\def\year{2022}\relax
\documentclass[letterpaper]{article} 
\usepackage{aaai22}  
\usepackage{times}  
\usepackage{helvet}  
\usepackage{courier}  
\usepackage[hyphens]{url}  
\usepackage{graphicx} 
\def\UrlFont{\rm}  
\usepackage{natbib}  
\usepackage{caption} 
\DeclareCaptionStyle{ruled}{labelfont=normalfont,labelsep=colon,strut=off} 
\usepackage[utf8]{inputenc} 
\usepackage{url}            
\usepackage{booktabs}       
\usepackage{amsfonts}       
\usepackage{nicefrac}       
\usepackage{microtype}      
\usepackage{xcolor}         
\usepackage{blindtext}

\usepackage{lipsum}
\usepackage{amsmath,amssymb,mathtools, amsthm}
\usepackage[noend]{algorithmic}
\usepackage{centernot}
\usepackage{scalerel} 
\usepackage{multirow}
\usepackage[inline]{enumitem}
\usepackage{tikz, subcaption}
\usepackage[T1]{fontenc}
\usetikzlibrary{shapes, arrows, arrows.meta, positioning}
\usepackage[ruled,vlined,linesnumbered]{algorithm2e}
\SetKw{Break}{break}
\usepackage{array}
\newcolumntype{M}[1]{>{\centering\arraybackslash}m{#1}}
\newcolumntype{N}{@{}m{0pt}@{}}

\DeclareMathAlphabet\mathbfcal{OMS}{cmsy}{b}{n}
\usepackage{todonotes}

\newtheorem{theorem}{Theorem}

\newtheorem{lemma}{Lemma}

\newenvironment{customlem}[1]{\renewcommand\thelemma{#1}\lemma}{\endlemma}
  
\newenvironment{customtheorem}[1]{\renewcommand\thetheorem{#1}\theorem}{\endtheorem}
  
\newtheorem{proposition}{Proposition}
\newenvironment{customprp}[1]{\renewcommand\theproposition{#1}\proposition}{\endproposition}
\newtheorem{definition}{Definition}

\newenvironment{myproof}[1][\proofname]{%
  \begin{proof}[#1]$ $\nobreak\ignorespaces
}{%
  \end{proof}
}

\newcommand{\independent}{\perp\mkern-9.5mu\perp}
\newcommand{\notindependent}{\centernot{\independent}}
\newcommand{\CI}[4]{#1 \independent_{#4} #2 \vert #3}
\newcommand{\notCI}[4]{#1 \notindependent_{#4} #2 \vert #3}
\newcommand{\dsep}[4]{ #1 \perp_{#4} #2 \vert #3}
\newcommand{\notdsep}[4]{#1 \centernot{\perp}_{#4} #2 \vert #3}

\newcommand{\MB}[1]{\textit{Mb}_{#1}}
\newcommand{\Mb}[2]{\textit{Mb}_{#2}(#1)}
\newcommand{\Pa}[2]{\textit{Pa}_{#2}(#1)}
\newcommand{\Ch}[2]{\textit{Ch}_{#2}(#1)}
\newcommand{\N}[2]{\textit{N}_{#2}(#1)}
\newcommand{\De}[2]{\textit{De}_{#2}(#1)}

\newcommand{\Cp}[2]{\textit{CP}_{#2}(#1)}
\newcommand{\din}[1]{\Delta_{\textit{in}}(#1)}

\newcommand{\V}{\mathbf{V}}
\newcommand{\OV}{\overline{\mathbf{V}}}
\newcommand{\E}{\mathbf{E}}
\newcommand{\G}{\mathcal{G}}

\newcommand{\PV}{P_\mathbf{V}}

\def\keywordname{{\bfseries Keywords}}%
\def\keywords#1{\par\addvspace\medskipamount{\rightskip=0pt
\def\and{\ifhmode\unskip\nobreak\fi\ $\cdot$
}\noindent\keywordname\enspace\ignorespaces#1\par}}

\title{Learning Bayesian Networks in the Presence of Structural Side Information}
\author {
    Ehsan Mokhtarian,\textsuperscript{\rm 1}
    Sina Akbari, \textsuperscript{\rm 1}
    Fateme Jamshidi,\textsuperscript{\rm 2}
    Jalal Etesami,\textsuperscript{\rm 1}
    Negar Kiyavash \textsuperscript{\rm 1 2}
}
\affiliations {
    \textsuperscript{\rm 1} Department of Computer and Communication Science, EPFL, Lausanne, Switzerland\\
    \textsuperscript{\rm 2} College of Management of Technology, EPFL, Lausanne, Switzerland\\
    \{ehsan.mokhtarian, sina.akbari, fateme.jamshidi, seyed.etesami, negar.kiyavash\}@epfl.ch
}
\begin{document}

\maketitle
\begin{abstract}
    We study the problem of learning a Bayesian network (BN) of a set of variables when structural side information about the system is available. It is well known that learning the structure of a general BN is both computationally and statistically challenging. However, often in many applications, side information about the underlying structure can potentially reduce the learning complexity. In this paper, we develop a recursive constraint-based algorithm that efficiently incorporates such knowledge (i.e., side information) into the learning process. In particular, we study two types of structural side information about the underlying BN: (I) an upper bound on its clique number is known, or (II) it is diamond-free. We provide theoretical guarantees for the learning algorithms, including the worst-case number of tests required in each scenario. As a consequence of our work, we show that bounded treewidth BNs can be learned with polynomial complexity. Furthermore, we evaluate the performance and the scalability of our algorithms in both synthetic and real-world structures and show that they outperform the state-of-the-art structure learning algorithms.
\end{abstract}

\section{Introduction}\label{sec:intro}
    Bayesian networks (BNs) are probabilistic graphical models that represent conditional dependencies in a set of random variables via directed acyclic graphs (DAGs).
    Due to their succinct representations and power to improve the prediction and to remove systematic biases in inference \cite{pearl2009causality,spirtes2000causation}, BNs have been widely applied in various areas including medicine \cite{flores2011incorporating}, bioinformatics \cite{friedman2000using}, ecology \cite{pollino2007parameterisation}, etc.
    Learning a BN from data is in general NP-hard \cite{chickering2004large}. 
    However, any type of side information about the network can potentially reduce the complexity of the learning task.
    
    BN structure learning algorithms are of three flavors:
    constraint-based, e.g., parent-child (PC) algorithm \cite{spirtes2000causation}, score-based, e.g., \cite{chickering2002optimal, solus2017consistency,notears,zhu2019causal}, and hybrid, e.g., MMHC algorithm \cite{tsamardinos2006max}. 
    
    
    Although constraint-based methods do not require any assumptions about the underlying generative model, they often require conditional independence (CI) tests with large conditioning sets or a large number of CI tests which grows exponentially as the number of variables increases\footnote{See \cite{scutari2014bayesian} for an overview on implementations of constraint-based algorithms.}. 
    Often in practice, we have side information about the network that can improve learning accuracy or reduce complexity. 
    We show in this work that such side information can reduce the learning complexity to polynomial in terms of the number of CI tests. 
    Our main contributions are as follows.
    \begin{itemize}[leftmargin=*]
        \item  We propose a constraint-based Recursive Structure Learning (RSL) algorithm to recover BNs. 
        In addition, we study two types of structural side information: (I) an upper bound on the clique number of the graph is known, or (II) the graph is diamond-free. In each case, we provide a learning algorithm.
        RSL follows a divide-and-conquer approach: it breaks the learning problem into several sub-problems that are similar to the original problem but smaller in size by eliminating \textit{removable} variables (see Definition \ref{def: removable}). Thus, in each recursion, both the size of the conditioning sets and the number of CI tests decrease. 
        
        \item 
        Learning BNs with bounded treewidth has recently attracted attention.
        Works such as \cite{korhonen2013exact,nie2014advances,ramaswamy2021turbocharging} aim to develop learning algorithms for BNs when an upper bound on the treewidth of the graph is given as side information. 
        Assuming bounded treewidth is  more restrictive than bounded clique number assumption, i.e., having a bound on the treewidth implies an upper bound on the clique number of the network.
        Hence, our proposed algorithm with structural side information of type (I) can also learn bounded treewidth BNs. 
        However, our algorithm has polynomial complexity, while the state-of-the-art exact learning algorithms have exponential complexity.
        
        \item We show that when the clique number of the underlying BN is upper bounded by $m$, i.e., $\omega(\G)\!\leq\! m$ (See Table \ref{table of notations} for the graphical notations), our algorithm requires $\mathcal{O}(n^2+n\Delta_{in}^{m+1})$ CI tests (Theorem \ref{thm: upperbound clique}). 
        Furthermore, when the graph is diamond-free, our algorithm requires $\mathcal{O}(n^2+n\Delta_{in}^3)$ CI tests (Theorem \ref{thm: upperbound}). 
        These bounds significantly improve over the state of the art.
    \end{itemize}
    
    \begin{table}[t]
        \centering
        \begin{tabular}{N M{1.4cm}|M{5.5cm}}
     		\toprule
            & $n$ & Number of variables\\
            & $\Delta(\G)$ & Maximum degree of DAG $\G$\\
            & $\Delta_{in}(\G)$ & Maximum in-degree of DAG $\G$\\
            & $\omega(\G)$ & Clique number of graph $\G$\\
            & $N_{\G}(X)$ & Neighbors of $X$ in DAG $\G$\\
            & $Ch_{\G}(X)$ & Children of $X$ in DAG $\G$\\
            & $Pa_{\G}(X)$ & Parents of $X$ in DAG $\G$\\
            & $CP_{\G}(X)$ & Co-parents of $X$ in DAG $\G$\\
            & $Mb_{\mathbf{V}}(X)$ & Markov boundary of $X$ among set $\mathbf{V}$\\
            & $\alpha(\G)$ & Maximum Mb size of $\G$\\
    		\bottomrule
    	\end{tabular}
    	\caption{Graphical notations that we use in this paper.}
	    \label{table of notations}
    \end{table}
    

    \paragraph{Related work:}
    Herein, we review the relevant work on BN learning methods as well as those with side information.
    
    The PC algorithm \cite{spirtes2000causation} is a classical example of constraint-based methods that requires $\mathcal{O}(n^\Delta)$ number of CI tests.
    CS \cite{pellet2008using} and MARVEL \cite{mokhtarian2021recursive} are two examples that focus on BN structure learning with small number of CI tests by using the Markov boundaries (Mbs).
    This results in $\mathcal{O}(n^2 2^\alpha)$ and $\mathcal{O}(n^2+n \Delta_\text{in}^2 2^{\Delta_\text{in}})$ number of CI tests for  CS and MARVEL, respectively.
    On the other hand, methods such as GS \cite{margaritis1999bayesian}, MMPC \cite{mmpc}, and HPC \cite{HPC} focus on reducing the size of the conditioning sets in their CI tests.
    However, the aforementioned methods are not equipped to take advantage of side information.
    Table \ref{table} compares the complexity of various constraint-based algorithms in terms of their CI tests. RSL$_\omega$ and RSL$_D$ are our proposed algorithms when an upper bound on the clique number is given and when the BN is diamond-free, respectively. 
    Note that in general, $\Delta_\text{in} \leq \Delta \leq \alpha$, and in a DAG with a constant in-degree, $\Delta$ and $\alpha$ can grow linearly with the number of variables. 
    \begin{table}[ht]
        \centering
        \begin{tabular}{N M{3cm}|M{3.5cm}}
     		\toprule
    	    &Algorithm &  \#CI tests \\
    		\hline
    		&PC & $\mathcal{O}(n^\Delta)$\\
    		&GS  & $\mathcal{O}(n^2+ n \alpha^2 2^\alpha)$\\
    		&MMPC, CS  & $\mathcal{O}(n^22^\alpha$)\\
    		&MARVEL  & $\mathcal{O}(n^2+n\Delta_{in}^22^{\Delta_{in}})$\\
    		&RSL$_D$ & $\mathcal{O}(n^2+n\Delta_{in}^3)$\\
    		&RSL$_\omega$ & $\mathcal{O}(n^2+n\Delta_{in}^{m+1})$\\
    		\bottomrule
    	\end{tabular}
    	\caption{Required number of CI tests in the worst case by various algorithms.}\label{table}
    \end{table}
    
    Side information about the underlying generative model has been harnessed for structure learning in limited fashion, e.g., \cite{sesen2013bayesian, flores2011incorporating, oyen2016bayesian, mclachlan2020bayesian}. 
    As an example,  \cite{takeishi2020knowledge} propose an approach to incorporate side knowledge about feature relations into the learning process.
    \cite{shimizu2019non} and \cite{sondhi2019reduced} study the structure learning problem when the data is from a linear structural equation model and propose LiNGAM and reduced PC algorithms, respectively.
    \cite{Sparsecausal, zheng2020learning} consider learning sparse BNs. 
    In particular, \cite{Sparsecausal} show that in sparse setting, BN recovery is no longer NP-hard, even in the presence of unobserved variables. That is for sparse graphs with maximum node degree of $\Delta$, a sound and complete BN can be obtained by performing $\mathcal{O}(n^{2(\Delta+2)})$ CI tests.
    
    Side information has been incorporated into score-based methods in limited fashions too, e.g., \cite{chen2016learning, constraints, bartlett2017integer}. 
    The side information in the aforementioned works is in the form of ancestral constraints which are about the absence or presence of a directed path between two vertices in the underlying BN. 
    \cite{bartlett2017integer} cast this problem as an integer linear program. 
    The proposed method by \cite{chen2016learning} recovers the network with guaranteed optimality but it does not scale beyond 20 random variables. 
    The method by \cite{constraints} scales up to 50 variables but it does not provide any optimality guarantee. 

    Another related problem is optimizing $\sum_{v\in \V}f_v(Pa(v))$ over a set of DAGs with vertices $\V$ and parent sets $\{Pa(v)\}_{v\in\V}$. In this problem  $\{f_v(\cdot)\}_{v\in\V}$ is a set of predefined local score functions.
    This problem is NP-hard \cite{chickering2004large}. 
    Note that the BN structure learning can be formulated as a special case of this problem by selecting appropriate local score functions. 
    \cite{korhonen2013exact} introduce an exact algorithm for solving this problem with complexity $3^n n^{t+\mathcal{O}(1)}$ under a constraint that the optimal BN has treewidth at most $t$.
    \cite{elidan2008learning} propose a heuristic algorithm that finds a sub-optimal DAG with bounded treewidth which runs in time polynomial in $n$ and $t$. 
    Knowing a bound on the treewidth is yet another type of structural side information that is more restrictive\footnote{In General, Treewidth $\!+1\!\geq\! \omega$, \cite{bodlaender1993pathwidth}.} than our structural assumptions. 
    Therefore, our algorithm in Section \ref{sec: generalization} can learn bounded treewidth BNs with polynomial complexity, i.e., $\mathcal{O}(n^2+n\Delta_{in}^{t+2})$, where $t$ is a bound on the treewidth and $\Delta_{in}<n$. 
    
    \cite{korhonen2015tractable} is another score-based method that study the BN structure learning when an upper bound $k$ on the vertex cover number of the underlying BN is available. Their algorithm has complexity $4^kn^{2k+\mathcal{O}(1)}$. Since the vertex cover number of a graph is greater than its clique number minus one, then our algorithm in Section \ref{sec: generalization} can also recover a bounded vertex cover numbers BN with complexity $\mathcal{O}(n^2+n\Delta_{in}^{k+2})$.
    \cite{gruttemeier2020learning} consider the structural constraint that the moralized graph can be transformed into a graph with maximum degree one by at most $r$ vertex deletions. 
    They show that under this constraint, an optimal network can be learned in $n^{\mathcal{O}(r^2)}$ time. 
    
\section{Preliminaries}\label{sec: pre}
    Throughout the paper, we use capital letters for random variables and bold letters for sets.
    Also, the graphical notations are presented in Table \ref{table of notations}.
    
    A graph is defined as a pair $\G=(\V, \E)$ where $\V$ is a finite set of vertices and $\E$ is the set of edges. 
    If $\E$ is a set of unordered pairs of vertices, the graph is called \emph{undirected} and if it is a set of ordered pairs, it is called \emph{directed}.
    An undirected graph is called \emph{complete} if $\E$ contains all edges. 
    A \emph{directed acyclic graph} (DAG) is a directed graph with no directed cycle. 
    In an edge $(X,Y)\in\E$ (or $\{X,Y\}\in\E$, in case of an undirected graph), the vertices $X$ and $Y$ are the endpoints of that edge and they are called \emph{neighbors}. 
    Let $\G=(\V, \E)$ be a (directed or undirected) graph and $\OV\subseteq \V$, then the \emph{induced subgraph} $\G[\OV]$ is the graph whose vertex set is $\OV$ and whose edge set consists of all of the edges in $\E$ that have both endpoints in $\OV$.
    The \emph{skeleton} of a graph ${\G}=(\V,\E)$ is its undirected version.
    The \emph{clique number} of an undirected graph $\G$ is the number of vertices in the largest induced subgraph of $\G$ that is complete. 
    
     
    Let $\textbf{X}, \textbf{Y}$, and $\mathbf{S}$ be three disjoint subsets of $\V$. 
    We use $\dsep{\textbf{X}}{\textbf{Y}}{\mathbf{S}}{\G}$ to indicate  $\mathbf{S}$ d-separates\footnote{See Appendix \ref{sec: apd-dsep} for the definition.}
    $\textbf{X}$ and $\textbf{Y}$ in $\G$. 
    In this case, the set $\mathbf{S}$ is called a \textit{separating} set for $\textbf{X}$ and $\textbf{Y}$.
    Suppose $\PV$ is the joint probability distribution of $\V$.
    We use $\CI{\textbf{X}}{\textbf{Y}}{\mathbf{S}}{\PV}$ to denote the Conditional Independence (CI) of  $\textbf{X}$ and $\textbf{Y}$ given $\mathbf{S}$. Also, a CI test refers to detecting whether $\CI{X}{Y}{\mathbf{S}}{\PV}$.
    A DAG $\G$ is said to be an \emph{independency map (I-map)} of $\PV$ if for every three disjoint subsets of vertices $\textbf{X}, \textbf{Y}$, and $\mathbf{S}$ we have $\dsep{X}{Y}{\mathbf{S}}{\G} \Rightarrow \CI{X}{Y}{\mathbf{S}}{\PV}$.
    A DAG $\G$ is a \emph{minimal} I-map of $\PV$ if it is an I-map of $\PV$ and the resulting DAG after removing any edge is no longer an I-map of $\PV$.
    A DAG $\G=(\V,\E)$ is called a \emph{Bayesian network} (BN) of $\PV$, if and only if $\G$ is a minimal I-map of $\PV$.
    The joint probability distribution $\PV$ with a BN $\G =(\V,\E)$ satisfies the Markov factorization property, that is $\PV = \prod_{X\in \V} \PV(X \vert \Pa{X}{\G})$  \cite{pearl1988probabilistic}.
    
    A joint distribution $\PV$ may have several BNs. 
    The \textit{Markov equivalence class} (MEC) of $\PV$, denoted by $\langle\PV\rangle$, is the set of all its BNs. 
    It has been shown that two DAGs belong to a MEC if and only if they share the same skeleton and the same set of v-structures\footnote{Three vertices $X,Y$, and $Z$ form a \emph{v-structure} if $X\to Y \gets Z$ while $X$ and $Z$ are not neighbors.} \cite{pearl2009causality}.
    A MEC $\langle\PV\rangle$ can be uniquely represented by a partially directed graph\footnote{It is a graph with both directed and undirected edges.} called \emph{essential graph}.
    A DAG $\G=(\V,\E)$ is called a \emph{dependency map (D-map)} of $\PV$ if for every three disjoint subsets of vertices $\textbf{X}, \textbf{Y}$, and $\mathbf{S}$,  $\CI{\textbf{X}}{\textbf{Y}}{\mathbf{S}}{\PV}$ implies $\dsep{\textbf{X}}{\textbf{Y}}{\mathbf{S}}{\G}$. This property is also known as \emph{faithfulness} in the causality literature \cite{pearl2009causality}.
    Furthermore, $\G$ is called a \emph{perfect map} if it is both an I-map and a D-map of $\PV$, i.e., $\dsep{\textbf{X}}{\textbf{Y}}{\mathbf{S}}{\G} \iff \CI{\textbf{X}}{\textbf{Y}}{\mathbf{S}}{\PV}$. 
    Note that if $\G$ is perfect map of $\PV$, then it belongs to $\langle\PV\rangle$, i.e., a perfect map is a BN. 

    \paragraph{Problem description:}
    The BN structure learning problem involves identifying $\langle\PV\rangle$ from $\PV$ on the population-level or from a set of samples of $\PV$.
    As mentioned earlier, the constraint-based methods perform this task using a series of CI tests.
    In this paper, we consider the BN structure learning problem using a constraint-based method, when we are given structural side information about the underlying DAG. 

\section{Learning Bayesian networks recursively} \label{sec: RSL}
    Suppose $\G =(\V,\E)$ is a perfect map of $\PV$ and let $\mathcal{H}$ denote its skeleton. 
    Recall that learning $\langle\PV\rangle$ requires recovering $\mathcal{H}$ and the set of v-structures of $\G$. 
    It has been shown that finding a separating set for each pair of non-neighbor vertices in $\G$ suffices to recover its set of v-structures \cite{spirtes2000causation}. 
    Thus, we propose an algorithm called Recursive Structure Learning (\textbf{RSL}) that recursively finds $\mathcal{H}$ along with a set of separating sets $\mathbfcal{S}_{\V}$ for non-neighbor vertices in $\V$. The pseudocode of \textbf{RSL} is presented in Algorithm \ref{alg: RSL}.
    \begin{algorithm}[ht]
        \caption{Recursive Structure Learning (RSL).}
        \label{alg: RSL}
        \begin{algorithmic}[1]
            \STATE {\bfseries Input:} $\V,\, \PV, \text{ SideInfo}$
            \STATE $\MB{\V} \gets  \textbf{ComputeMb}(\V,\, \PV)$
            \STATE $(\mathcal{H},\, \mathbfcal{S}_{\V}) \gets \textbf{RSL}(\V,\, \PV,\, \MB{\V}, \text{ SideInfo})$
        \end{algorithmic}
        \hrulefill
        \begin{algorithmic}[1]
            \STATE {\bfseries RSL}$(\OV,\, P_{\OV},\, \MB{\OV}, \text{ SideInfo})$
            \IF{$|\OV|=1$}
                \RETURN $((\OV,\varnothing), \varnothing)$
            \ELSE
                \STATE $X \gets \textbf{FindRemovable}(\OV,\,  P_{\OV},\, \MB{\OV}, \text{ SideInfo})$
                \STATE $(\N{X}{\G[\OV]},\, \mathbfcal{S}_X) \gets \textbf{FindNeighbors}(X,\, \OV,\,  P_{\OV},\, \Mb{X}{\OV}, \text{ SideInfo})$ 
                \STATE $\MB{\OV\setminus\{X\}} \gets \textbf{UpdateMb}(X,\, P_{\OV} ,\, \N{X}{\G[\OV]}, \, \MB{\OV})$
                \STATE $(\mathcal{H}[\OV\setminus\{X\}],\, \mathbfcal{S}_{\OV\setminus\{X\}}) \gets \textbf{RSL}(\OV \setminus \{X\},\, P_{\OV\setminus\{X\}},\, \MB{\OV\setminus\{X\}}, \text{ SideInfo})$
                \STATE Construct $\mathcal{H}[\OV]$ by $\mathcal{H}[\OV\setminus\{X\}]$ and undirected edges between $X$ and $\N{X}{\G[\OV]}$.
                \STATE $\mathbfcal{S}_{\OV} \gets \mathbfcal{S}_{\OV\setminus\{X\}} \cup \mathbfcal{S}_X$
                \RETURN $(\mathcal{H}[\OV],\, \mathbfcal{S}_{\OV})$
            \ENDIF
        \end{algorithmic}
    \end{algorithm}
    
    \textbf{RSL}'s inputs comprise a subset $\OV \subseteq \V$ with its joint distribution $P_{\OV}$\footnote{In practice, the finite sample data at hand is used instead of $P_{\OV}$.} 
    such that $\G[\OV]$ is a perfect map of $P_{\OV}$, and their Markov boundaries $\MB{\OV}$ (see Definition \ref{def: Mb}), along with structural side information, which can be either diamond-freeness, or an upper bound on the clique number.
    In this case, \textbf{RSL} outputs $\mathcal{H}[\OV]$ and a set of separating sets $\mathbfcal{S}_{\OV}$ for non-neighbor vertices in $\OV$.
    The \textbf{RSL} consists of three main sub-algorithms: \textbf{FindRemovable}, \textbf{FindNeighbors}, and \textbf{UpdateMb}. 
    It begins by calling \textbf{FindRemovable} in line 5 to find a vertex $X\in\OV$ such that the resulting graph after removing $X$ from the vertex set, $\G[\OV\setminus\{X\}]$, remains a perfect map of $P_{\OV\setminus \{X\}}$.
    Afterwards, in line 6, \textbf{FindNeighbors} identifies the neighbors of $X$ in $\G[\OV]$ and a set of separating sets for $X$ and each of its non-neighbors in this graph.
    In lines 7 and 8, \textbf{RSL} updates the Markov boundaries and calls itself to learn the remaining graph after removing vertex $X$, i.e.,  $\G[\OV\setminus\{X\}]$, respectively.
    The two functions \textbf{FindRemovable} and \textbf{FindNeighbors} take advantage of the provided side information, as we shall discuss later.

    As mentioned above, it is necessary for $\G[\OV]$ to remain a perfect map of $P_{\OV}$ at each iteration. This cannot be guaranteed if $X$ is chosen arbitrarily. \cite{mokhtarian2021recursive} introduced the notion of removability in the context of causal graphs and showed that removable variables are the ones that preserve the perfect map assumption after the distribution is marginalized over them.
    In this work, we introduce a similar concept in the context of BN structure recovery.
    \begin{definition}[Removable] \label{def: removable}
        Suppose $\G=(\V,\E)$ is a DAG and $X\in \V$.
        Vertex $X$ is called removable in $\G$ if the d-separation relations in $\G$ and $\G[\V \setminus \{X\}]$ are equivalent over $\V \setminus \{X\}$. 
        That is, for any vertices $Y,Z \in \V \setminus \{X\}$ and $\mathbf{S} \subseteq \V \setminus\{X,Y,Z\}$,
        \begin{equation}\label{eq: d-sepEquivalence}
          \dsep{Y}{Z}{\mathbf{S}}{\G}
    	    \iff
    	    \dsep{Y}{Z}{\mathbf{S}}{\G[\V \setminus \{X\}]}.
    	\end{equation}
    \end{definition}
    
    \begin{proposition} \label{prp: removable}
        Suppose $\G=(\V,\E)$ is a perfect map of $\PV$. For each variable $X\in \V$, $\G[\V \setminus \{X\}]$ is a perfect map of $P_{\V\setminus\{X\}}$ if and only if $X$ is a removable vertex in $\G$. 
    \end{proposition}
    All proofs appear in Appendix \ref{sec: apd_proofs}.
    
    \paragraph{Markov boundary (Mb):}
    Our proposed algorithm uses the notion of Markov boundary.
        \begin{definition}[Mb] \label{def: Mb}
           Suppose $\PV$ is the joint distribution on $\V$. 
           The Mb of $X\in \V$, denoted by $\Mb{X}{\V}$, is a minimal set $\mathbf{S} \subseteq \V\setminus\{X\}$ s.t. $\CI{X}{\V \setminus (\mathbf{S} \cup \{X\})}{\mathbf{S}}{\PV}$. 
           We denote $(\Mb{X}{\V}\!:\:X\in \V)$ by $\MB{\V}$.
        \end{definition}
        \begin{definition}[co-parent]
            Two non-neighbor variables are called co-parents in $\G$, if they share at least one child. For $X\in \V$, the set of co-parents of $X$ is denoted by $\Cp{X}{\G}$.
        \end{definition}
    If $\G$ is a perfect map of $\PV$, for every vertex $X\in\V$, $\Mb{X}{\V}$ is unique \cite{pearl1988probabilistic} and it is equal to
	\begin{equation} \label{eq: Mb with perfect map assumption}
	    \Mb{X}{\V}= \Pa{X}{\G} \cup \Ch{X}{\G} \cup \Cp{X}{\G}.
	\end{equation}
	
	The subroutines \textbf{FindRemovable} and \textbf{FindNeighbors} need the knowledge of Mbs to perform their tasks. 
	Several constraint-based and scored-based algorithms have been developed in  literature such as TC \cite{pellet2008using}, GS \cite{margaritis1999bayesian}, and others \cite{tsamardinos2003algorithms} that can recover the Mbs of a set of random variables. 
	Initially, any of the aforementioned algorithms could be used in \textbf{ComputeMb} to find $\MB{\V}$ and pass it to the \textbf{RSL}. 
	After eliminating a removable vertex $X$, the Mbs of the remaining graph will change. Therefore, we need to update and pass $\MB{\OV \setminus \{X\}}$ to the next recall of \textbf{RSL}.
	This is done by function \textbf{UpdateMb} in line 7 of Algorithm \ref{alg: RSL}.
	We propose Algorithm \ref{alg: update Mb} for \textbf{UpdateMb} and prove its soundness and complexity in Proposition \ref{prp: updatemb}. 
	Further discussion about this algorithm is presented in Appendix \ref{sec: apd_alg_updat}.
	\begin{algorithm}[ht]
    \caption{Updates Markov boundaries (Mbs).}
    \label{alg: update Mb}
    \begin{algorithmic}[1]
        \STATE {\bfseries UpdateMb}$(X,\, P_{\OV} ,\, \N{X}{\G[\OV]}, \, \MB{\OV})$
        \STATE $\MB{\OV \setminus \{X\}} \gets (\Mb{Y}{\OV}\!:\: Y\in \OV \setminus \{X\})$ 
        \FOR{$Y\in \Mb{X}{\OV}$}
            \STATE Remove $X$ from $\Mb{Y}{\OV \setminus \{X\}}$.
        \ENDFOR
        \IF{$\N{X}{\G[\OV]}=\Mb{X}{\OV}$}
            \FOR{$Y,Z \in \N{X}{\G[\OV]}$} 
            \IF{$\CI{Y}{Z}{\Mb{Y}{\OV \setminus \{X\}}\setminus \{Y,Z\}}{P_{\OV}}$}
                \STATE Remove $Z$ from $\Mb{Y}{\OV \setminus \{X\}}$ 
                \STATE Remove $Y$ from $\Mb{Z}{\OV \setminus \{X\}}$
            \ENDIF
        \ENDFOR
        \ENDIF
        \RETURN $\MB{\OV \setminus \{X\}}$
    \end{algorithmic}
    \end{algorithm}
	\begin{proposition}\label{prp: updatemb}
	    Suppose $\mathcal{G}[\OV]$ is a perfect map of $P_{\OV}$ and $X$ is a removable variable in $\mathcal{G}[\OV]$. Algorithm \ref{alg: update Mb} correctly finds $\MB{\OV \setminus \{X\}}$ by performing at most $\binom{|\N{X}{\G[\OV]}|}{2}$ CI tests. 
	\end{proposition}

\section{Learning BN with known upper bound on the clique number} \label{sec: generalization}



    In this section, we consider the BN structure learning problem when we are given an upper bound $m$ on the clique number of the underlying BN and propose algorithms \ref{alg: FindRemovable clique} and \ref{alg: FindNeighbors clique} to efficiently find removable vertices along with their neighbors. We denote the resulting \textbf{RSL} with these implementations of \textbf{FindRemovable} and \textbf{FindNeighbors} by RSL$_\omega$. First, we present a sufficient removability condition in such networks, which is the foundation of Algorithm \ref{alg: FindRemovable clique}.
    \begin{lemma} \label{lemma: FindRemovable clique}
        Suppose $\G=(\OV,\E)$ is a DAG and a perfect map of $P_{\OV}$ such that 
        $\omega(\G)\leq m$. Vertex $X\in \OV$ is removable in $\G$ if for any $\mathbf{S}\subseteq\Mb{X}{\OV}$ with $\left\vert\mathbf{S}\right\vert\leq m-2$, we have
        \begin{equation} \label{eq: removable-clique}
        \begin{aligned} 
            &\forall Y,Z \in \Mb{X}{\OV} \setminus \mathbf{S}\!:\\
            &\hspace{.8cm} \notCI{Y}{Z}{\big(\Mb{X}{\OV} \cup \{X\}\big) \setminus\big(\{Y,Z\}\cup\mathbf{S}\big)}{P_{\OV}},\\
            & \text{and}\ \forall Y \in \Mb{X}{\OV}\setminus \mathbf{S}\!: \\
            &\hspace{.8cm} \notCI{X}{Y}{\Mb{X}{\OV} \setminus (\{Y\}\cup\mathbf{S})}{P_{\OV}}. 
        \end{aligned}
        \end{equation}
        Also, the set of vertices that satisfy Equation \eqref{eq: removable-clique} is nonempty.
    \end{lemma}

    \begin{algorithm}
    \caption{Finds a removable vertex.}
    \label{alg: FindRemovable clique}
        \begin{algorithmic}[1]
        \STATE {\bfseries FindRemovable}($\OV,\, P_{\OV},\, \MB{\OV},\, \text{ SideInfo (m)}$)
        \STATE $\textbf{X}=(X_{1},..., X_{|\OV|})\gets \OV$
        \STATE Sort $\textbf{X}$ s.t.
        $|\Mb{X_1}{\OV}|\leq |\Mb{X_2}{\OV}|\dots \leq |\Mb{X_{|\OV|}}{\OV}|.$
        \FOR{$i=1$ to $|\OV|$}
            \IF{\eqref{eq: removable-clique} holds for $X=X_i$}
                \RETURN $X_i$
            \ENDIF
        \ENDFOR
        \end{algorithmic}
    \end{algorithm}
    
    Algorithm \ref{alg: FindRemovable clique} first sorts the vertices in $\OV$ based on their Mb size and checks their removability, starting with the vertex with the smallest Mb.
    This ensures that both the number of CI tests and the size of the conditioning sets remain bounded. 
    
    \begin{proposition}\label{prop:bounded_clique_neighbors}
        Suppose $\G[\OV]$ is a DAG and a perfect map of $P_{\OV}$ s.t. $\omega(\mathcal{G[\OV]})\leq m$. Algorithm \ref{alg: FindRemovable clique} returns a removable vertex in $\G[\OV]$ by performing $\mathcal{O}(|\OV| \din{\G[\OV]}^{m})$ CI tests.
    \end{proposition}
    We now turn to the function \textbf{FindNeighbors}. Recall that the purpose of this function is to find the neighbors of a removable vertex $X$ and its separating sets.
    Since for every vertex $Y\not\in\Mb{X}{\OV}$, we have $\CI{Y}{X}{\Mb{X}{\OV}}{P_{\OV}}$,  $\Mb{X}{\OV}$ is a separating set for all vertices outside of $\Mb{X}{\OV}$. 
    Therefore, it suffices to find the non-neighbors of $X$ within $\Mb{X}{\OV}$ or equivalently the co-parents of $X$.
    Next result characterizes the co-parents of a removable vertex $X$.
    \begin{lemma} \label{lemma: FindNeighbors clique}
        Suppose $\G[\OV]$ is a DAG and a perfect map of $P_{\OV}$ with  $\omega(\mathcal{G[\OV]})\leq m$. Let $X\in \OV$ be a vertex that satisfies Equation \eqref{eq: removable-clique} and $Y\in \Mb{X}{\OV}$. Then, $Y\in \Cp{X}{\G}$ iff
        \begin{multline} \label{eq: neighborhood-clique}
            \exists \mathbf{S}\subseteq\Mb{X}{\OV}\setminus\{Y\}\!: \\ \left\vert\mathbf{S}\right\vert=(m-1)\,,\ \ \CI{X}{Y}{\Mb{X}{\OV}\setminus(\{Y\}\cup\mathbf{S})}{P_{\OV}}.
        \end{multline}
    \end{lemma}
     Algorithm \ref{alg: FindNeighbors clique} is designed based on Lemma \ref{lemma: FindNeighbors clique}. We use $\langle X\vert \mathbf{Z} \vert Y \rangle$ to denote that $\mathbf{Z}$ is a separating set for $X$ and $Y$.

    \begin{algorithm}
        \caption{Finds neighbors and separating sets in a graph with bounded clique number.}
        \label{alg: FindNeighbors clique}
        \begin{algorithmic}[1]
            \STATE {\bfseries FindNeighbors}($X,\, \OV,\, P_{\OV},\, \Mb{X}{\OV}, \text{ SideInfo(m)}$)
            \FOR{$Y\in \OV\setminus \Mb{X}{\OV}$}
                \STATE Add $\langle X \vert \Mb{X}{\OV} \vert Y  \rangle$ to $\mathbfcal{S}_X$. 
            \ENDFOR
            \FOR{$Y \in \Mb{X}{\OV}$}
                \IF{\eqref{eq: neighborhood-clique} holds}
                    \STATE Add $\langle X \vert \Mb{X}{\OV}\setminus\{Y,Z\}\vert Y \rangle$ to $\mathbfcal{S}_X$. 
                \ELSE
                    \STATE Add $Y$ to $\N{X}{\G[\OV]}$.
                \ENDIF
            \ENDFOR
            \RETURN $(\N{X}{\G[\OV]},  \mathbfcal{S}_X)$
        \end{algorithmic}
    \end{algorithm}


    
    \begin{theorem}\label{thm: upperbound clique}
        Suppose $\G=(\V,\E)$ is a DAG and a perfect map of $\PV$ with $\omega(\mathcal{G})\leq m$. Then, \textbf{RSL} (Algorithm \ref{alg: RSL}) with sub-algorithms \ref{alg: FindRemovable clique} and \ref{alg: FindNeighbors clique} is sound and complete, and performs $\mathcal{O}(|\V|^2 \din{\G}^{m})$ CI tests.
    \end{theorem}

\section{Learning BN without side information} \label{sec: without side information}
    We showed in Theorem \ref{thm: upperbound clique} that if the upper bound on the clique number is correct, i.e., $\omega(\G)\leq m$, then RSL$_\omega$ learns the DAG correctly. But what happens if $\omega(\G)> m$?
    In this case, there are two possibilities: either Algorithm \ref{alg: FindRemovable clique} fails to find any removables and consequently, RSL$_\omega$ fails or RSL$_\omega$ terminates with output $(\mathcal{\Tilde{H}},\mathbfcal{S}_{\V})$.
    Next result shows that the clique number of $\mathcal{\Tilde{H}}$ is greater or equal to $\omega(\G)$ and thus, it is strictly larger than $m$. 
    \begin{proposition}[Verifiable] \label{prp: verifiable}
        Suppose $\G=(\V,\E)$ is a DAG with skeleton $\mathcal{H}$ that is a perfect map of $\PV$. 
        If the \textbf{RSL} with sub-algorithms \ref{alg: FindRemovable clique} and \ref{alg: FindNeighbors clique}, and input $m>0$ terminates, then the clique number of the learned skeleton is 
        at least  $\omega(\mathcal{G})$.
    \end{proposition}
    This result implies that executing RSL$_\omega$ with input $m$ either outputs a graph with clique number at most $m$, which is guaranteed to be the true BN, or indicates that the upper bound $m$ is incorrect.
    As a result, we can design Algorithm \ref{alg: without side information} using RSL$_\omega$ when no bound on the clique number is given.
    \begin{algorithm}
        \caption{Learns BN without side information.}
        \label{alg: without side information}
        \begin{algorithmic}[1]
            \STATE {\bfseries Input:} $\V,\, \PV$
            \STATE $\MB{\V} \gets  \textbf{ComputeMb}(\V,\, \PV)$
            \FOR{$m$ from 1 to $|\V|$}
                \STATE $\hat{\mathcal{G}} \gets \textbf{RSL}(\V,\, \PV,\, \MB{\V},\, \text{ SideInfo(m)})$
                \IF{\textbf{RSL} terminates and $\omega(\hat{\mathcal{G}})\leq m$}
                    \RETURN $\hat{\mathcal{G}}$
                \ENDIF
            \ENDFOR
        \end{algorithmic}
    \end{algorithm}

\section{Learning diamond-free BNs} \label{sec: diamond-free}
In this section, we consider a well-studied class of graphs, namely diamond-free graphs.
These graphs appear in many real-world applications (see Appendix \ref{sec: apd_rep}).
Diamond-free graphs also occur with high probability in a wide range of random graphs. 
For instance, an Erdos-Renyi graph G(n,p) is diamond-free with high probability, if $pn^{0.8}\to 0$ (See Lemma \ref{lemma: erdos-renyi} in Section \ref{sec: discussion}).
Various NP-hard problems such as maximum weight stable set, maximum weight clique, domination and coloring have been shown to be linearly or polynomially solvable for diamond-free graphs \cite{brandstadt2004p5, dabrowski2017colouring}.
We show that the structure learning problem for diamond-free graphs is also polynomial-time solvable.
    \begin{definition}[diamond-free graphs]
        The graphs depicted in Figure \ref{fig: diamond} are called diamonds. A diamond-free graph is a graph that contains no diamond as an induced subgraph.
    \end{definition}
Note that triangle-free graphs are a subset of diamond-free graphs.
From Section \ref{sec: generalization}, we know that RSL$_\omega$ with $m=2$ can lean a triangle-free BN with complexity $\mathcal{O}(|\V|^2\din{\G}^2)$.
Herein, we propose new subroutines for \textbf{FindRemovable} and \textbf{FindNeighbors} with which, RSL can learn diamond-free BNs with the same complexity as triangle-free networks. 
We start with providing a necessary and sufficient condition for removability in a diamond-free graph.
    \begin{figure}[ht] 
	    \centering
		\tikzstyle{block} = [circle, inner sep=1.3pt, fill=black]
		\tikzstyle{input} = [coordinate]
		\tikzstyle{output} = [coordinate]
		\begin{subfigure}[b]{0.14\textwidth}
    		\centering
            \begin{tikzpicture}
                \tikzset{edge/.style = {->,> = latex',-{Latex[width=2mm]}}}
                \node[block] (A) at  (0,1) {};
                \node[] ()[above = -0.06cm of A]{$A$};
                \node[block] (B) at  (-0.9,0.5) {};
                \node[] ()[above left = -0.05cm and -0.2 cm of B]{$B$};
                \node[block] (C) at  (1,0.5) {};
                \node[] ()[above right = -0.05cm and -0.2 cm of C]{$C$};
                \node[block] (D) at  (0,0) {};
                \node[] ()[below = -0.06cm of D]{$D$};
                \draw[edge] (A) to (B);
                \draw[edge] (A) to (C);
                \draw[edge] (A) to (D);
                \draw[edge] (B) to (D);
                \draw[edge] (C) to (D);
            \end{tikzpicture}
        \end{subfigure}\hfill
        \begin{subfigure}[b]{0.14\textwidth}
    		\centering
            \begin{tikzpicture}
                \tikzset{edge/.style = {->,> = latex',-{Latex[width=2mm]}}}
                \node[block] (A) at  (0,1) {};
                \node[] ()[above = -0.06cm of A]{$A$};
                \node[block] (B) at  (-0.9,0.5) {};
                \node[] ()[above left = -0.05cm and -0.2 cm of B]{$B$};
                \node[block] (C) at  (1,0.5) {};
                \node[] ()[above right = -0.05cm and -0.2 cm of C]{$C$};
                \node[block] (D) at  (0,0) {};
                \node[] ()[below = -0.06cm of D]{$D$};
                \draw[edge] (A) to (B);
                \draw[edge] (C) to (A);
                \draw[edge] (A) to (D);
                \draw[edge] (B) to (D);
                \draw[edge] (C) to (D);
            \end{tikzpicture}
        \end{subfigure}\hfill
        \begin{subfigure}[b]{0.14\textwidth}
    		\centering
            \begin{tikzpicture}
                \tikzset{edge/.style = {->,> = latex',-{Latex[width=2mm]}}}
                \node[block] (A) at  (0,1) {};
                \node[] ()[above = -0.06cm of A]{$A$};
                \node[block] (B) at  (-0.9,0.5) {};
                \node[] ()[above left = -0.05cm and -0.2 cm of B]{$B$};
                \node[block] (C) at  (1,0.5) {};
                \node[] ()[above right = -0.05cm and -0.2 cm of C]{$C$};
                \node[block] (D) at  (0,0) {};
                \node[] ()[below = -0.06cm of D]{$D$};
                \draw[edge] (B) to (A);
                \draw[edge] (C) to (A);
                \draw[edge] (A) to (D);
                \draw[edge] (B) to (D);
                \draw[edge] (C) to (D);
            \end{tikzpicture}
        \end{subfigure}
        \caption{Diamond graphs.}
        \label{fig: diamond}
    \end{figure}
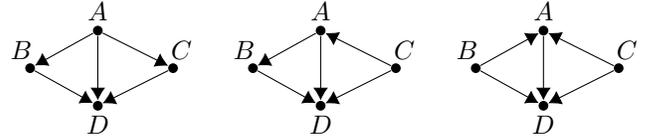 
    \begin{lemma} \label{lemma: FindRemovable}
        Suppose $\G=(\OV,\E)$ is a diamond-free DAG and a perfect map of $P_{\OV}$. Vertex $X\in \OV$ is removable in $\G$ if and only if $\:\forall Y,Z \in \Mb{X}{\OV}\!:$
        \begin{equation}\label{eq: removable-diamond}
            \notCI{Y}{Z}{(\Mb{X}{\OV} \cup \{X\} )\setminus \{Y,Z\}}{P_{\OV}}. 
        \end{equation}
        Furthermore, the set of removable vertices is nonempty.
    \end{lemma}
    
    Based on Lemma \ref{lemma: FindRemovable}, the pseudocode for \textbf{FindRemovable} function is identical to Algorithm \ref{alg: FindRemovable clique}, except that it gets the diamond-freeness as input instead of $m$ and it checks for \eqref{eq: removable-diamond} instead of \eqref{eq: removable-clique} in line 5.
    
    Similar to RSL$_\omega$, we have the following result. 
    
    \begin{proposition} \label{prp: findremovable}
        Suppose $\G[\OV]$ is a diamond-free DAG and a perfect map of $P_{\OV}$. \textbf{FindRemovable} returns a removable vertex in $\G[\OV]$ by performing at most $|\OV| \binom{\din{\G[\OV]}}{2}$ CI tests.
    \end{proposition}
    Analogous to the case with bounded clique number, the next result characterizes the co-parents of a removable vertex in a diamond-free graph.
    \begin{lemma} \label{lemma: FindNeighbors}
        Suppose $\G=(\OV,\E)$ is a diamond-free DAG and a perfect map of $P_{\OV}$. Let $X\in \OV$ be a removable vertex in $\G$, and $Y\in \Mb{X}{\OV}$. In this case, $Y\in \Cp{X}{\G}$ if and only if 
        \begin{equation}\label{eq: neighborhood-diamond}
            \exists Z\in \Mb{X}{\OV}\setminus \{Y\}\!:\ \  \CI{X}{Y}{\Mb{X}{\OV}\setminus\{Y,Z\}}{P_{\OV}}.
        \end{equation}
    \end{lemma}
    Accordingly, \textbf{FindNeighbors} is identical to Algorithm \ref{alg: FindNeighbors clique}, except that diamond-freeness is input to it rather than $m$ and it checks for \eqref{eq: neighborhood-diamond} instead of \eqref{eq: neighborhood-clique} in line 5.
    
    \begin{theorem}\label{thm: upperbound}
        Suppose $\G=(\V,\E)$ is a diamond-free DAG and a perfect map of $\PV$. RSL$_D$ is sound and complete, and performs $\mathcal{O}(|\V|^2\din{\G}^2)$ CI tests.
    \end{theorem}
    A limitation of RSL$_D$ is that diamond-freeness is not verifiable, unlike a bound on the clique number.
    However, even if the BN has diamonds, RSL$_D$ correctly recovers all the existing edges with possibly extra edges, i.e., RSL$_D$ has no false negative (see Appendix \ref{sec: apd_verify} for details.)
    Further, as we shall see in Section \ref{sec: experiments}, RSL$_D$ achieves the best accuracy in almost all cases in practice, even when the graph has diamonds. 

    
\section{Discussion}\label{sec: discussion}
\paragraph{Complexity analysis:}

Theorems \ref{thm: upperbound clique} and \ref{thm: upperbound} present the maximum number of CI tests required by Algorithm \ref{alg: RSL} to learn a DAG with bounded clique number and a diamond-free DAG, respectively. 
However, this algorithm may perform a CI test several times. 
We present an implementation of \textbf{RSL} in Appendix \ref{sec: apd_impl} that avoids such unnecessary duplicate tests (by keeping track of the performed CI tests, using mere logarithmic memory space) and achieves $\mathcal{O}(|\V|\din{\G}^3)$ and $\mathcal{O}(|\V|\din{\G}^{m+1})$ CI tests in diamond-free graphs and those with bounded clique number, respectively.
Recall that Algorithm \ref{alg: RSL} initially takes $\MB{\V}$ as an input, and
finding the Mbs requires an additional $\mathcal{O}(|\V|^2)$ number of CI tests.

Due to the recursive nature of \textbf{RSL}, the size of conditioning sets in each iteration reduces. 
Furthermore, since the size of the Mb of a removable variable is bounded by the maximum in-degree\footnote{See Lemma \ref{lemma:  Mb size of removable} in  Appendix \ref{sec: apd_proofs}.}, \textbf{RSL} performs CI tests with small conditioning sets. 
Having small conditioning sets in each CI test is essential to reduce sample complexity of the learning task.
In Section \ref{sec: experiments}, we empirically show that our proposed algorithms outperform the state-of-the-art algorithms both having lower number of CI tests and smaller conditioning sets. 

\paragraph{Random BNs:}
    As discussed earlier, diamond-free graphs or BNs with bounded clique numbers appear in some specific applications. Herein, we show that such structures also appear with high probability in networks whose edges appear independently and therefore, are essentially realizations of Erdos-Renyi graphs \cite{erdHos1960evolution}.
    \begin{lemma} \label{lemma: erdos-renyi}
        A random graph $\G$ generated from Erdos-Renyi model $G(n,p)$ is diamond-free with high probability when $pn^{0.8}\!\rightarrow\!0$ and $\omega(\G)\leq m$ when $p n^{2/m}\!\rightarrow\!0$.
    \end{lemma}

\section{Experiment}\label{sec: experiments}
    \begin{figure*}[ht] 
        \centering
        \captionsetup{justification=centering}
        \begin{subfigure}[b]{\textwidth}
            \centering
            \includegraphics[width=0.47\textwidth]{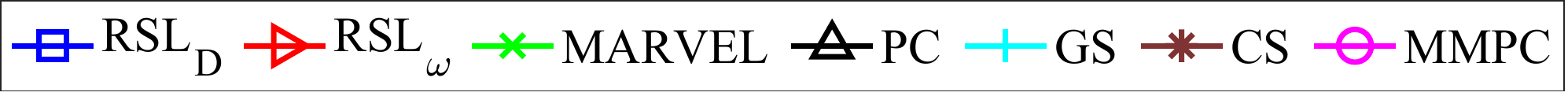}
        \end{subfigure}
        
        \begin{subfigure}[b]{0.23\textwidth}
            \centering
            \includegraphics[width=\textwidth]{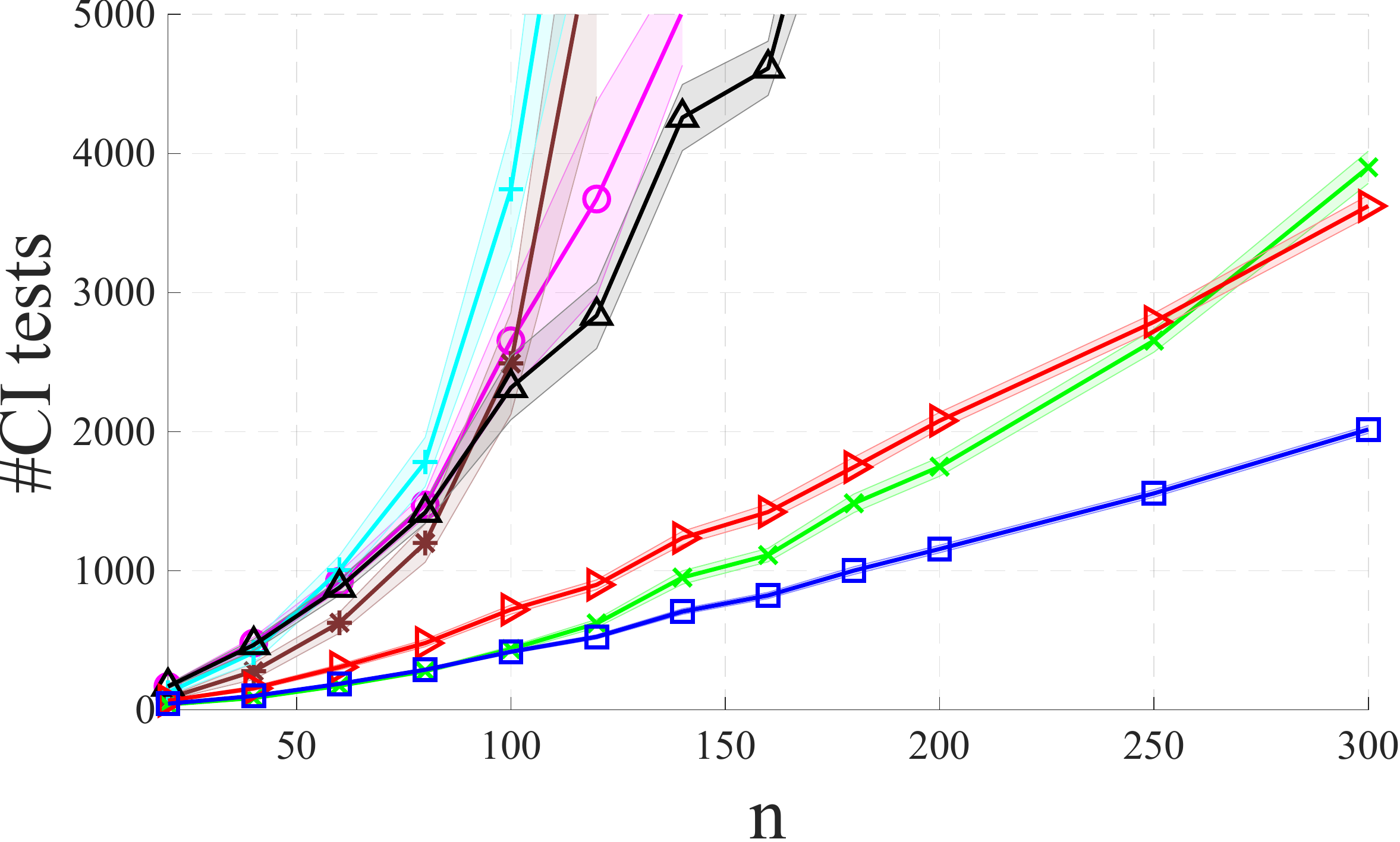}
            \caption{Oracle; $p=n^{-0.82}$.}
            \label{fig: oracle 0.82}
        \end{subfigure} \hspace{0.044\textwidth}
        \begin{subfigure}[b]{0.23\textwidth}
            \centering
            \includegraphics[width=\textwidth]{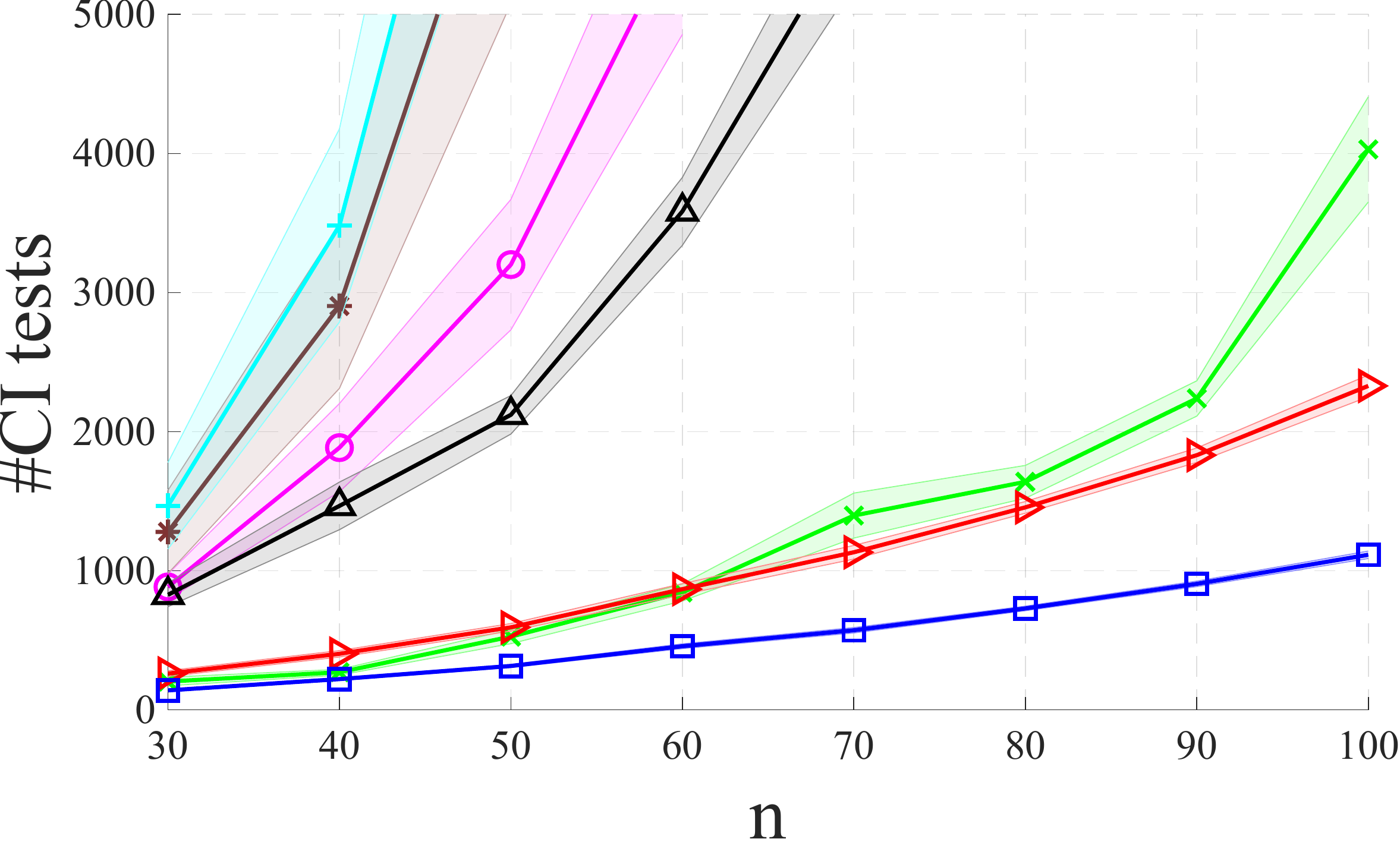}
            \caption{Oracle; $p= n^{-0.72}$.}
            \label{fig: oracle 0.72}
        \end{subfigure} \hspace{0.044\textwidth}
        \begin{subfigure}[b]{0.23\textwidth}
            \centering
            \includegraphics[width=\textwidth]{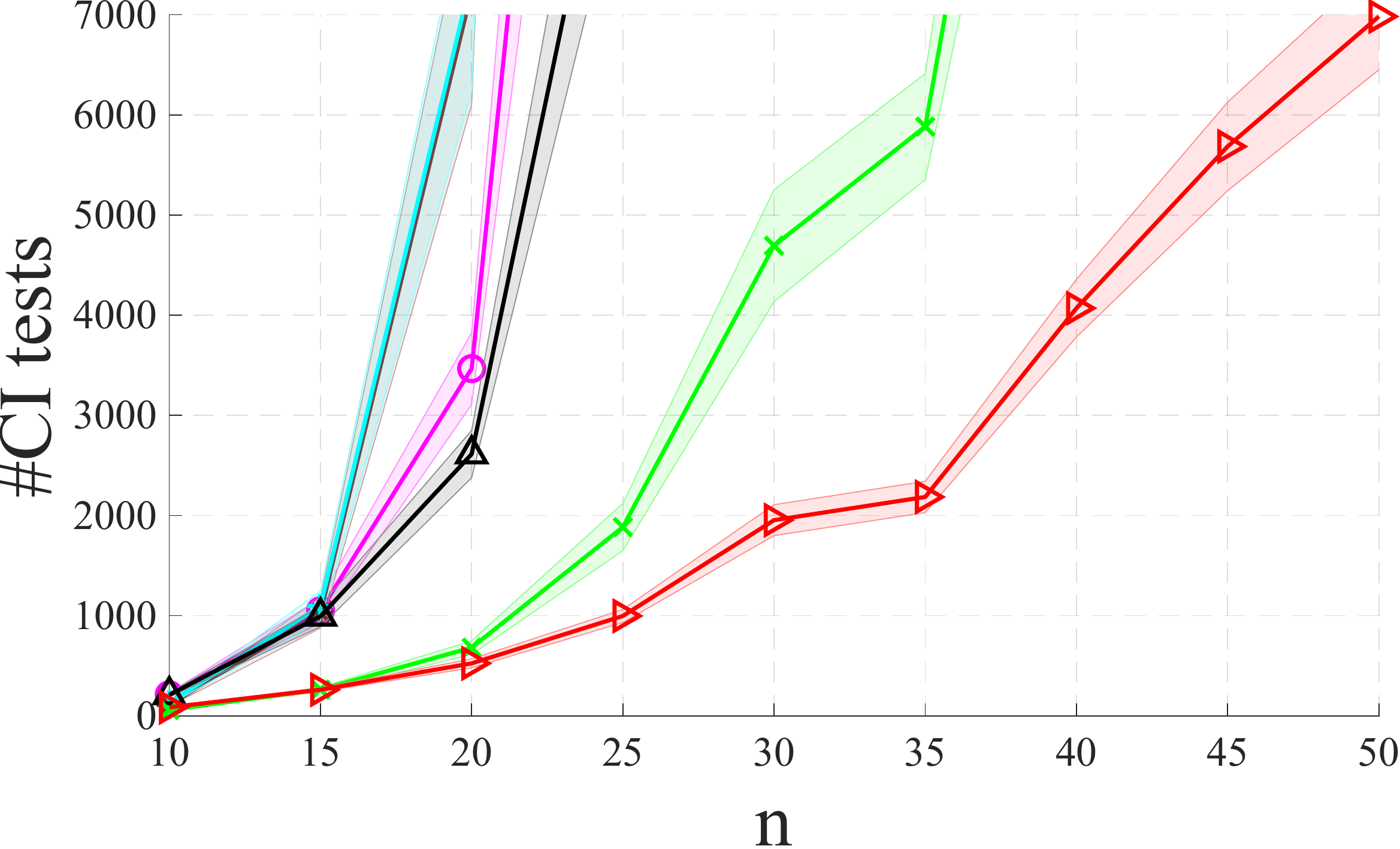}
            \caption{Oracle; $p=n^{-0.53}$.}
            \label{fig: oracle 0.53}
        \end{subfigure}
        
        \begin{subfigure}[b]{\textwidth}
            \centering
            \includegraphics[width=0.23\textwidth]{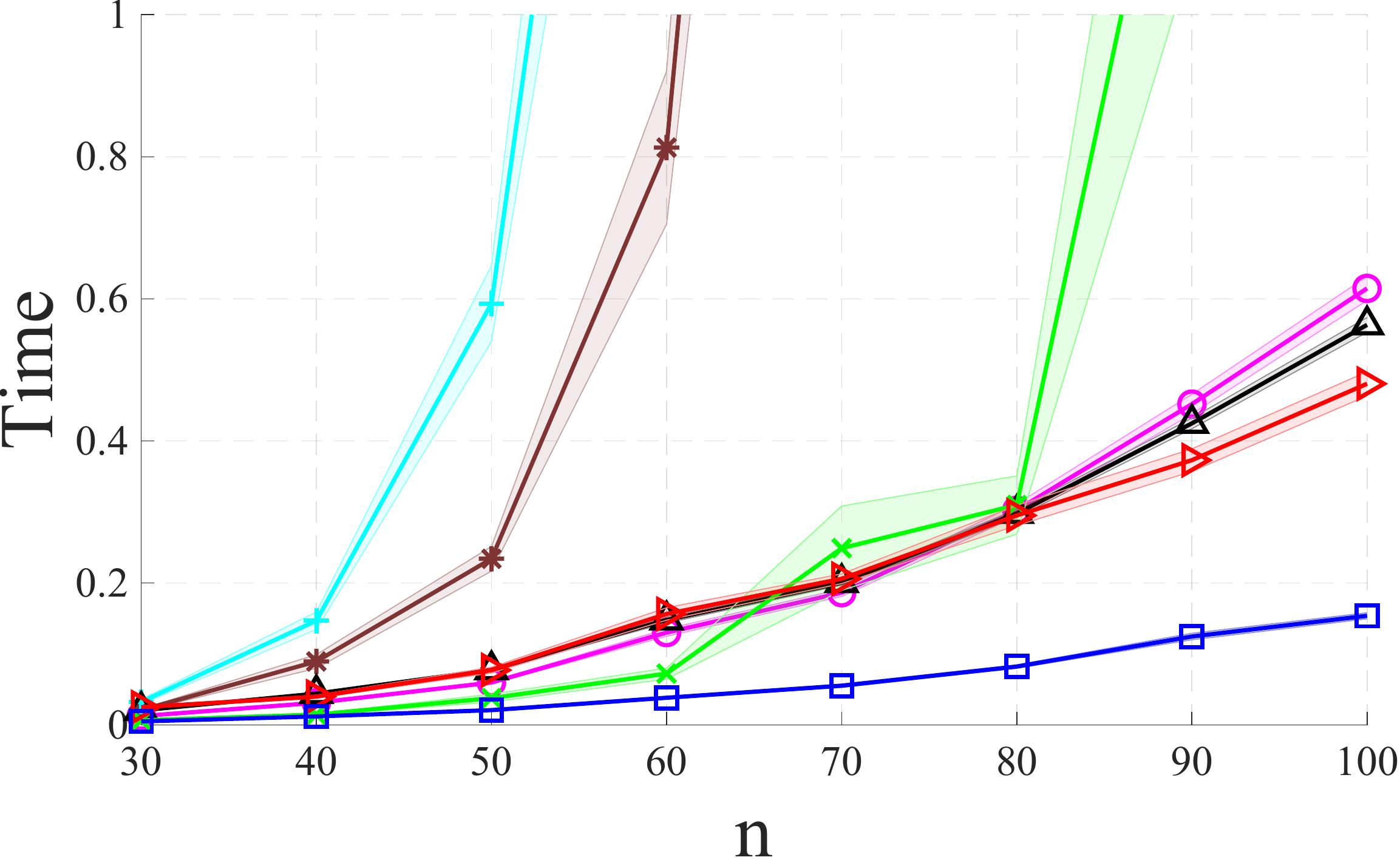}
            \hspace{0.05\textwidth}
            \includegraphics[width=0.23\textwidth]{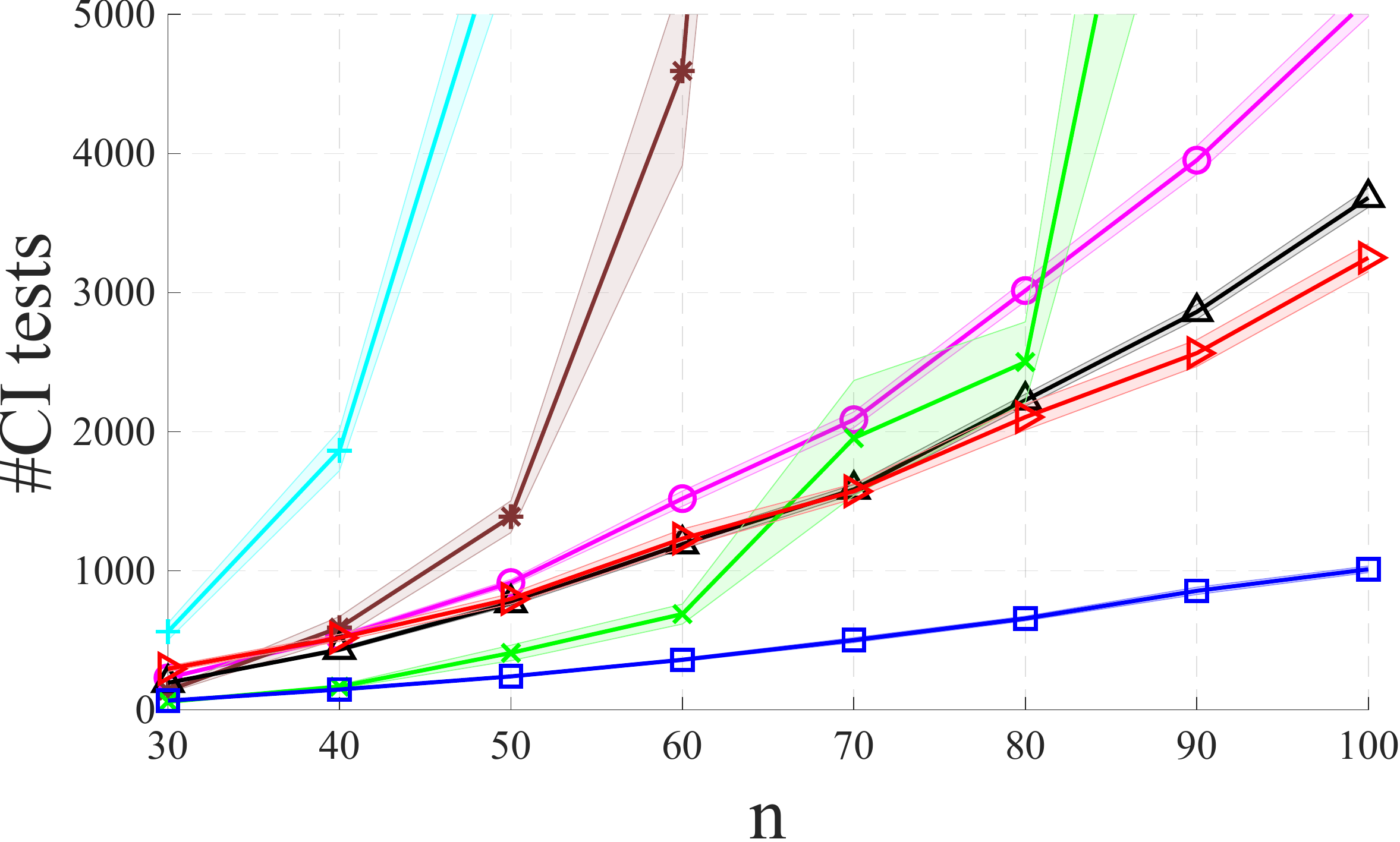}
            \hspace{0.05\textwidth}
            \includegraphics[width=0.23\textwidth]{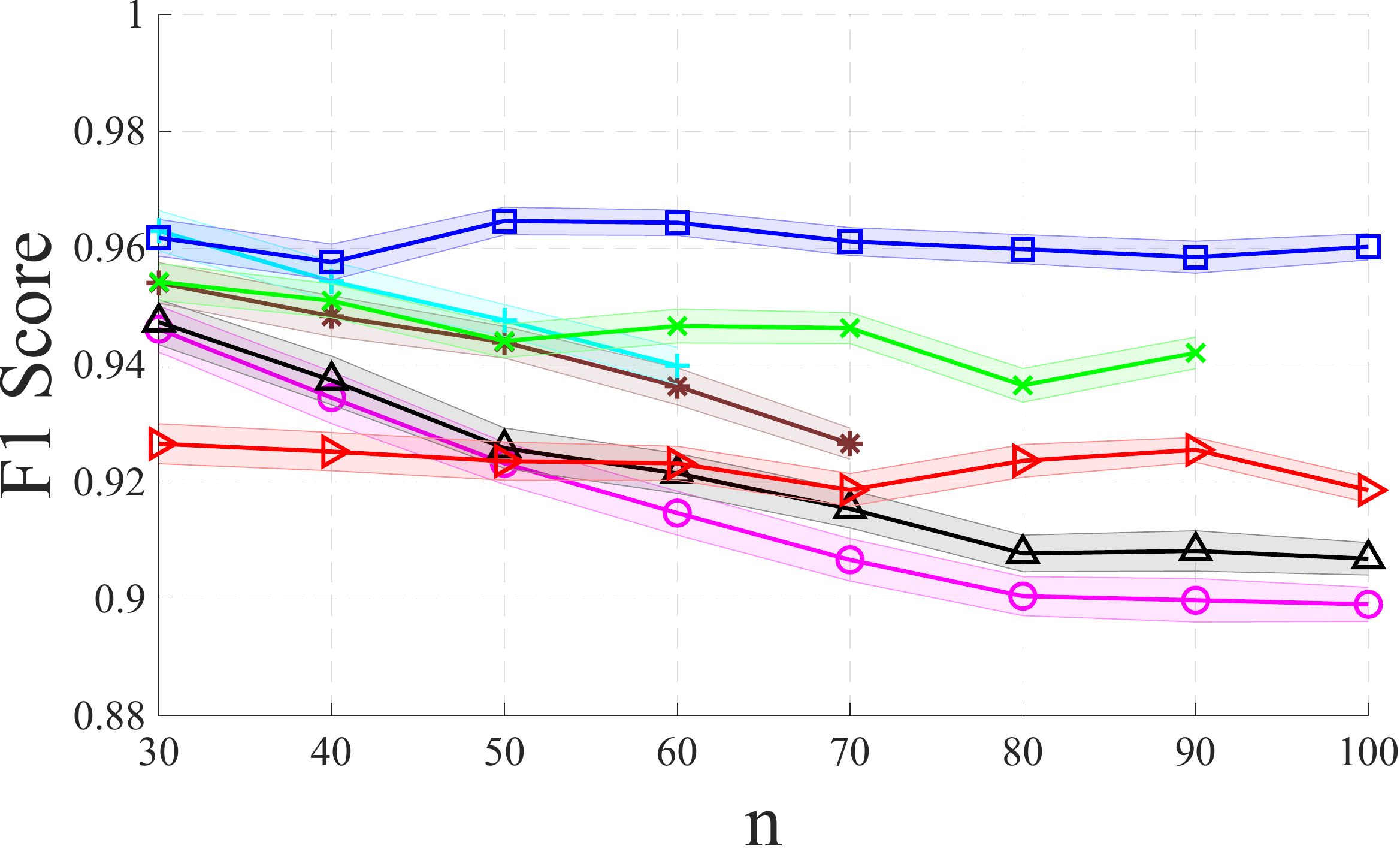}
            \caption{On data; $G(n,p)$ with $p= n^{-0.72}$ and sample size = $50n$.}
            \label{fig: data 0.72}
        \end{subfigure}
        
        \begin{subfigure}[b]{\textwidth}
            \centering
            \includegraphics[width=0.23\textwidth]{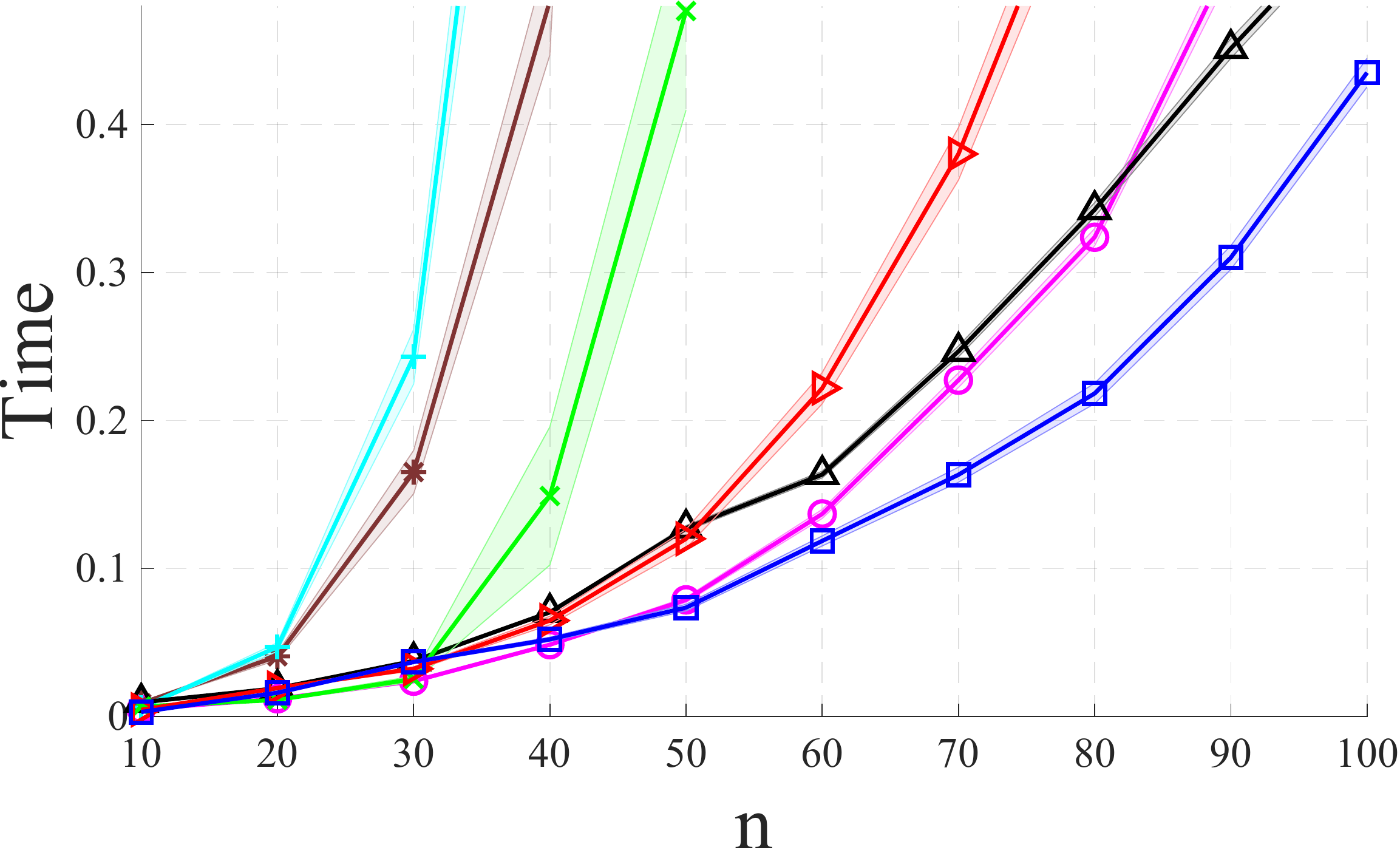}
            \hspace{0.05\textwidth}
            \includegraphics[width=0.23\textwidth]{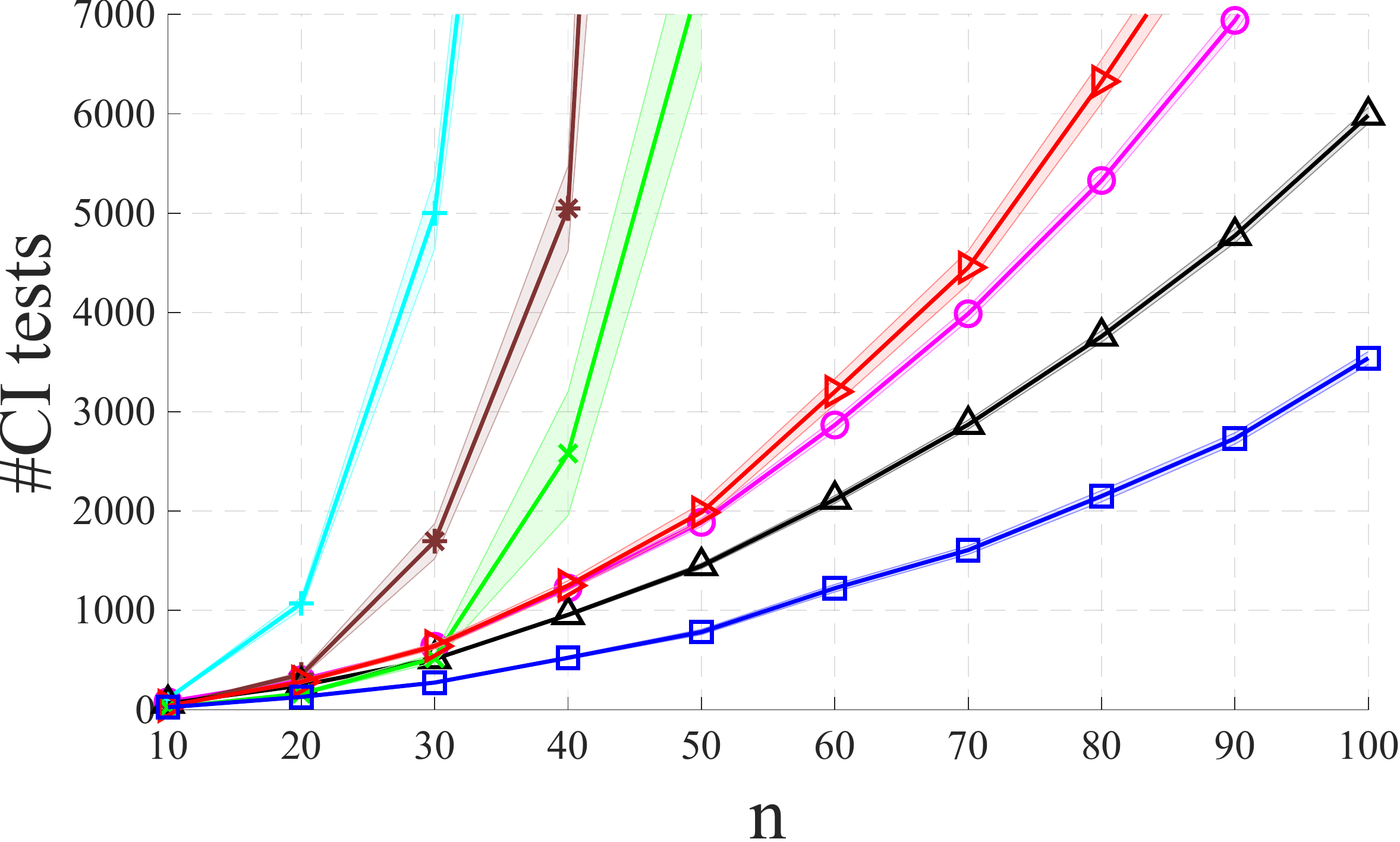}
            \hspace{0.05\textwidth}
            \includegraphics[width=0.23\textwidth]{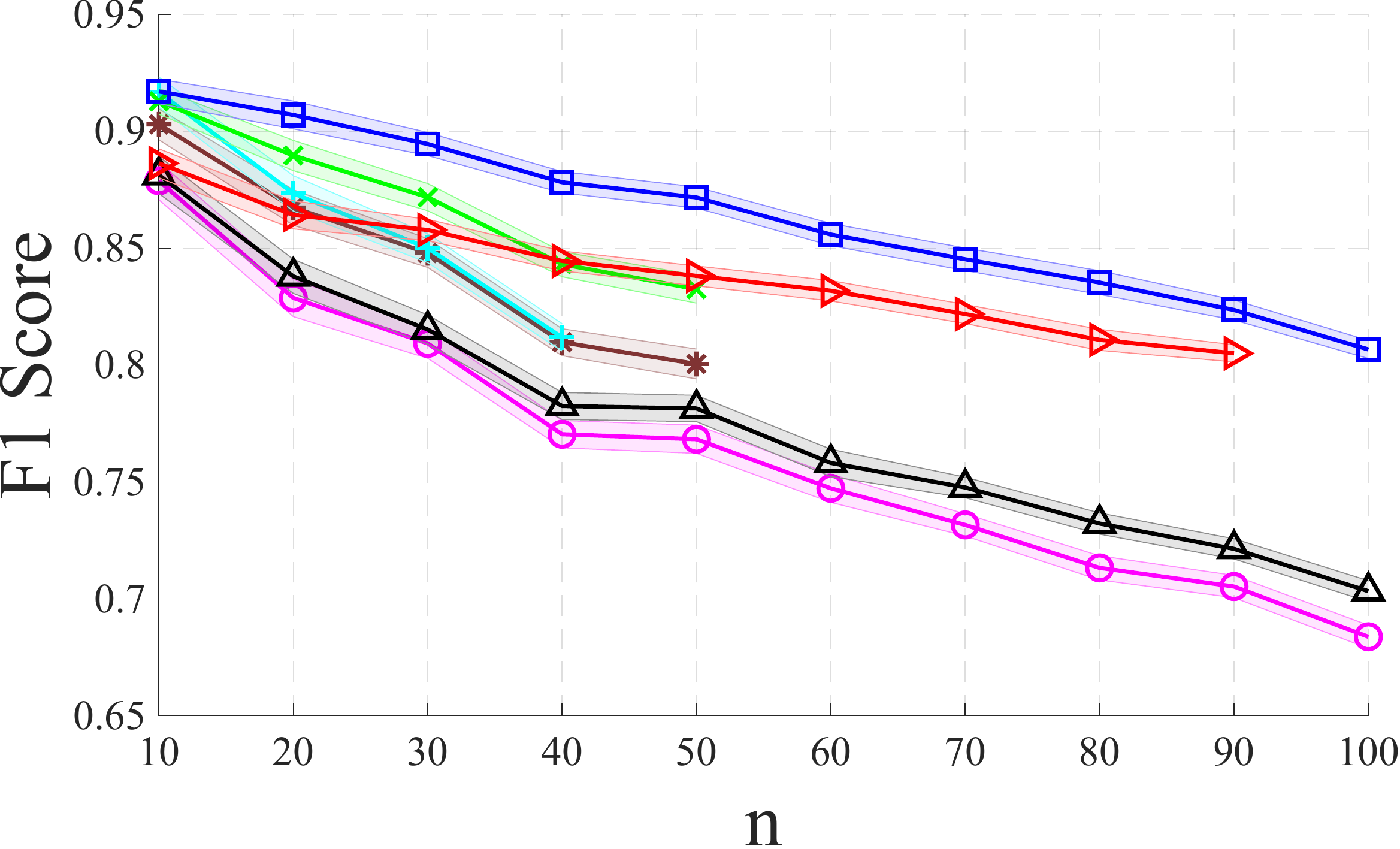}
            \caption{On data; $G(n,p)$ with $p= n^{-0.65}$ and sample size = $20n$.}
            \label{fig: data 0.65}
        \end{subfigure}
        \caption{Performance of various algorithms on random graphs generated from Erdos-Renyi models.}
        \label{fig: random graphs}
    \end{figure*}	

    \begin{figure*}[!ht] 
        \centering
        \captionsetup{justification=centering}
        \begin{subfigure}[b]{1\textwidth}
            \centering
            \includegraphics[width=0.23\textwidth]{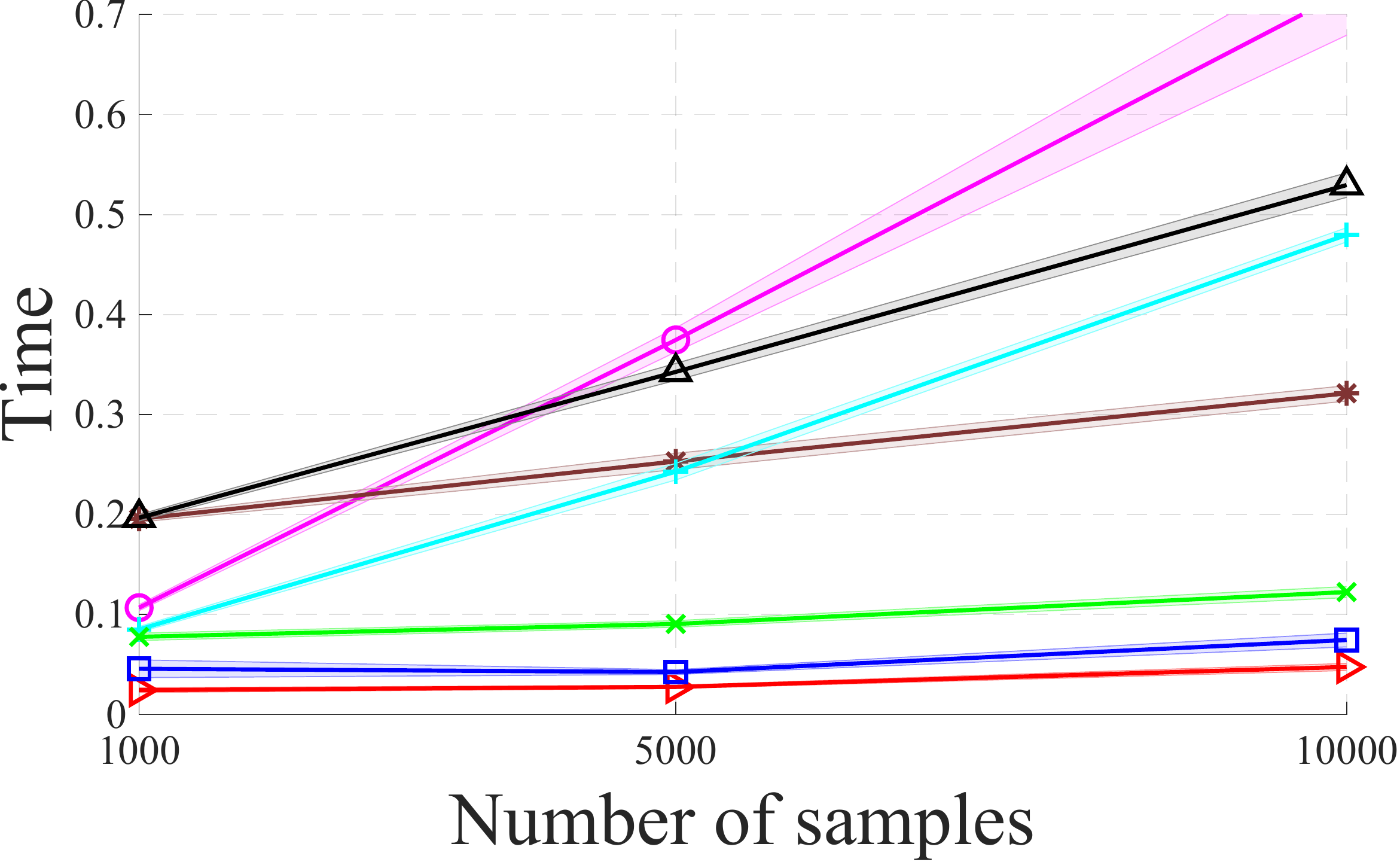}
            \hspace{0.05\textwidth}
            \includegraphics[width=0.23\textwidth]{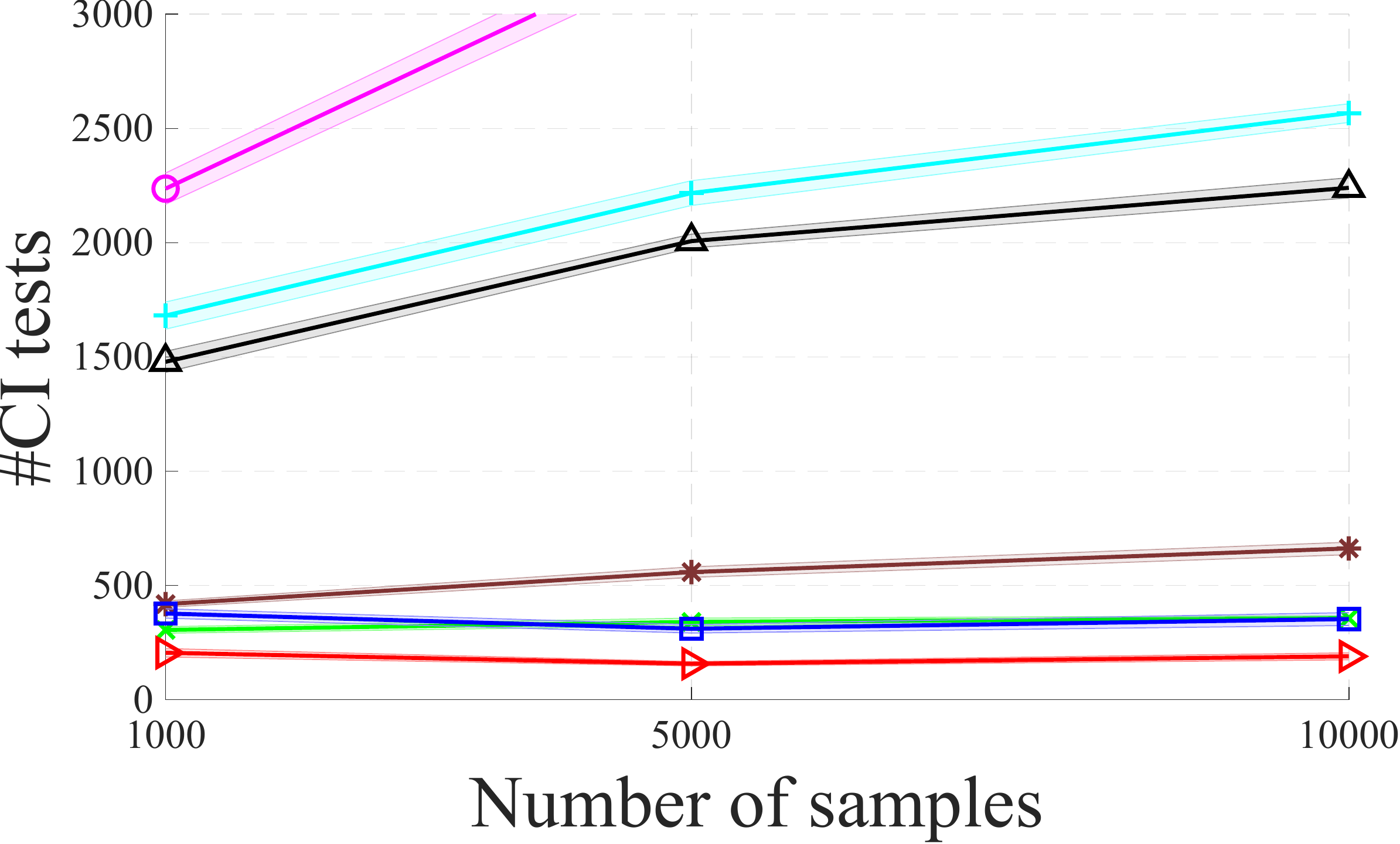}
            \hspace{0.05\textwidth}
            \includegraphics[width=0.23\textwidth]{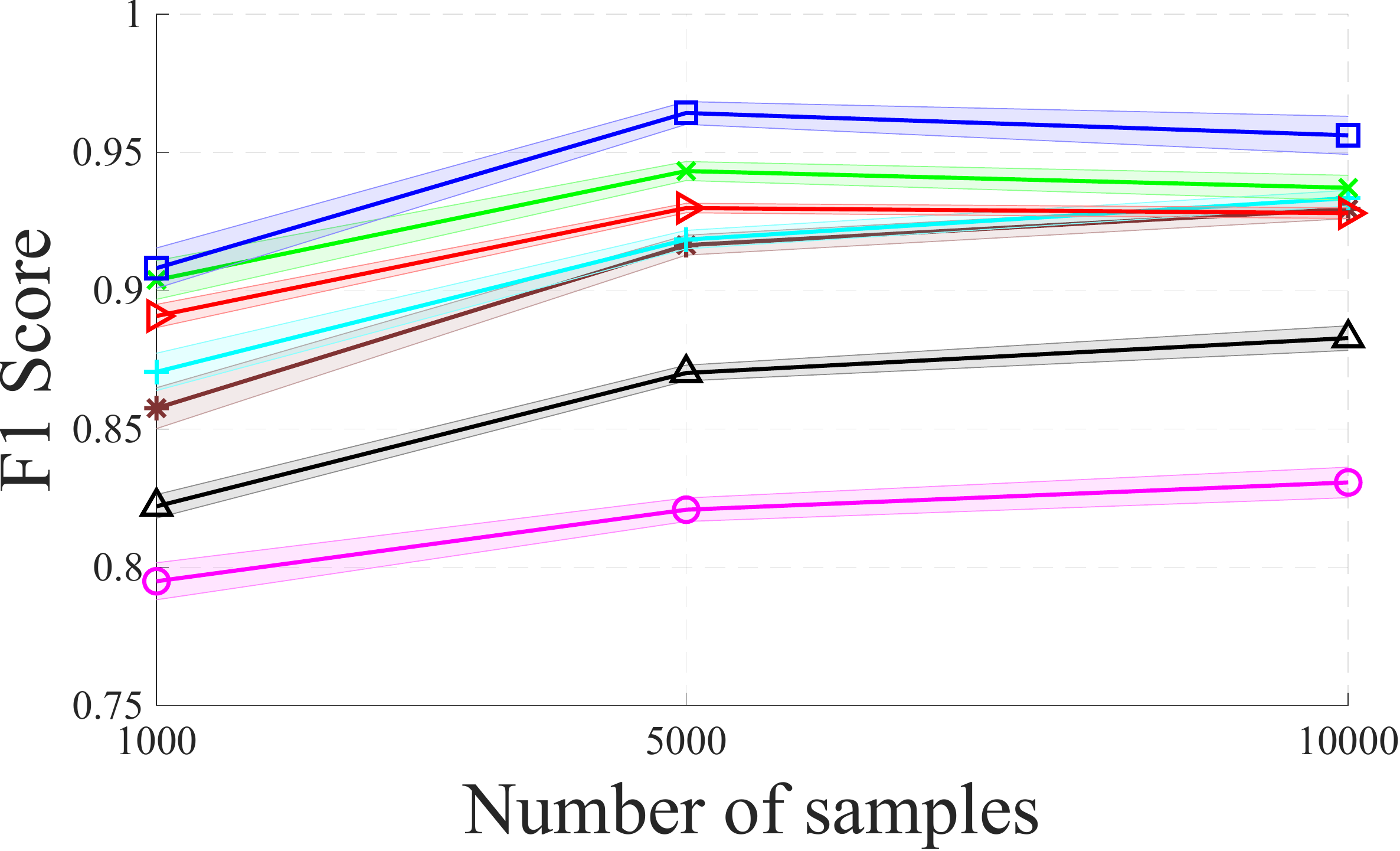}
            \caption{On data; Diabetes ($n=104, |\E|=148, \omega = 3, \Delta_{in}=2, \Delta=7, \alpha=12$, diamond-free).}
            \label{fig: Diabetes}
        \end{subfigure}
        
        \begin{subfigure}[b]{\textwidth}
            \centering
            \includegraphics[width=0.23\textwidth]{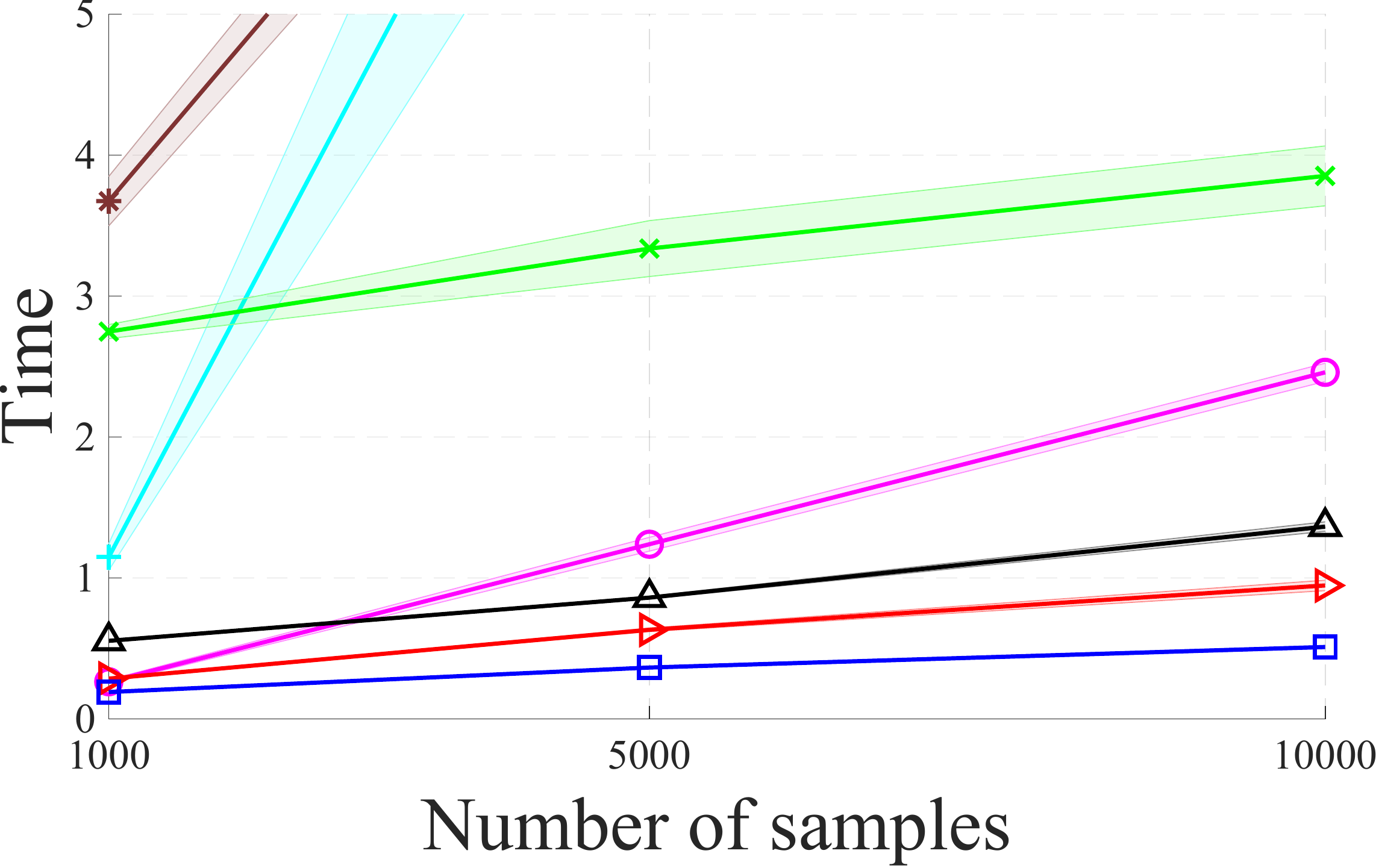}
            \hspace{0.05\textwidth}
            \includegraphics[width=0.23\textwidth]{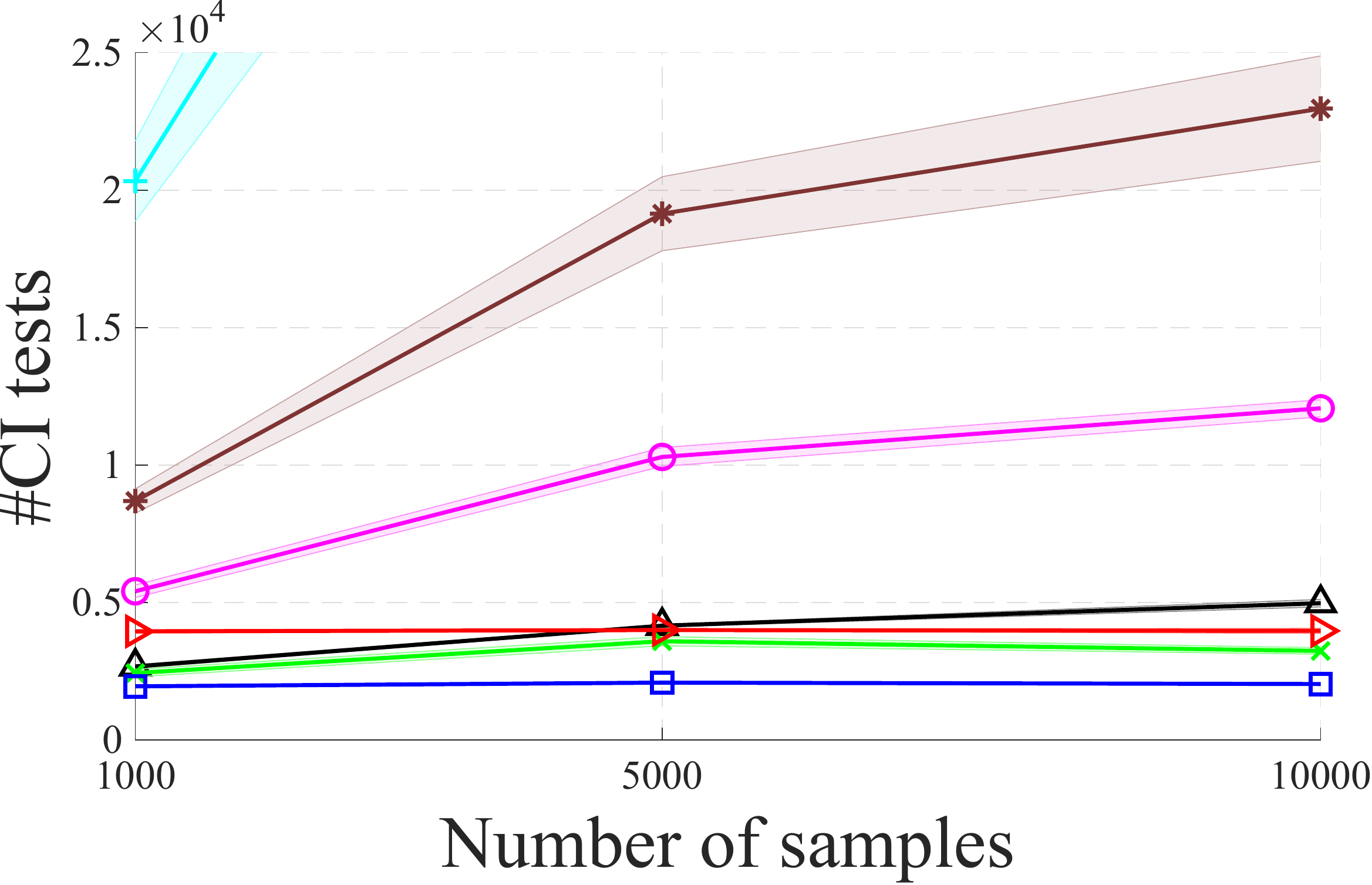}
            \hspace{0.05\textwidth}
             \includegraphics[width=0.23\textwidth]{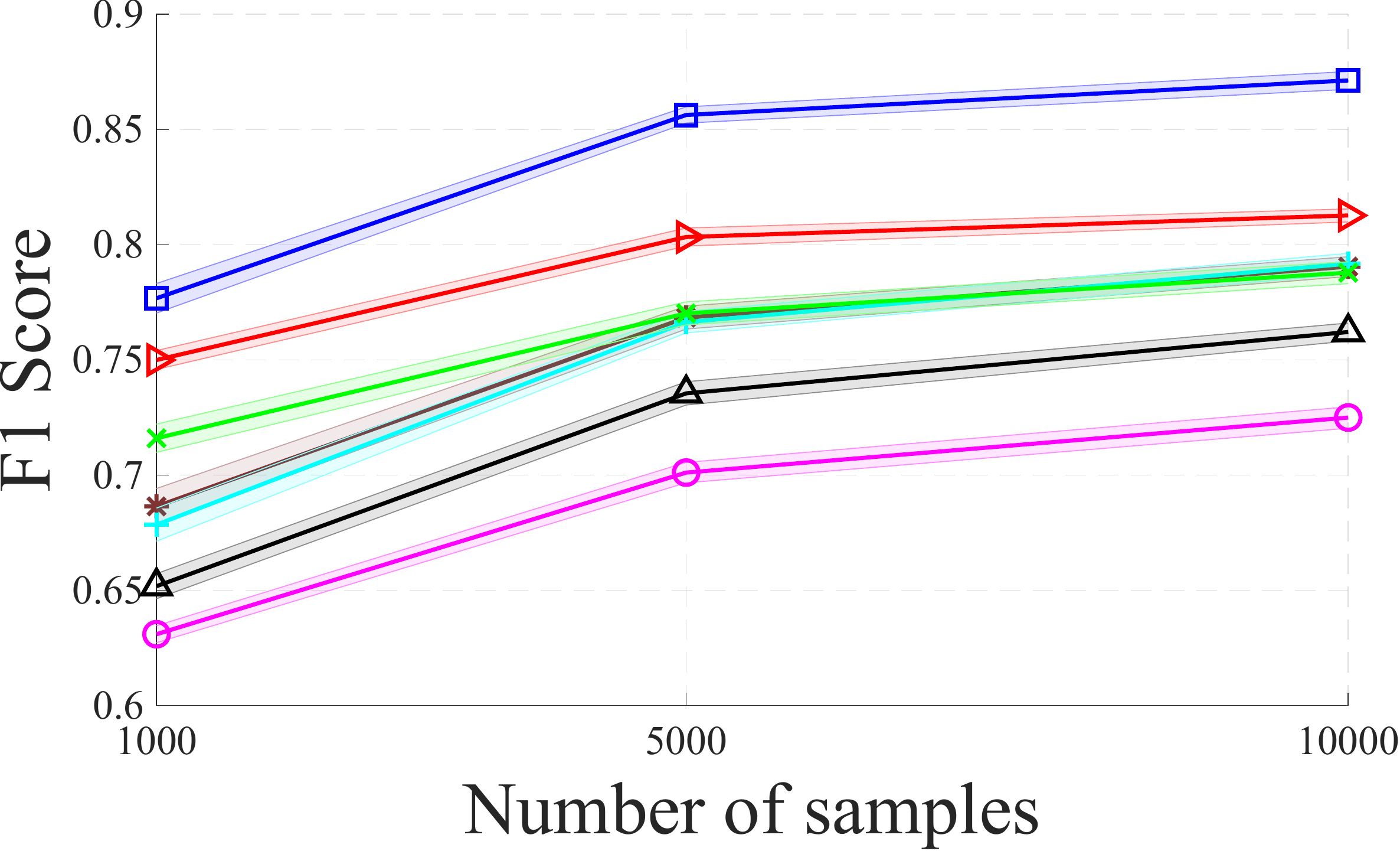}
            \caption{On data; Andes ($n=223, |\E|=328, \omega = 3, \Delta_{in}=6, \Delta=12, \alpha=23$, not diamond-free).}
            \label{fig: Andes}
        \end{subfigure}
        \caption{Performance of various algorithms on real-world structures.}
        \label{fig: real-world}
    \end{figure*}	    
    In this section, we present a set of experiments to illustrate the effectiveness of our proposed algorithms.
    The MATLAB implementation of our algorithms is publicly available\footnote{https://github.com/Ehsan-Mokhtarian/RSL.}. 
    We compare the performance of our algorithms, $RSL_{D}$ and RSL$_\omega$ with MARVEL \cite{mokhtarian2021recursive}, a modified version of PC \cite{spirtes2000causation, pellet2008using} that uses Mbs, GS \cite{margaritis1999bayesian}, CS \cite{pellet2008using}, and MMPC \cite{tsamardinos2003time} on both real-world structures and Erdos-Renyi random graphs.
    
    All aforementioned algorithms are Mb based.
    Thus, we initially use TC \cite{pellet2008using} algorithm to compute $\MB{\V}$, and then pass it to each of the methods for the sake of fair comparison.
    The algorithms are compared in two settings: I) oracle, and II) finite sample.
    In the oracle setting, we are working in the population level, i.e., the CI tests are queried through an oracle that has access to the true CI relations among the variables.
    In the latter setting, algorithms have access to a dataset of finite samples from the true distribution. Hence, the CI tests might be noisy. 
    The samples are generated using a linear model where each variable is a linear combination of its parents plus an exogenous noise variable; 
    the coefficients are chosen uniformly at random from $[-1.5,-1] \cup [1,1.5]$, and the noises are generated from $\mathcal{N}(0,\sigma^2)$, where $\sigma$ is selected uniformly at random from $[\sqrt{0.5}, \sqrt{1.5}]$.
    As for the CI tests, we use Fisher Z-transformation \cite{fisher1915frequency} with significance level 0.01 
    in the algorithms (alternative values did not alter our experimental results) and $\frac{2}{n^2}$ for Mb discovery \cite{pellet2008using}.
    These are standard evaluations' scenarios often performed in the structure learning literature \cite{colombo2014order,amendola2020structure, mokhtarian2021recursive, huang2012sparse, ghahramani2001propagation, scutari2019learning}. 
    We compare the algorithms in terms of runtime, the number of performed CI tests, and the f1-scores of the learned skeletons. 
    In Appendix \ref{sec: apd_rep}, we further report other measurements (average size of conditioning sets, precision, recall, structural hamming distance) of the learned skeletons, and accuracy of the learned separating sets.
    
    Figure \ref{fig: random graphs} illustrates the performance of BN learning algorithms on random Erdos-Renyi $G(n,p)$ model graphs.
    Each point is reported as the average of 100 runs, and the shaded areas indicate the $80\%$ confidence intervals.
    Runtime and the number of CI tests are reported after Mb discovery.
    Figures \ref{fig: oracle 0.82}, \ref{fig: oracle 0.72} and \ref{fig: oracle 0.53} demonstrate the number of CI tests each algorithm performed in the oracle setting, for the values of $p=n^{-0.82}$,$n^{-0.72}$, and $n^{-0.53}$, respectively.
    In \ref{fig: oracle 0.82}, the graphs are diamond-free with high probability (see Section \ref{sec: discussion} for details).
    In \ref{fig: data 0.72}, $\omega \leq 3$ with high probability, but the graphs are not necessarily diamond-free.
    In \ref{fig: oracle 0.53}, $\omega \leq 4$, with high probability.
    We have not included the result of RSL$_D$ in Figure \ref{fig: oracle 0.53}, as the graphs contain diamonds with high probability, and RSL$_D$ has no theoretical guarantee despite of low complexity. 
    Figures \ref{fig: data 0.72} and \ref{fig: data 0.65} demonstrate the performance of the algorithms in the finite sample setting, when $50n$ and $20n$ samples were available, respectively.
    Although RSL$_D$ does not have any theoretical correctness guarantee to recover the network (graphs are not diamond-free), both RSL$_D$ and RSL$_\omega$ outperform  other algorithms in terms of both accuracy and computational complexity in most cases.
    The lower runtime of MARVEL and MMPC compared to RSL$_\omega$ in Figure \ref{fig: data 0.65} can be explained through their significantly low accuracy due to skipping numerous CI tests.
    
    Figure \ref{fig: real-world} illustrates the performance of BN learning algorithms on two real-world structures, namely Diabetes \cite{andreassen1991model} and Andes \cite{conati1997line} networks,
    over a range of different sample sizes.
    Each point is reported as the average of 10 runs.
    As seen in Figures \ref{fig: Diabetes} and \ref{fig: Andes}, both RSL algorithms outperform other algorithms in both accuracy and complexity.
    Note that although Andes is not a diamond-free graph, RSL$_D$ achieves the best accuracy in Figure \ref{fig: Andes}.    
    Similar experimental results for five real-world structures in both oracle and finite sample settings along with detailed information about these structures appear in Appendix \ref{sec: apd_rep}.

\section{Conclusion}
    In this work, we presented the RSL algorithm for BN structure learning. 
    Although our generic algorithm has exponential complexity, we showed that it could harness structural side information to learn the BN structure in polynomial time.
    In particular, we considered two types of side information about the underlying BN: I) when an upper bound on its clique number is known, and II) when the BN is diamond-free.
    We provided theoretical guarantees and upper bounds on the number of CI tests required by our algorithms. 
    We demonstrated the superiority of our proposed algorithms in both synthetic and real-world structures.
    We showed in the experiments that even when the graph is not diamond-free, RSL$_D$ outperforms various algorithms both in time complexity and accuracy. 
\section*{Acknowledgments}
    The work presented in this paper was in part supported by Swiss National Science Foundation (SNSF) under grant number 200021\_20435.
\bibliography{bibliography}
\clearpage
\onecolumn
\appendix

\section*{Appendix}
Table \ref{tab: appendix} summarizes how the Appendix is organized.
\begin{table*}[ht]
        \centering
        \begin{tabular}{N M{1cm}|p{6cm} }
            \toprule
            & Section &  Title\\
            \hline
            & \ref{sec: apd-dsep}  & d-separation   \\
            & \ref{sec: apd_proofs} & Technical proofs  \\
            & \ref{sec: apd_verify}   & RSL$_D$ on networks with Diamonds  \\
            & \ref{sec: apd_alg_updat}      & Discussion on Algorithm \ref{alg: update Mb} \\
            & \ref{sec: apd_impl}      & Discussion on the implementation of \textbf{RSL}  \\
           & \ref{sec: apd_rep}        &  Reproducibility and additional experiments \\
            \bottomrule
        \end{tabular}
        \caption{Organization of the Appendix.}
        \label{tab: appendix}
    \end{table*}
\section{d-separation}\label{sec: apd-dsep}
    \begin{definition}[Path \& Directed Path]
        Let $X_1,...,X_m$ be a set of distinct vertices in a DAG $\G=(\V,\E)$ such that $X_i\in \N{X_{i+1}}{\G}$. This is called a path between $X_1$ and $X_m$. 
        When $X_i\in \Pa{X_{i+1}}{\G}$ for $1\leq i\leq m-1$, then we have a directed path from $X_1$ to $X_m$.
    \end{definition}
    
    \begin{definition}[$\De{X}{\G}$] \label{descendent}
        Suppose $X, Y \in \V$ in a DAG $\G=(\V,\E)$. $Y$ is called a descendent of $X$ if there exists a directed path from $X$ to $Y$. We show by $\De{X}{\G}$ the set of all descendants of $X$. 
    \end{definition}
    
    \begin{definition}[Blocked] \label{blocked}
        A path between $X_1$ and $X_m$ is called blocked by a set $\textbf{S}$ (with neither $X_1$ nor $X_n$ in $\textbf{S}$) whenever there is a node $X_k$, such that one of the following two possibilities holds:
        \begin{enumerate}
            \item $X_k\in \textbf{S}$ and $X_{k-1}\rightarrow X_k\rightarrow X_{k+1}$ or $X_{k-1}\leftarrow X_k\leftarrow X_{k+1}$ or $X_{k-1}\leftarrow X_k\rightarrow X_{k+1}$.
            \item $X_{k-1}\rightarrow X_k\leftarrow X_{k+1}$ and neither $X_k$ nor any of its descendants is in $\textbf{S}$.
        \end{enumerate}
    \end{definition}
    
    \begin{definition}[d-separation]
        Let $\textbf{X}, \textbf{Y}$, and $\mathbf{S}$ be three disjoint subsets of vertices of a DAG $\G$. We say $\textbf{X}$ and $\textbf{Y}$ are d-separated by $\textbf{S}$, denoted by $\dsep{\textbf{X}}{\textbf{Y}}{\mathbf{S}}{\G}$, if every path between vertices in $\textbf{X}$ and $\textbf{Y}$ is blocked by $\textbf{S}$.
    \end{definition}

\section{Technical proofs}\label{sec: apd_proofs}

To present the proofs, we need the following prerequisites.
    

    \begin{definition}[Collider] \label{collider}
        Let $X_1,...,X_m$ be a path between $X_1$ and $X_m$ in a DAG $\G=(\V,\E)$. A vertex $X_j$, $1<j<m$ on this path is called a collider if both $X_{j}\in\Ch{X_{j+1}}{\G}$ and $X_{j}\in\Ch{X_{j-1}}{\G}$. i.e.,  $X_{j-1}\rightarrow X_j\leftarrow X_{j+1}$.
    \end{definition}
    
    \begin{theorem}[\cite{mokhtarian2021recursive}] \label{theorem: removability}
    	$X$ is removable in a DAG $\G$ if and only if the following two conditions are satisfied for every $W\in\Ch{X}{\G}.$
		\begin{description}
			\item 
			    Condition 1: $\N{X}{\G}\subseteq \N{W}{\G}\cup\{W\}.$
			\item 
			    Condition 2: $\Pa{Y}{\G} \subseteq \Pa{W}{\G}$ for any $Y\in\Ch{X}{\G}\cap \Pa{W}{\G}$.
		\end{description}
    \end{theorem}
    
    \begin{lemma}[\cite{mokhtarian2021recursive}] \label{lemma:  Mb size of removable}
        Suppose $\G=(\V,\E)$ is a perfect map of $P_{\V}$ and $X\in \V$ is a removable vertex in $\G$. In this case, $|\Mb{X}{\V}|\leq  \din{\G}$. 
    \end{lemma}

    \subsection{Proofs of Section \ref{sec: RSL}:}
    \begin{customprp}{\ref{prp: removable}}\label{prp: removable proof}
        Suppose $\G=(\V,\E)$ is a perfect map of $\PV$. For each variable $X\in \V$, $\G[\V \setminus \{X\}]$ is a perfect map of $P_{\V\setminus\{X\}}$ if and only if $X$ is a removable vertex in $\G$. 
    \end{customprp}
    \begin{myproof}[Proof]
        \emph{If part:}
        We need to show that $\G[\V \setminus \{X\}]$ is an I-map and a D-map of $P_{\V \setminus \{X\}}$:
        \begin{itemize}[leftmargin=*]
            \item I-map: 
            Suppose $Y,Z \in \V \setminus \{X\}$ and $\mathbf{S}\subseteq \V \setminus \{X,Y,Z\}$ such that $\dsep{Y}{Z}{\mathbf{S}}{\G[\V \setminus \{X\}]}$. Equation \eqref{eq: d-sepEquivalence} implies 
            $\dsep{Y}{Z}{\mathbf{S}}{\G}$. Since $\G$ is an I-map of $\PV$, $Y$ and $Z$ are independent conditioned on $\mathbf{S}$ in $\PV$ and therefore in $P_{\V\setminus\{X\}}$. Hence, $\G[\V \setminus \{X\}]$ is an I-map of $P_{\V\setminus\{X\}}$.
            
            \item D-map: 
            Suppose $Y,Z \in \V \setminus \{X\}$ and $\mathbf{S}\subseteq \V \setminus \{X,Y,Z\}$ such that $\notdsep{Y}{Z}{\mathbf{S}}{\G[\V \setminus \{X\}]}$.
            Equation \eqref{eq: d-sepEquivalence} implies 
            $\notdsep{Y}{Z}{\mathbf{S}}{\G}$.
            Since $\G$ is a D-map of $\PV$, $Y$ and $Z$ are dependent conditioned on $\mathbf{S}$ in $\PV$ and therefore in $P_{\V\setminus\{X\}}$. Hence, $\G[\V \setminus \{X\}]$ is a D-map of $P_{\V\setminus\{X\}}$.
        \end{itemize}
        \emph{Only if part:}
            We need to prove that $X$ is removable in $\G$. Suppose $Y,Z\in \V \setminus \{X\}$ and $\mathbf{S}\subseteq \V \setminus \{X,Y,Z\}$.
            Since $\G[\V \setminus \{X\}]$ is a perfect map of $P_{\V\setminus \{X\}}$, 
            \[ 
                \dsep{Y}{Z}{\mathbf{S}}{\G[\V\setminus \{X\}]} \iff \CI{Y}{Z}{\mathbf{S}}{P_{\V\setminus\{X\}}}\iff \CI{Y}{Z}{\mathbf{S}}{P_{\V}}.
            \]
            Since $\G$ is a perfect map of $P_{\V}$, we obtain 
            \[ 
                 \CI{Y}{Z}{\mathbf{S}}{P_{\V}}  \iff \dsep{Y}{Z}{\mathbf{S}}{\G[\V]},
            \]
            and consequently,
            \[
             \dsep{Y}{Z}{\mathbf{S}}{\G[\V\setminus \{X\}]} \iff \dsep{Y}{Z}{\mathbf{S}}{\G[\V]}. 
            \]
            Therefore, due to the definition of removability (Definition \ref{def: removable}), $X$ is a removable vertex in $\G$. 
    \end{myproof}
    
    \begin{customprp}{\ref{prp: updatemb}}\label{prp: updatemb proof}
        Suppose $\mathcal{G}[\OV]$ is a perfect map of $P_{\OV}$ and $X$ is a removable variable in $\mathcal{G}[\OV]$. Algorithm \ref{alg: update Mb} correctly finds $\MB{\OV \setminus \{X\}}$ by performing at most $\binom{|\N{X}{\G[\OV]}|}{2}$ CI tests. 
	\end{customprp}
	\begin{myproof}[Proof]	   
	\emph{Soundness:}
	   Suppose $X$ is a removable vertex in $\G[\OV]$ with the set of neighbors $\N{X}{\G[\OV]}$. Since $\G[\OV]$ is a perfect map of $P_{\OV}$, Proposition {\ref{prp: removable proof}} implies that $\G[\OV \setminus \{X\}]$ is a perfect map of $P_{{\OV} \setminus\{X\}}$. Hence, for any $Y\in \OV$,
	   \[\Mb{Y}{\OV} = \N{Y}{\G[\OV]} \cup \Cp{Y}{\G[\OV]},\]
	   and for any $Y \in \OV \setminus \{X\}$,
	   \[\Mb{Y}{\OV\setminus \{X\}} = \N{Y}{\G[\OV\setminus \{X\}]} \cup \Cp{Y}{\G[\OV\setminus \{X\}]}.\]
	   First, note that for any $Y\in \OV \setminus \{X\}$,
	   \[\N{Y}{\G[\OV\setminus \{X\}]} = \N{Y}{\G[\OV]} \setminus \{X\},\]
	   and 
	   \[\Cp{Y}{\G[\OV\setminus \{X\}]} \subseteq  \Cp{Y}{\G[\OV]}.\]
	   Algorithm \ref{alg: update Mb} initializes $\MB{\OV \setminus\{X\}}$ with $\MB{\OV}$ in line 2, and remove $X$ from them in line 4. Hence, we only need to identify the variables in $D(Y):= \Cp{Y}{\G[\OV]} \setminus \Cp{Y}{\G[\OV\setminus \{X\}]}$ and remove them from $\Mb{Y}{\OV \setminus \{X\}}$.
	   
	   Suppose $Y \in \OV \setminus \{X\}$ such that $D(Y)\neq \varnothing$. 
	   Let $Z$ be a vertex in $D(Y)$. 
	   In this case, $Y$ and $Z$ have a common child in $\G[\OV]$, but do not have any common child in $\G[\OV\setminus \{X\}]$. 
	   Hence, 
	   $X$ is the only common child of $Y$ and $Z$ in $\G[\OV]$.
	   However, this cannot happen if $X$ has a child in $\G[\OV]$, because if $W\in \Ch{X}{\G[\OV]}$, Condition 1 of Theorem \ref{theorem: removability} implies that $Y,Z \in \Pa{W}{\G[\OV]}$, i.e., $W$ is a common child for $Y,Z$ which is not possible. 
	   Hence, 
	   \begin{equation} \label{eq: if X has child}
	       \Ch{X}{\G[\OV]}\neq \varnothing 
	       \Longrightarrow 
	       D(Y)=\varnothing,\,\hspace{5pt}  \forall Y \in \OV \setminus \{X\}. 
	   \end{equation}
	   In line 5 of Algorithm \ref{alg: update Mb}, if the condition $\N{X}{\G[\OV]}=\Mb{X}{\OV}$ does not hold, then $X$ has at least one co-parent, and therefore, at least one child. 
	   In this case, 
	   Equation \eqref{eq: if X has child} implies that Algorithm \ref{alg: update Mb} correctly returns $\MB{\OV\setminus \{X\}}$.
	   
	   As we mentioned above, if $Z\in D(Y)$, then $Y,Z \in \Pa{X}{\G[\OV]}\subseteq \N{X}{\G[\OV]}$. 
	   In the next step (lines 6-9), similar to the methods presented in \cite{mokhtarian2021recursive} and \cite{margaritis1999bayesian}, for all pairs $Y,Z \in \N{X}{\G[\OV]}$ we can either perform $\CI{Y}{Z}{\Mb{Y}{\OV \setminus \{X\}}\setminus \{Y,Z\}}{P_{\OV}}$ or $\CI{Y}{Z}{\Mb{Z}{\OV \setminus \{X\}}\setminus \{Y,Z\}}{P_{\OV}}$ to decide whether they remain in each other's Markov boundary after removing $X$ from $\G[\OV]$.
	   
	   \emph{Complexity:}
	   The algorithm possibly performs CI tests only in line 7 for a pair of distinct variables in $\N{X}{\G[\OV]}$. 
	   Hence, it perform at most $\binom{|\N{X}{\G[\OV]}|}{2}$ CI tests.
	\end{myproof}
 \subsection{Proofs of Section \ref{sec: generalization}:}\label{sec: apd_bounded}

    We first present Lemma \ref{lem: rem-clique-k} that will be used for the proofs of this section.
    \begin{lemma} \label{lem: rem-clique-k}
        Suppose $\G=(\OV,\E)$ is a DAG and a perfect map of $P_{\OV}$, $X\in \OV$, and $1 \leq k \leq \vert\Ch{X}{\mathcal{G}}\vert$ is a fixed number.
        If Equation \eqref{eq: removable-clique} (or \eqref{eq: removable-clique proof}) holds for all $\mathbf{S}\subseteq\Mb{X}{\OV}$ with $\left\vert\mathbf{S}\right\vert \leq k-1$, then there exists $\mathbf{W}=\{W_1,\dots,W_k\} \subseteq \Ch{X}{\G}$ such that the following hold.
        \begin{enumerate}
            \item $W_j \in \Pa{W_i}{\G}$, for all $1\leq i<j\leq k$.
            \item $Y\in \Pa{W_i}{\G}$, for all $1\leq i \leq k$ and $Y\in \Mb{X}{\OV}\setminus \mathbf{W}$.
        \end{enumerate}
    \end{lemma}
    \begin{proof}
        We construct $\mathbf{W}$ iteratively, and for choosing $W_1$ we use Lemma \ref{lemma: limb1} and Equation \eqref{eq: proof diamond-free} that are presented in Appendix \ref{sec: diamond-free proofs}.
        First note that for $|\mathbf{S}|=0$, Equation \eqref{eq: removable-clique proof} reduces to Equation \eqref{eq: proof diamond-free}. 
        Thus, Lemma \ref{lemma: limb1} implies that there exists $W_1\in \Ch{X}{\G}$ such that $\Mb{X}{\OV}\cup \{X\} = \Pa{W_1}{\G}\cup \{W_1\}$.
        Hence, $\mathbf{W}=\{W_1\}$ satisfies the following conditions:
        \begin{equation}\label{eq: lem-W-k}
            \begin{aligned}
                &W_j \in \Pa{W_i}{\G}, \forall\: 1\leq i<j\leq |\mathbf{W}|,\\
                &Y\in \Pa{W_i}{\G}, \forall\: 1\leq i \leq |\mathbf{W}|\: \&\: Y\in \Mb{X}{\OV}\setminus \mathbf{W}.
            \end{aligned}
        \end{equation}
        Now, suppose there exists $\mathbf{W}=\{W_1,\dots, W_m\}\subseteq \Ch{X}{\G}$ that satisfies Equation \eqref{eq: lem-W-k}.
        In order to prove the lemma, it suffices to show that there exists a set $\mathbf{W^{'}}$ with $|\mathbf{W^{'}}|=m+1$ that satisfies Equation \eqref{eq: lem-W-k}. To this end, we introduce $W_{m+1}\in \Ch{X}{\G}\setminus \mathbf{W}$ such that $\mathbf{W^{'}}=\mathbf{W}\cup \{W_{m+1}\}$ satisfies Equation \eqref{eq: lem-W-k}.
        
        Let $W_{m+1}$ be a variable in $\mathbf{A}:=\Mb{X}{\OV}\setminus \mathbf{W}$ that does not have any descendants in $\mathbf{A}$. 
        Note that $\mathbf{A}$ is non-empty since $m<|\Ch{X}{\G}|$, and such a vertex exists due to the assumption that $\G$ is a DAG.
        
        \textbf{Claim 1:} $W_{m+1}\in \Ch{X}{\G}$.
        
        \textit{proof of claim 1.} As $m<|\Ch{X}{\G}|$, $\mathbf{A} \cap \Ch{X}{\G}\neq \varnothing$ and $W_{m+1}$ can be either a child or a co-parent of $X$ (parents of $X$ have at least one descendant in $\mathbf{A}$.)
        Assume by contradiction that $W_{m+1}\in \Cp{X}{\G}$.
        In this case, $W_{m+1}$ does not have any common child with $X$ other than the variables in $\mathbf{W}$, because $W_{m+1}$ does not have any descendants in $\mathbf{A}$.
        
        Next, we will show that 
        \[ \dsep{X}{W_{m+1}}{\Mb{X}{\OV}\setminus (\{W_{m+1}\} \cup \mathbf{W})}{\G},\]
        which is against Equation \eqref{eq: removable-clique proof} for $\mathbf{S}=\mathbf{W}$ and $Y=W_{m+1}$. Therefore, $W_{m+1}\in \Ch{X}{\G}$, i.e., claim 1 holds.
        
        Suppose $\mathbf{B}:= \Mb{X}{\OV}\setminus (\{W_{m+1}\} \cup \mathbf{W})$ and let $\mathcal{P}=(X,V_1,\dots,V_t,W_{m+1})$ be a path between $X$ and $W_{m+1}$. 
        Note that $\mathcal{P}$ has length at least two since $X$ and $W_{m+1}$ are not neighbors.
        We have the following possibilities.
        \begin{itemize}[leftmargin=*]
            \item $V_1 \in \Pa{X}{\G}$:
                $V_1$ blocks $\mathcal{P}$ since $V_1$ is not a collider on $\mathcal{P}$ and $V_1 \in \mathbf{B}$.
            \item $V_1 = W_i\in \mathbf{W}$:
                From the hypothesis of the previous iteration we have $W_{m+1} \in \Pa{W_i}{\G}$. 
                Hence, $\mathcal{P}$ is not a directed path from $X$ to $W_{m+1}$ and has at least one collider.
                Let $Z$ be the collider on $\mathcal{P}$ closest to $X$.
                Hence, $Z \in \De{W_i}{\G}\cup \{W_i\}$, and neither $Z$ nor its descendants belongs to $\mathbf{B}$. 
                Therefore, $Z$ blocks $\mathcal{P}$.
            \item $V_1 \notin \mathbf{W}$, $V_1 \in \Ch{X}{\G}$, and $V_1$ is not a collider on $\mathcal{P}$: 
                $V_1$ blocks $\mathcal{P}$ since $V_1 \in \mathbf{B}$.
            \item $V_1\notin \mathbf{W}$, $V_1 \in \Ch{X}{\G}$, and $V_1$ is a collider on $\mathcal{P}$: 
                As $W_{m+1}$ does not have any common child with $X$ other than the variables in $\mathbf{W}$, $\mathcal{P}$ has at least length 3 ($t\geq 2$).
                In this case, $V_2 \neq W_{m+1}$ is a parent of $V_1$ and a co-parent of $X$. 
                Thus, $V_2 \in \mathbf{B}$. 
                Therefore, $V_2$ blocks $\mathcal{P}$ as it cannot be a collider on $\mathcal{P}$.
        \end{itemize}
        
        We now show that the two properties of Equation \eqref{eq: lem-W-k} hold for $\mathbf{W}\cup \{W_{m+1}\}$.
        The first property directly follows from  $W_{m+1}\in \Mb{X}{\OV}\setminus \mathbf{W}$ and the definition of $\mathbf{W}$.
        To verify the second property, suppose $Y\in \Mb{X}{\OV} \setminus (\mathbf{W}\cup \{W_{m+1}\})$.
        We need to show that $Y\in \Pa{W_{m+1}}{\G}$.
        
        Let us define $\mathbf{C}:= (\Mb{X}{\OV} \cup \{X\}) \setminus (\{Y,W_{k+1}\}\cup \mathbf{W})$.
        We now show that $\mathbf{C}$ blocks all the paths of length at least two between $Y$ and $W_{m+1}$. 
        Let $\mathcal{P}=(Y,V_1,\dots,V_t,W_{m+1})$ be a path of length at least two between $Y$ and $W_{m+1}$. 
        If the last edge in this path is $V_t \to W_{m+1}$, $V_t$ blocks this path since it is not a collider and it is a member of $\mathbf{C}$. 
        Now, suppose the last edge in this path is $V_t \gets W_{m+1}$. 
        Note that $Y \notin \De{W_{m+1}}{\G}$ by definition of $W_{m+1}$. 
        Thus, there exists at least one collider on $\mathcal{P}$. 
        Let $V_i$ be the closest collider of $\mathcal{P}$ to $W_{m+1}$. 
        Since $V_i\in\De{W_{m+1}}{\G}$, neither $V_i$ nor any of its descendants appear in $\mathbf{C}$. 
        Therefore, $V_i$ blocks $\mathcal{P}$. 
        Hence, $\mathbf{C}$ blocks all the paths of length at least two between $Y$ and $W_{m+1}$. 
        On the other hand, Equation \eqref{eq: removable-clique proof} implies that 
        \[
        \notdsep{Y}{W_{m+1}}{\mathbf{C}}{\G}.
        \]
        Hence, there must be a path of length one between $Y$ and $W_{m+1}$, i.e., $Y\in\N{W_{m+1}}{\G}$. 
        $Y$ cannot be a child of $W_{m+1}$ by the definition of $W_{m+1}$, and therefore $Y\in\Pa{W_{m+1}}{\G}$.
        This completes the proof.
    \end{proof}
    
    \begin{customlem} {\ref{lemma: FindRemovable clique}} \label{lemma: FindRemovable clique proof}
        Suppose $\G=(\OV,\E)$ is a DAG and a perfect map of $P_{\OV}$ such that $\omega(\G)\leq m$. 
        Vertex $X\in \OV$ is removable in $\G$ if for any $\mathbf{S}\subseteq\Mb{X}{\OV}$ with $\left\vert\mathbf{S}\right\vert\leq m-2$, we have
            \begin{equation} \label{eq: removable-clique proof}
                \begin{aligned}
                    &\notCI{Y}{Z}{\big(\Mb{X}{\OV} \cup \{X\}\big) \setminus \big(\{Y,Z\}\cup\mathbf{S}\big)}{P_{\OV}}, & \forall Y,Z \in \Mb{X}{\OV} \setminus \mathbf{S}, \\
                    &\notCI{X}{Y}{\Mb{X}{\OV} \setminus (\{Y\}\cup\mathbf{S})}{P_{\OV}}, & \forall Y \in \Mb{X}{\OV} \setminus \mathbf{S}.
                \end{aligned}
            \end{equation}
            Furthermore, the set of vertices that satisfy Equation \eqref{eq: removable-clique proof} is nonempty.
        \end{customlem}
    \begin{myproof}[Proof]
        First, we show that $|\Ch{X}{\G}|\leq m-1$:
        
        Assume by contradiction that $|\Ch{X}{\G}|>m-1$.
        Lemma \ref{lem: rem-clique-k} for $k=m-1$ implies that there exists $\mathbf{W}\subseteq \Ch{X}{\G}$ with $|\mathbf{W}|=m-1$ such that I) all the variables in $\mathbf{W}$ are neighbors, and II) each variable in $\Mb{X}{\OV}\setminus \mathbf{W}$ is a parent of each of the variables in $\mathbf{W}$.
        Since $|\Ch{X}{\G}|>m-1$, let $Z$ be a variable in $\Ch{X}{\G}\setminus \mathbf{W}$.
        Hence, $\mathcal{G}[\mathbf{W}\cup \{X,Z\}]$ is a complete graph of size $m+1$ which is against the assumption that $\omega(\G)\leq m$.
        Therefore, $|\Ch{X}{\G}|\leq m-1$.
        
        Now, Lemma \ref{lem: rem-clique-k} for $k=|\Ch{X}{\G}|$ implies that there exists $\mathbf{W} \subseteq \Ch{X}{\G}$ with $|\mathbf{W}|=|\Ch{X}{\G}|$ such that I) all the variables in $\mathbf{W}$ are neighbors, and II) each variable in $\Mb{X}{\OV}\setminus \mathbf{W}$ is a parent of each of the variables in $\mathbf{W}$.
        In this case, $\mathbf{W}$ must be equal to $\Ch{X}{\G}$ which indicates that all the variables in $\Mb{X}{\OV} \setminus \Ch{X}{\G}$ are parents of the children of $X$ while the children of $X$ are neighbors of each other. 
        Hence, Theorem \ref{theorem: removability} implies that $X$ is a removable variable in $\G$.
        
        Finally, if $X$ has no child, Equation \eqref{eq: removable-clique proof} holds. 
        Also, there exists at least one vertex with no child in every DAG.
        Hence, the set of vertices that satisfy Equation \eqref{eq: removable-clique proof} is nonempty.
    \end{myproof}

    \begin{customprp} {\ref{prop:bounded_clique_neighbors}} \label{prop:bounded_clique_neighbors_proof}
    Suppose $\G[\OV]$ is a DAG and a perfect map of $P_{\OV}$ s.t. $\omega(\mathcal{G[\OV]})\leq m$. Algorithm \ref{alg: FindRemovable clique} returns a removable vertex in $\G[\OV]$ by performing $\mathcal{O}(|\OV| \din{\G[\OV]}^{m})$ CI tests.
    \end{customprp}
    \begin{myproof}[Proof]
        The soundness of Algorithm \ref{alg: FindRemovable clique} follows directly from Lemma \ref{lemma: FindRemovable clique proof}.
        
        \emph{Complexity:}
            Suppose $X_{i^*}$ is the output of Algorithm \ref{alg: FindRemovable clique}. 
            In this case, the algorithm has performed CI tests in line 5 to verify Equation \eqref{eq: removable-clique proof} for $X_1,\dots, X_{i^*}$. 
            For each $1\leq j \leq i^*$, it performs at most 
            \[
                \sum_{s=0}^{m-2} \binom{|\Mb{X_j}{\OV}|}{s} [\binom{|\Mb{X_j}{\OV}|-s}{2} + |\Mb{X_j}{\OV}|-s]
            \]
            CI tests. 
            Since $X_{i^*}$ is removable in $\G[\OV]$, Lemma \ref{lemma: Mb size of removable} implies that $|\Mb{X_{j}}{\OV}|\leq |\Mb{X_{i^*}}{\OV}| \leq \din{\G[\OV]}$. 
            Hence, Algorithm \ref{alg: FindRemovable clique} has performed
            \[
                \mathcal{O}(i^* \sum_{s=0}^{m-2} \din{\G[\OV]}^{s+2}) 
                = \mathcal{O}(i^* \din{\G[\OV]}^{m})
            \]
            CI tests.
            Note that $i^*\leq |\OV|$.
      \end{myproof}
      
    
    \begin{customlem} {\ref{lemma: FindNeighbors clique}} \label{lemma: FindNeighbors clique proof}
        Suppose $\G[\OV]$ is a DAG and a perfect map of $P_{\OV}$ with  $\omega(\mathcal{G[\OV]})\leq m$. Let $X\in \OV$ be a vertex that satisfies Equation \eqref{eq: removable-clique} and $Y\in \Mb{X}{\OV}$. In this case, $Y\in \Cp{X}{\G}$ if and only if
        \begin{equation} \label{eq: neighborhood-clique proof}
            \exists \mathbf{S}\subseteq\Mb{X}{\OV}\setminus\{Y\}\!:\ \ \left\vert\mathbf{S}\right\vert=(m-1)\,,\ \ \CI{X}{Y}{\Mb{X}{\OV}\setminus(\{Y\}\cup\mathbf{S})}{P_{\OV}}.
        \end{equation}
    \end{customlem}
    \begin{myproof}[Proof]
        The if part is straightforward as $\Mb{X}{\OV}=\N{X}{\G}\cup \Cp{X}{\G}$.
        
        \emph{only if part:}
        As shown in the proof of Lemma \ref{lemma: FindRemovable clique proof}, I) $|\Ch{X}{\G}|\leq m-1$, II) children of $X$ are neighbors of each other, and III) each variable in $\Mb{X}{\OV}\setminus \Ch{X}{\G}$ is a parent of all the children of $X$.
        
        We first show that if $Y\in \Cp{X}{\G}$, then
        \[
        \dsep{X}{Y}{\Mb{X}{\OV}\setminus(\{Y\}\cup \Ch{X}{\G})}{\G}.
        \]
        Let $\mathcal{P}=(X,V_1,\dots,V_t,Y)$ be a path between $X$ and $Y$.
        If $V_1\in \Pa{X}{\G}$, $V_1$ blocks $\mathcal{P}$ as it is not a collider on $\mathcal{P}$ and $V_1 \in \Mb{X}{\OV}\setminus(\{Y\}\cup \Ch{X}{\G})$.
        Now, suppose $V_1\in \Ch{X}{\G}$.
        Since $Y\in \Pa{V_1}{\G}$, $Y\notin \De{V_1}{\G}$ and $\mathcal{P}$ is not a directed path from $X$ to $Y$.
        Hence, $\mathcal{P}$ contains at least one collider. 
        Let $Z$ be the collider on $\mathcal{P}$ closest to $V_1$.
        In this case, $Z\in \De{V_1}{\G}$.
        Hence, $Z$ blocks $\mathcal{P}$ since all the variables in $\Mb{X}{\OV}\setminus(\{Y\}\cup\Ch{X}{\G})$ are parents of $V_1$ and they cannot be a descendent of $Z$.
        Therefore, $\Mb{X}{\OV}\setminus(\{Y\}\cup \Ch{X}{\G})$ d-separates $X$ and $Y$.
        
        On the other hand, if $\Cp{X}{\G}\neq \varnothing$ and  $|\Ch{X}{\G}|\leq m-2$, then Equation \eqref{eq: removable-clique} with $\mathbf{S}=\Ch{X}{\G}$ implies that 
        $\notCI{X}{Y}{\Mb{X}{\OV}\setminus(\{Y\}\cup \Ch{X}{\G})}{P_{\OV}}$ which is against what we just proved.
        Hence, either $\Cp{X}{\G}= \varnothing$ or $|\Ch{X}{\G}|= m-1$.
        In the first case, the claim is trivial, and for the second case, Equation \eqref{eq: neighborhood-clique proof} holds for $\mathbf{S}= \Ch{X}{\G}$.
    \end{myproof}

    \begin{customtheorem} {\ref{thm: upperbound clique}} \label{thm: upperbound clique proof}
    Suppose $\G=(\V,\E)$ is a DAG and a perfect map of $\PV$ with $\omega(\mathcal{G})\leq m$. Then, \textbf{RSL} (Algorithm \ref{alg: RSL}) with sub-algorithms \ref{alg: FindRemovable clique} and \ref{alg: FindNeighbors clique} is sound and complete, and performs $\mathcal{O}(|\V|^2 \din{\G}^{m})$ CI tests.
    \end{customtheorem}
    \begin{myproof}[Proof] 
    \emph{Soundness:}
        Suppose $\G[\OV]$ is a perfect map of $P_{\OV}$ in a recursion.
        According to Proposition \ref{prop:bounded_clique_neighbors}, \textbf{FindRemovable} function correctly finds a removable variable in $\G[\OV]$. 
        Then, \textbf{FindNeighbors} function correctly finds $\N{X}{\G[\OV]}$ and $\mathbfcal{S}_X$ according to Lemma \ref{lemma: FindNeighbors clique}.
        Then, \textbf{UpdateMb} function correctly updates $\MB{\OV \setminus \{X\}}$ according to Proposition \ref{prp: updatemb}.
        Hence, we call the \textbf{RSL} function for the next recursion with correct Markov boundaries.
        Moreover, Proposition \ref{prp: removable} implies that $\G[\OV \setminus \{X\}]$ is a perfect map of $P_{\OV \setminus \{X\}}$. 
        Therefore, as we initially assume that $\G$ is a perfect map of $\PV$, $\G[\OV]$ is a perfect map of $P_{\OV}$ throughout all the recursions, and \textbf{RSL} correctly outputs $(\mathcal{H}[\OV],\, \mathbfcal{S}_{\OV})$.
    
    \emph{Complexity:} 
        In each recursion, Lemma \ref{lemma:  Mb size of removable} implies that $|\Mb{X}{\OV}| \leq \din{\G[\OV]}$.
        Hence, According to Propositions \ref{prp: updatemb}, \ref{prop:bounded_clique_neighbors} and Lemma \ref{lemma: FindNeighbors clique}, Algorithm \ref{alg: RSL} performs $\mathcal{O}(|\V|\din{\G}^m)$ CI tests at each recursion. 
        Therefore, it performs $\mathcal{O}(|\V|^2 \din{\G}^{m})$ CI tests in total.
    \end{myproof}
    
    \subsection{Proofs of Section \ref{sec: without side information}:}
    
       \begin{customprp}
        {\ref{prp: verifiable}}
        [Verifiable]
        \label{prp: verifiable app}
        Suppose $\G=(\V,\E)$ is a DAG with skeleton $\mathcal{H}$ that is a perfect map of $\PV$. 
        If the \textbf{RSL} with sub-algorithms \ref{alg: FindRemovable clique} and \ref{alg: FindNeighbors clique}, and input $m>0$ terminates, then the clique number of the learned skeleton is 
        at least  $\omega(\mathcal{G})$.
    \end{customprp}

    \begin{myproof}
        To show this result, it suffices to prove that if an edge $e=\{X,Y\}$ belongs to $\mathcal{H}$, it also belongs to the output of \textbf{RSL} with $m>0$ and sub-algorithms \ref{alg: FindRemovable clique} and \ref{alg: FindNeighbors clique}. 
        
        Since $X$ and $Y$ are neighbors, then there exists no separating set for them in any subgraph $\G[\OV]$, where $X,Y\in\OV$. 
        Hence, sub-algorithm \ref{alg: FindNeighbors clique} will not eliminate $Y$ from $\N{X}{\G[\OV]}$ and vice versa. 
        Furthermore, if $W$ is a removable vertex in $\G[\OV]$, function \textbf{UpdateMB} in Algorithm \ref{alg: update Mb} will not remove $X$ and $Y$ from $\Mb{Y}{\OV \setminus \{W\}}$ and $\Mb{X}{\OV \setminus \{W\}}$, respectively. 
        This ensures that \textbf{RSL} with any input $m>0$ will not remove the edge between $X$ and $Y$.
    \end{myproof}
    
\subsection{Proofs of Section \ref{sec: diamond-free}:} \label{sec: diamond-free proofs}
    \begin{customlem}{\ref{lemma: FindRemovable}}\label{lemma: FindRemovable proof}
        Suppose $\G=(\OV,\E)$ is a diamond-free DAG and a perfect map of $P_{\OV}$. Vertex $X\in \OV$ is removable in $\G$ if and only if
        \begin{align} \label{eq: proof diamond-free}
            &\notCI{Y}{Z}{(\Mb{X}{\OV} \cup \{X\} )\setminus \{Y,Z\}}{P_{\OV}},\ \ \ \ \forall Y,Z \in \Mb{X}{\OV}.
        \end{align}
        Furthermore, the set of removable vertices is nonempty.
    \end{customlem}
    \begin{myproof}[Proof]
        We first show in Lemma \ref{lemma: limb1} that Equation \eqref{eq: proof diamond-free} holds for a vertex $X$ if and only if
            \begin{align}\label{eq: proof_l1_short}
                \exists W\in\Ch{X}{\G}\cup\{X\}\ \ \text{such that}\ \  \Mb{X}{\OV} \cup \{X\} = \Pa{W}{\G} \cup \{W\}.
            \end{align}
        Afterwards, in lemmas \ref{lemma: limb2} and \ref{lemma: limb3}, we show that in a diamond-free graph $\G$, vertex $X$ is removable if and only if \eqref{eq: proof_l1_short} holds. This concludes the proof of Lemma \ref{lemma: FindRemovable}. 
        
        To show that the set of removable variables is nonempty, we have that if $X$ has no children, then \eqref{eq: removable-diamond} holds. 
        On the other hand, a DAG always contains a vertex with no children. Thus, the set of removable vertices of a diamond-free DAG is non-empty.
    \end{myproof}

    \begin{lemma} \label{lemma: limb1}
        Suppose $\G=(\OV,\E)$ is a DAG and a perfect map of $P_{\OV}$. 
        Equation \eqref{eq: proof diamond-free} holds for a vertex $X\in \OV$ if and only if there exists $W\in\Ch{X}{\G}\cup\{X\}$, such that
        \begin{equation} \label{eq: limb}
            \Mb{X}{\OV} \cup \{X\} = \Pa{W}{\G} \cup \{W\}.
        \end{equation}
    \end{lemma}
    \begin{proof}
        \emph{If part:} 
        Suppose $Y,Z\in \Mb{X}{\OV}$ and $\mathbf{S}= \Mb{X}{\OV} \cup \{X\} \setminus \{Y,Z\}$. 
        We need to show that $\notdsep{Y}{Z}{\mathbf{S}}{\G}$. 
        If either $W=Y$ or $W=Z$, then $Y$ and $Z$ are neighbors because of Equation \eqref{eq: limb}, and therefore, they are not d-separable. 
        Otherwise, $Y,Z\in\Pa{W}{\G}$ and $W\in \mathbf{S}$. 
        In this case, $Y\to W \gets Z$ is an active path and $\mathbf{S}$ does not d-separate $Y$ and $Z$.
        
        \emph{Only if part:}
        Suppose $X\in \OV$ satisfies Equation \eqref{eq: proof diamond-free}. 
        Take $W$ to be a vertex in $\Ch{X}{\G} \cup \{X\}$ that has no nontrivial descendants\footnote{A nontrivial descendant of a variable is a descendant other than itself.} in $\Mb{X}{\OV} \cup \{X\}$. 
        Note that such a vertex exists due to the acyclicity of $\G$. 
        Since $W\in \Ch{X}{\G} \cup \{X\}$, $\Pa{W}{\G} \cup \{W\} \subseteq \Mb{X}{\OV} \cup \{X\}$. 
        It is enough to show that $\Mb{X}{\OV} \cup \{X\} \subseteq \Pa{W}{\G} \cup \{W\}$. 
        Suppose $T \in \Mb{X}{\OV} \setminus\{W\}$ and $\mathbf{S}=\Mb{X}{\OV}\cup\{X\}\setminus\{T,W\}$. 
        It suffices to show that $T\in \Pa{W}{\G}$. 
        We now show that $\mathbf{S}$ blocks all the paths of length at least two between $T$ and $W$. 
        Let $\mathcal{P}=(T,V_1,\dots,V_k,W)$ be a path of length at least two between $T$ and $W$. 
        If the last edge in this path is $V_k \to Z$, $V_k$ blocks this path since it is not a collider and it is a member of $\mathbf{S}$. 
        Now suppose the last edge in this path is $V_k \gets W$. 
        Note that $T \notin \De{W}{\G}$ by definition of $W$. 
        Thus, there exists at least one collider on $\mathcal{P}$. 
        Let $V_i$ be the closest collider of $\mathcal{P}$ to $W$. 
        Since $V_i\in\De{W}{\G}$, neither $V_i$ nor any of its descendants appear in $\mathbf{S}$. 
        Therefore, $V_i$ blocks $\mathcal{P}$. 
        Hence, $\mathbf{S}$ blocks all the paths of length at least two between $T$ and $W$. 
        Since Equation \eqref{eq: proof diamond-free} holds, there must be a path of length one between $T$ and $W$, i.e., $T\in\N{W}{\G}$. 
        $T$ cannot be a child of $W$ by definition of $W$, and therefore $T\in\Pa{W}{\G}$.
    \end{proof} 
    
    \begin{lemma}\label{lemma: limb2}
        Suppose $\G = (\OV,\E)$ is a DAG and a perfect map of $P_{\OV}$, and $X\in \OV$ is a removable variable in $\G$. There exists a variable $W\in\Ch{X}{\G}\cup\{X\}$ such that Equation \eqref{eq: limb} holds.
    \end{lemma}
    \begin{proof}
        Take $W$ as a vertex in $\Ch{X}{\G}\cup\{X\}$ that has no children in $\Mb{X}{\OV}\cup\{X\}$. Note that such a vertex exists due to acyclicity. We will show Equation \eqref{eq: limb} holds for this $W$.
        
        If $W=X$, then the claim is trivial as $X$ has no children and $\Mb{X}{\OV}=\Pa{X}{\G}$. 
        Otherwise, $W\in\Ch{X}{\G}$. 
        Hence, $\Pa{W}{\G}\cup\{W\}\subseteq\Mb{X}{\OV}\cup\{X\}$. 
        Now take an arbitrary vertex $T \in\Mb{X}{\OV} \setminus \{W\}$. 
        It suffices to show that $T\in \Pa{W}{\G}$.
        
        If $T \in \N{X}{\G}\setminus\{W\}$, according to Condition 1 of removability (Theorem \ref{theorem: removability}), all of the neighbors of $X$ are adjacent to the children of $X$. 
        Hence, $T$ and $W$ are neighbors. 
        Since $W$ has no children in $\Mb{X}{\OV}\cup\{X\}$, $T\in\Pa{W}{\G}$. 
        
        If $T \in \Cp{X}{\G}$, then $X$ and $T$ have at least a common child $Y$. If $Y = W$, then $W\in\Pa{Z}{\G}$. Otherwise, $Y \in \Ch{X}{\G}\setminus\{W\}$. As we showed in the previous case, $Y$ is a parent of $W$, and therefore, $Y\in\Ch{X}{\G}\cap\Pa{W}{\G}$. Hence, Condition 2 of Theorem \ref{theorem: removability} implies that $T\in\Pa{W}{\G}$, which completes the proof. 
    \end{proof}
    
    \begin{lemma} \label{lemma: limb3}
        Suppose $\G=(\OV,\E)$ is a diamond-free DAG and a perfect map of $P_{\OV}$. $X\in \OV$ is removable in $\G$ if there exists $W\in\Ch{X}{\G} \cup \{X\}$ such that $\Mb{X}{\OV} \cup \{X\} = \Pa{W}{\G} \cup \{W\}$. 
    \end{lemma}
    \begin{proof}
        In order to prove the removability of $X$ in $\G$, we show that the conditions of Theorem \ref{theorem: removability} are satisfied. 
        Take an arbitrary $W'\in\Ch{X}{\G}$. 
        If $W'=W$, both conditions are satisfied as all vertices in $\Mb{X}{\OV}$ are parents of $W'$. 
        Now suppose $W'\in\Ch{X}{\G}\setminus\{W\}$. 
        In this case, $W'\in\Pa{W}{\G}$.
        
        Condition 1: 
        Suppose $Y\in\N{X}{\G}$. 
        We need to show that $Y$ and $W'$ are neighbors. 
        Consider the induced subgraph $\G[\{X,W,W',Y\}]$,
        where we have $X,W',Y\in\Pa{W}{\G}$, $X\in \Pa{W'}{\G}$ and $X \in \N{Y}{\G}$. 
        Since $\G$ is diamond-free, $W'$ and $Y$ must be neighbors.
        
        Condition 2: 
        Suppose $Y \in \Ch{X}{\G}\cap\Pa{W'}{\G}$, and $T \in \Pa{Y}{\G}$. 
        We need to show that $T\in \Pa{W'}{\G}$. 
        Now, consider the induced subgraph $\G[\{T,Y,W',W\}]$,
        where we have $T,Y,W'\in\Pa{W}{\G}$, $T\in\Pa{Y}{\G}$ and $Y\in\Pa{W'}{\G}$. 
        Since $\G$ is diamond-free, $T$ and $W'$ must be neighbors, which implies that $T\in\Pa{W'}{\G}$ due to acyclicity of $\G$.
    \end{proof}
	
	\begin{customprp} {\ref{prp: findremovable}} \label{prp: findremovable proof}
        Suppose $\G[\OV]$ is a diamond-free DAG and a perfect map of $P_{\OV}$. \textbf{FindRemovable} returns a removable vertex in $\G[\OV]$ by performing at most $|\OV| \binom{\din{\G[\OV]}}{2}$ CI tests.
    \end{customprp}
    \begin{myproof} [Proof]
    \emph{Soundness:}
    $\G[\OV]$ is a diamond-free graph and a perfect map of $P_{\OV}$.
    Hence, Lemma \ref{lemma: FindRemovable} implies that I) $\G[\OV]$ has at least one removable vertex, and II) Equation \eqref{eq: removable-diamond} holds for $X_i \in \OV$ if and only if $X_i$ is removable in $\G[\OV]$.
    Therefore, the output of \textbf{FindRemovable} function is a removable vertex. 
    
    \emph{Complexity:}
    Suppose $X_{i^*}$ is the output of \textbf{FindRemovable}. 
    In this case, the function has performed CI tests in line 5 to verify Equation \eqref{eq: removable-diamond} for $X_1,\dots, X_{i^*}$. 
    For each $1\leq j \leq i^*$, it performs at most $\binom{|\Mb{X_j}{\OV}|}{2}$ CI tests. 
    Since $X_{i^*}$ is removable in $\G[\OV]$, Lemma \ref{lemma: Mb size of removable} implies that $|\Mb{X_{j}}{\OV}| \leq |\Mb{X_{i^*}}{\OV}| \leq \din{\G[\OV]}$. 
    Hence, \textbf{FindRemovable} has performed at most $i^* \binom{\din{\G[\OV]}}{2}$ CI tests. 
    Note that $i^* \leq |\OV|$.
    \end{myproof}
    
    \begin{customlem}{\ref{lemma: FindNeighbors}}\label{lemma: find neighborhood proof}
    Suppose $\G=(\OV,\E)$ is a diamond-free DAG and a perfect map of $P_{\OV}$. Let $X\in \OV$ be a removable vertex in $\G$, and $Y\in \Mb{X}{\OV}$. In this case, $Y\in \Cp{X}{\G}$ if and only if 
        \begin{equation}\label{eq: neighborhood-diamond proof}
            \exists Z\in \Mb{X}{\OV}\setminus \{Y\}\!:\ \  \CI{X}{Y}{\Mb{X}{\OV}\setminus\{Y,Z\}}{P_{\OV}}.
        \end{equation}
    \end{customlem}
    \begin{myproof} [Proof]
        \emph{If part:}
        Since $\G$ is a perfect map of $P_{\OV}$,
        \[\Mb{X}{\OV} = \N{X}{\G} \cup \Cp{X}{\G}.\]
	    Equation \eqref{eq: neighborhood-diamond proof} implies that $X$ and $Y$ are d-separable. Hence, they are not neighbors and $Y \in \Cp{X}{\G}$.
	    
        \emph{Only if part:}
        Since $\G$ is a perfect map of $P_{\OV}$ and $X$ is a removable variable in $\G$, Lemma \ref{lemma: limb2} implies that there exists a variable $Z \in \Ch{X}{\G} \cup \{X\}$ such that
        \begin{equation} \label{eq: Mb is parent of Z}
            \Mb{X}{\OV} \cup \{X\} = \Pa{Z}{\G} \cup \{Z\}.
        \end{equation}
        We will show that Equation \eqref{eq: neighborhood-diamond proof} holds for this $Z$.
        
        We first show that if $Z\neq X$, then $Z$ is the only common child of $X$ and $Y$: 
        Assume by contradiction that $W\neq Z$ is a common child of $X$ and $Y$.
        Since $W \in \Mb{X}{\OV}\cup \{X\}$, $W$ is a parent of $Z$. 
        Therefore, $\G[\{X,W,Y,Z\}]$ is a diamond graph which is against our assumption. 
        
        Now, we need to prove that $\mathbf{S}=\Mb{X}{\OV}\setminus\{Y,Z\}$ d-separates $X$ and $Y$, i.e., blocks all the paths between them. 
        Let $\mathcal{P}=(X,V_1,\dots,V_k,Y)$ be a path between $X$ and $Y$. 
        Note that $\mathcal{P}$ has length at least two since $X$ and $Y$ are not neighbors.
        We have the following possibilities.
        \begin{itemize}[leftmargin=*]
            \item $V_1 \in \Pa{X}{\G}$:
                $V_1$ blocks $\mathcal{P}$ since $V_1$ is not a collider on $\mathcal{P}$ and $V_1 \in \mathbf{S}$.
            \item $V_1 = Z$:
                In this case, $Z\in \Ch{X}{\G}$ as $Z\neq X$.
                Since $Y\in \Mb{X}{\OV}\setminus\{Z\}\subset \Pa{Z}{\G}$, $\mathcal{P}$ is not a directed path from $X$ to $Y$ and has at least one collider.
                Let $W$ be the collider on $\mathcal{P}$ closest to $X$.
                Hence, $W \in \De{Z}{\G}\cup \{Z\}$, and neither $W$ nor its descendants belongs to $\mathbf{S}$. 
                Therefore, $W$ blocks $\mathcal{P}$.
            \item $V_1 \neq Z$, $V_1 \in \Ch{X}{\G}$, and $V_1$ is not a collider on $\mathcal{P}$: 
                $V_1$ blocks $\mathcal{P}$ since $V_1 \in \mathbf{S}$.
            \item $V_1\neq Z$, $V_1 \in \Ch{X}{\G}$, and $V_1$ is a collider on $\mathcal{P}$: 
                Since $\Ch{X}{\G}\neq\varnothing$, $Z\neq X$, and $Z$ is the only common child of $X$ and $Y$. 
                Hence, $\mathcal{P}$ has at least length 3 ($k\geq 2$).
                In this case, $V_2 \neq Y$ is a parent of $V_1$ and a co-parent of $X$. 
                Thus, $V_2 \in \mathbf{S}$. 
                Therefore, $V_2$ blocks $\mathcal{P}$ as it cannot be a collider on $\mathcal{P}$.
        \end{itemize}
        In all the cases, $\mathbf{S}$ blocks $\mathcal{P}$, and therefore, d-separates $X$ and $Y$.
    \end{myproof}
    
    In order to prove Theorem \ref{thm: upperbound}, we first show the following result. 
    \begin{lemma} \label{prp: find neighbor}
        Suppose $\G[\OV]$ is a diamond-free DAG and a perfect map of $P_{\OV}$. 
        If $X$ is removable in $\G[\OV]$, then \textbf{FindNeighbors} function for learning $\N{X}{\G[\OV]}$ and $\mathbfcal{S}_X$ is sound, and performs $\mathcal{O}(\din{\G[\OV]}^2)$ number of CI tests.
    \end{lemma}
    \begin{myproof}[Proof] 
    \emph{Soundness:} 
    According to the definition of Markov boundary, for every $Y \in \OV \setminus \Mb{X}{\OV}$,
    \[\CI{X}{Y}{\Mb{X}{\OV}}{P_{\OV}}.\] 
    Since $\G[\OV]$ is a perfect map of $P_{\OV}$, 
    \[\CI{X}{Y}{\Mb{X}{\OV}}{P_{\OV}} \iff \dsep{X}{Y}{\Mb{X}{\OV}}{\G[\OV]}.\]
    Therefore, line 3 of Algorithm \ref{alg: FindNeighbors clique} correctly finds a separating set for $X$ and the variables in $\OV \setminus \Mb{X}{\OV}$.
   
    Since $\G[\OV]$ is a perfect map of $P_{\OV}$, Equation \eqref{eq: Mb with perfect map assumption} implies that
    \[\Mb{X}{\OV} = \N{X}{\G[\OV]} \cup \Cp{X}{\G[\OV]}.\]
    For every $Y \in \Mb{X}{\OV}$, Algorithm \ref{alg: FindNeighbors clique} checks whether Equation \eqref{eq: neighborhood-diamond} holds for $Y$. 
    If Equation \eqref{eq: neighborhood-diamond} holds for $Y,$ Lemma \ref{lemma: FindNeighbors} implies that $Y \in \Cp{X}{\G[\OV]}$ and $\Mb{X}{\G[\OV]}\setminus \{Y,Z\}$ is a separating set for $X$ and $Y.$ 
    Otherwise, it implies that $Y \in \N{X}{\G[\OV]}$. 
    Hence, Algorithm \ref{alg: FindNeighbors clique} correctly finds a separating set for $X$ and $\Cp{X}{\G[\OV]}$ in line 6 and correctly identifies $\N{X}{\G[\OV]}$ in line 8.
    
    \emph{Complexity:} Algorithm \ref{alg: FindNeighbors clique} performs at most $|\Mb{X}{\OV}|-1$ CI tests in line 5 to check Equation \eqref{eq: neighborhood-diamond} for each $Y \in \Mb{X}{\OV}$. 
    Since $X$ is removable in $\G[\OV]$, Lemma \ref{lemma:  Mb size of removable} implies that $|\Mb{X}{\OV}| \leq \din{\G[\OV]}$. Thus, Algorithm \ref{alg: FindNeighbors clique} performs $\mathcal{O}(\din{\G[\OV]}^2)$ CI tests.
    \end{myproof}

    \begin{customtheorem} {\ref{thm: upperbound}} \label{thm: upper bound proof}
        Suppose $\G=(\V,\E)$ is a diamond-free DAG and a perfect map of $\PV$. RSL$_D$ is sound and complete, and performs $\mathcal{O}(|\V|^2\din{\G}^2)$ CI tests.
    \end{customtheorem}
    \begin{myproof}[Proof]
    \emph{Soundness:}
        Suppose $\G[\OV]$ is a perfect map of $P_{\OV}$ in a recursion.
        According to Proposition \ref{prp: findremovable}, \textbf{FindRemovable} function correctly finds a removable variable in $\G[\OV]$. 
        Then, \textbf{FindNeighbors} function correctly finds $\N{X}{\G[\OV]}$ and $\mathbfcal{S}_X$ according to Lemma \ref{prp: find neighbor}.
        Then, \textbf{UpdateMb} function correctly updates $\MB{\OV \setminus \{X\}}$ according to Proposition \ref{prp: updatemb}.
        Hence, we call the \textbf{RSL} function for the next recursion with correct Markov boundaries.
        Moreover, Proposition \ref{prp: removable} implies that $\G[\OV \setminus \{X\}]$ is a perfect map of $P_{\OV \setminus \{X\}}$. 
        Therefore, as we initially assume that $\G$ is a perfect map of $\PV$, $\G[\OV]$ is a perfect map of $P_{\OV}$ throughout all the recursions, and \textbf{RSL} correctly outputs $(\mathcal{H}[\OV],\, \mathbfcal{S}_{\OV})$.
    
    \emph{Complexity:} 
        In each recursion, Lemma \ref{lemma:  Mb size of removable} implies that $|\Mb{X}{\OV}| \leq \din{\G[\OV]}$.
        Hence, According to Propositions \ref{prp: updatemb}, \ref{prp: findremovable} and Lemma \ref{prp: find neighbor}, Algorithm \ref{alg: RSL} performs $\mathcal{O}(|\V|\din{\G}^2)$ CI tests at each recursion. 
        Therefore, it performs $\mathcal{O}(|\V|^2\din{\G}^2)$ CI tests in total.
    \end{myproof}
    \subsection{Proofs of Section \ref{sec: discussion}:}
        \begin{customlem}{\ref{lemma: erdos-renyi}}
            A random graph $\G$ generated from Erdos-Renyi model $G(n,p)$ is diamond-free with high probability when $pn^{0.8}\!\rightarrow\!0$ and $\omega(\G)\leq m$ when $p n^{2/m}\!\rightarrow\!0$.
        \end{customlem}
   \begin{proof}
       From the theory of random graphs \cite{gilbert1959random}, we know that any fixed graph $\G'$, with $n_{\G'}$ vertices and $e_{\G'}$ edges does not appear in an Erdos-Renyi graph $G(n,p)$ with high probability as long as $pn^{1/f(\G')}\!\rightarrow\!0$, as $n\!\rightarrow\!\infty$, where $f(\G')\!:=\!\max\{e_{\mathcal{K}}/n_{\mathcal{K}}: \mathcal{K}\subseteq \G'\}$.
        This implies that the realizations of $G(n,p)$ are diamond-free with high probability when $pn^{0.8}\!\rightarrow\!0$ and their clique numbers are bounded by $m$ when $p n^{2/m}\!\rightarrow\!0$.
   \end{proof}
   
\section{RSL$_D$ on networks with diamonds} \label{sec: apd_verify}
    \begin{figure}[ht] 
	    \centering
		\tikzstyle{block} = [circle, inner sep=1.3pt, fill=black]
		\tikzstyle{input} = [coordinate]
		\tikzstyle{output} = [coordinate]
		\begin{subfigure}[b]{0.19\textwidth}
    		\centering
            \begin{tikzpicture}
                \tikzset{edge/.style = {->,> = latex',-{Latex[width=2mm]}}}
                \node[block] (A) at  (0,1) {};
                \node[] ()[above = -0.06cm of A]{$A$};
                \node[block] (B) at  (-0.9,0.5) {};
                \node[] ()[above left = -0.05cm and -0.2 cm of B]{$B$};
                \node[block] (C) at  (1,0.5) {};
                \node[] ()[above right = -0.05cm and -0.2 cm of C]{$C$};
                \node[block] (D) at  (0,0) {};
                \node[] ()[below = -0.06cm of D]{$D$};
                \node[] ()[below = 0.5cm of D]{};
                \draw[edge] (A) to (B);
                \draw[edge] (A) to (C);
                \draw[edge] (A) to (D);
                \draw[edge] (B) to (D);
                \draw[edge] (C) to (D);
            \end{tikzpicture}
            \caption{The true DAG}
            \label{fig: apd_ver dag}
        \end{subfigure}\hspace{0.05\textwidth}
        \begin{subfigure}[b]{0.19\textwidth}
    		\centering
            \begin{tikzpicture}
                \tikzset{edge/.style = {->,> = latex',-{Latex[width=2mm]}}}
                \node[block] (A) at  (0,1) {};
                \node[] ()[above = -0.06cm of A]{$A$};
                \node[block] (B) at  (-0.9,0.5) {};
                \node[] ()[above left = -0.05cm and -0.2 cm of B]{$B$};
                \node[block] (C) at  (1,0.5) {};
                \node[] ()[above right = -0.05cm and -0.2 cm of C]{$C$};
                \node[block] (D) at  (0,0) {};
                \node[] ()[below = -0.06cm of D]{$D$};
                \draw[edge, -] (B) to (A);
                \draw[edge, -] (C) to (A);
                \draw[edge] (A) to (D);
                \draw[edge] (B) to (D);
                \draw[edge] (C) to (D);
            \end{tikzpicture}
            \caption{Learned essential graph when $D$ is removed first}
            \label{fig: apd_ver true ess}
        \end{subfigure}\hspace{0.05\textwidth}
        \begin{subfigure}[b]{0.19\textwidth}
    		\centering
            \begin{tikzpicture}
                \tikzset{edge/.style = {-}}
                \node[block] (A) at  (0,1) {};
                \node[] ()[above = -0.06cm of A]{$A$};
                \node[block] (B) at  (-0.9,0.5) {};
                \node[] ()[above left = -0.05cm and -0.2 cm of B]{$B$};
                \node[block] (C) at  (1,0.5) {};
                \node[] ()[above right = -0.05cm and -0.2 cm of C]{$C$};
                \node[block] (D) at  (0,0) {};
                \node[] ()[below = -0.06cm of D]{$D$};
                \draw[edge] (A) to (B);
                \draw[edge, red] (C) to (B);
                \draw[edge] (C) to (A);
                \draw[edge] (A) to (D);
                \draw[edge] (B) to (D);
                \draw[edge] (C) to (D);
            \end{tikzpicture}
            \caption{Learned essential graph when $A$ is removed first}
            \label{fig: apd_ver wrong ess}
        \end{subfigure}
        \caption{$RSL_D$ on networks including diamonds.}
        \label{fig: apd_verify}
    \end{figure}
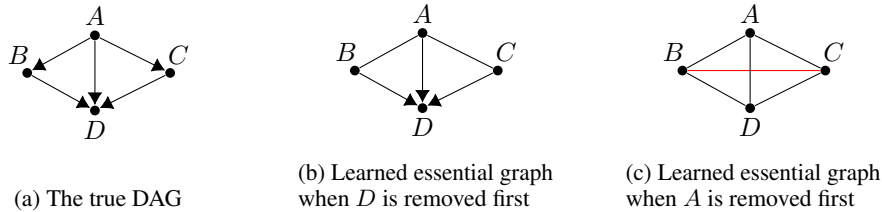 
    In this section, we discuss how $RSL_D$ performs on graphs including diamonds.
    Consider the DAG in Figure \ref{fig: apd_ver dag}, which is one of the forbidden structures in Figure \ref{fig: diamond}.
    All vertices $A,B,C$ and $D$ satisfy the conditions of Equation \eqref{eq: removable-diamond}, and all of them have a Markov boundary of size 3.
    As a result, $RSL_D$ will randomly choose one of them as the first removable node and continue with the remaining structure.
    However, the vertex $A$ is not removable according to Definition \ref{def: removable}.
    As a result, if $RSL_D$ removes one of $B,C$, or $D$ first, it will recover the true essential graph depicted in Figure \ref{fig: apd_ver true ess}.
    On the other hand, if $RSL_D$ removes $A$ first, an erroneous extra edge between the vertices $B$ and $C$ will appear in the output structure, as depicted in red in Figure \ref{fig: apd_ver wrong ess}.
    Although $RSL_D$ can make errors in recovering the BNs with diamonds as in this example, the next proposition shows that $RSL_D$ can only infer extra edges, i.e., the recall of $RSL_D$ is always equal to one, even if the true BN includes diamonds. 
    
    \begin{proposition}\label{prp: apd_ver}
        If RSL$_D$ terminates, the recovered BN structure contains no false negative edges. 
    \end{proposition}
    \begin{proof}
        Suppose the true BN is the DAG $\G=(\mathbf{V},\mathbf{E})$, and $\hat{\G}$ is the graph structure recovered from the output of $RSL_D$.
        It suffices to prove every edge in $\G$ also exists in $\hat{\G}$.
        Take an arbitrary edge $(X,Y)\in \mathbf{E}$.
        Since $X$ and $Y$ are neighbors, $X$ and $Y$ are in each other's Markov boundary at the beginning.
        Throughout $RSL_D$, Markov boundaries are only updated through Algorithm \ref{alg: update Mb}, where vertices $X$ and $Y$ are removed from each other's Markov boundary if either a separating set for them is found, or one of them is removed.
        Since $X$ and $Y$ are neighbors, no separating set for them exists.
        As a result, as long as none of $X$ and $Y$ are removed, they stay in each other's Mb.
        Suppose without loss of generality that $X$ is removed first, where the set of remaining variables is $\overline{\V}$.
        Since $Y$ is still in $\Mb{X}{\overline{\V}}$, and no separating set exists for $X$ and $Y$ (Equation \ref{eq: neighborhood-diamond} does not hold for any $Z$), function \textbf{FindNeighbors} adds $Y$ to $N_{\G[\overline{\V}]}(X)$ in line 8.
        Therefore, $X$ and $Y$ are neighbors in the output of $RSL_D$, and $\hat{\G}$, which completes the proof.
    \end{proof}

    Based on Proposition \ref{prp: apd_ver}, we propose the following implementation of \textbf{FindRemovable} for $RSL_D$, which does not require the diamond-freeness of the true BN, and yet guarantees a recall of 1.
    If no removable vertex is found at the end of for loop of line 4 of function \textbf{FindRemovable}, return the node with the smallest Mb size. 
    If the true BN is diamond-free, this will not affect the algorithm as there always exists a vertex that satisfies Equation \eqref{eq: removable-diamond} according to Lemma \ref{lemma: FindRemovable}.
    Otherwise, i.e., if the true BN contains diamonds, this implementation ensures that $RSL_D$ never gets stuck, and from Proposition \ref{prp: apd_ver} it has no false negatives in its output.
    Note that $RSL_D$ might still recover the true BN if it has diamonds. 
    
\section{Discussion on Algorithm \ref{alg: update Mb}}\label{sec: apd_alg_updat}
    Algorithm \ref{alg: update Mb} takes as input a removable variable $X$ in $\G[\OV]$ with its set of neighbors $\N{X}{\G[\OV]}$, and the set of Markov boundaries $\MB{\OV}$.
    The output of the algorithm is the set of Markov boundaries after removing $X$ from $\G[\OV]$, i.e., $\MB{\OV \setminus \{X\}}$. 
    See the proof of Proposition \ref{prp: updatemb} for more details.
    
    Function \textbf{UpdateMb} initializes $\Mb{Y}{\OV\setminus \{X\}}$ with $\Mb{Y}{\OV}$ and then removes the extra variables as follows.
    Since $X$ is removable in $\G[\OV]$, for any $Y \in \OV \setminus \{X\}$,
    \begin{equation}
        \begin{aligned}
            &\Mb{Y}{\OV} = \N{Y}{\G[\OV]} \cup \Cp{Y}{\G[\OV]}, \\
            &\Mb{Y}{\OV\setminus \{X\}} = \N{Y}{\G[\OV\setminus \{X\}]} \cup \Cp{Y}{\G[\OV\setminus \{X\}]}.
        \end{aligned}
    \end{equation}
    First, note that $\N{Y}{\G[\OV\setminus \{X\}]} \subseteq \N{Y}{\G[\OV]}$ and $\Cp{Y}{\G[\OV\setminus \{X\}]} \subseteq  \Cp{Y}{\G[\OV]}$.
    We define 
    \begin{equation}
        \begin{aligned}
            &D_1(Y) := \N{Y}{\G[\OV]} \setminus \N{Y}{\G[\OV\setminus \{X\}]}, \text{ and } \\
            &D_2(Y):= \Cp{Y}{\G[\OV]} \setminus \Cp{Y}{\G[\OV\setminus \{X\}]}.
        \end{aligned}
    \end{equation}
    One can see that $D_1(Y)$ is empty for $Y\in \OV \setminus (\Mb{X}{\OV}\cup \{X\})$, and is $\{X\}$ for $Y\in \Mb{X}{\OV}$.
    Hence, Algorithm \ref{alg: update Mb} removes $X$ from $\Mb{Y}{\OV}$ when $Y\in \Mb{X}{\OV}$ in line 4.
    
    The algorithm identifies the rest of the extra variables in lines 5-9 which are the variables in $D_2(Y)$.
    We show in the proof Proposition \ref{prp: updatemb} that if $X$ has at least one child, then $D_2(Y)$ is empty for all $Y\in \OV \setminus \{X\}$.
    In line 5 of Algorithm \ref{alg: update Mb}, if the condition $\N{X}{\G[\OV]}=\Mb{X}{\OV}$ does not hold, then $X$ has at least one co-parent, and therefore, at least one child. 
    In this case, there is no extra variables in the Markov boundaries and Algorithm \ref{alg: update Mb} correctly returns $\MB{\OV\setminus \{X\}}$. 
    
    In the next step (lines 6-9), similar to the methods presented in \cite{mokhtarian2021recursive} and \cite{margaritis1999bayesian}, for all pairs $Y,Z \in \N{X}{\G[\OV]}$ we can either perform $\CI{Y}{Z}{\Mb{Y}{\OV \setminus \{X\}}\setminus \{Y,Z\}}{P_{\OV}}$ or $\CI{Y}{Z}{\Mb{Z}{\OV \setminus \{X\}}\setminus \{Y,Z\}}{P_{\OV}}$ to decide whether they remain in each other's Markov boundary after removing $X$ from $\G[\OV]$.
	In our implementation, we chose the CI test with the smaller conditioning set among these two.

\section{Discussion on the implementation of \textbf{RSL}}\label{sec: apd_impl}
    Herein, we discuss an implementation of \textbf{RSL} that avoids unnecessary tests and reaches $\mathcal{O}(|\V|\din{\G}^3)$ number of CI tests in the case of diamond-free graphs and $\mathcal{O}(|\V|\din{\G}^{m+1})$ CI tests when the clique number is bounded by $m$. 
    
    The main idea for this implementation is to assign a boolean flag to each vertex of $\G$ with possible values "True" or "False". 
    In this implementation, the removability of a vertex will be checked by \textbf{FindRemovable} if and only if its flag is "True". 
     
    Initially, we set all the flags to be "True," and whenever a vertex is checked by the function \textbf{FindRemovable}, and it has not been identified as removable, its flag will is set to "False".
    The flag of a vertex remains "False" until its Markov boundary is changed by the function \textbf{UpdateMb}.
    This ensures that we do not check the removability of a non-removable vertex $X$ whose Markov boundary has not been updated since the last check. 
    This is because the removability of a vertex only depends on its Markov boundary. Thus, if a vertex is non-removable and its Markov boundary has not been changed, it remains non-removable. 
    
    To compute the number of CI tests performed in this implementation, it is crucial to observe that  \textbf{RSL} with the aforementioned implementation will check the removability of a vertex $X$ at most $\din{\G}$ times. 
    This is because of two reasons: (i) function \textbf{FindRemovable} only checks the removability of a vertex whose Markov boundary is bounded by $\din{\G}$ due to Lemma \ref{lemma: limb2}. 
    (ii) If a vertex is non-removable, it will be rechecked if and only if its Markov boundary has been updated. On the other hand, whenever an update happens, its Markov boundary size will decrease. Therefore, a vertex will be checked by \textbf{FindRemovable} at most $\din{\G}$ times, where each of these checks requires $\mathcal{O}(\din{\G}^2)$ and $\mathcal{O}(\din{\G}^{m})$ CI tests in the cases of diamond-free and bounded clique number, respectively. 
    In the worst-case, \textbf{RSL} requires $\mathcal{O}(|\V|\din{\G}^3)$ and $\mathcal{O}(|\V|\din{\G}^{m+1})$ CI tests in the cases of diamond-free and bounded clique number, respectively.

\section{Reproducibility and additional experiments}\label{sec: apd_rep}
    
    In this section, we provide complementary experiment results on real-world structures\footnote{All of the experiments were run in MATLAB on a MacBook Pro laptop equipped with a 1.7 GHz Quad-Core Intel Core i7 processor and a 16GB, 2133 MHz, LPDDR3 RAM.}. 

    As mentioned in Section \ref{sec: diamond-free}, diamond-free graphs appear in many real-world applications. 
    For instance, we randomly chose 17 real-world structures from a database which has become the benchmark in BN structure learning literature\footnote{https://www.bnlearn.com/bnrepository/}, and observed that 15 out of these 17 graphs were diamond-free.
    This suggests that $RSL_D$ can be employed in many real-world applications.
    It is important to note that even in the case of BNs with diamonds, our experimental results showed that $RSL_D$ might still work well on these structures (see Figure \ref{fig: Andes}, and the column corresponding to the Andes structure in Table \ref{table: real-world}.) We further discussed a theoretical guarantee in Section \ref{sec: apd_verify} for graphs containing diamonds.
    
    In what follows, we have reported the performance of BN learning algorithms in five real-world structures, namely \textit{Insurance}, \textit{Hepar2}, \textit{Diabetes}, \textit{Andes}, and \textit{Pigs}.
    Details of these structures can be found in Table \ref{tab: structure details},
    where $n,\, e,\, \omega,\,\Delta_{in},\, \Delta$, and $\alpha$ denote the number of vertices, number of edges, clique number, maximum in-degree, maximum degree, and maximum Markov boundary size of the structures, respectively.
    \begin{table*}[ht]
        \centering
        \begin{tabular}{N M{2cm}||M{1.5cm} M{1.5cm} M{1cm} M{1cm} M{1cm} M{1cm}}
            \toprule
            & Graph name & $n$ & $e$ & $\omega$ &$\Delta_{in}$ &$\Delta$ & $\alpha$\\
            \hline
            & Insurance  & 27  & 51   & 3  & 3 & 9 & 10\\
            & Hepar2     & 70  & 123  & 4  & 6 &19 & 26\\
            & Diabetes   &104  & 148  & 3  & 2 & 7 & 12\\
            & Andes      & 223 & 328  & 3  & 6 & 12 & 23\\
            & Pigs       & 441 & 592  & 2  & 2 & 41 & 68\\
            \bottomrule
        \end{tabular}
        \caption{Detailed information of the real-world structures used in the experiments.}
        \label{tab: structure details}
    \end{table*}
    
\begin{figure*}[!ht] 
        \centering
        \captionsetup{justification=centering}
        \begin{subfigure}[b]{\textwidth}
            \centering
            \includegraphics[width=0.5\textwidth]{Figures/leg.pdf}
        \end{subfigure}
        \begin{subfigure}[b]{1\textwidth}
            \centering
            \includegraphics[width=0.25\textwidth]{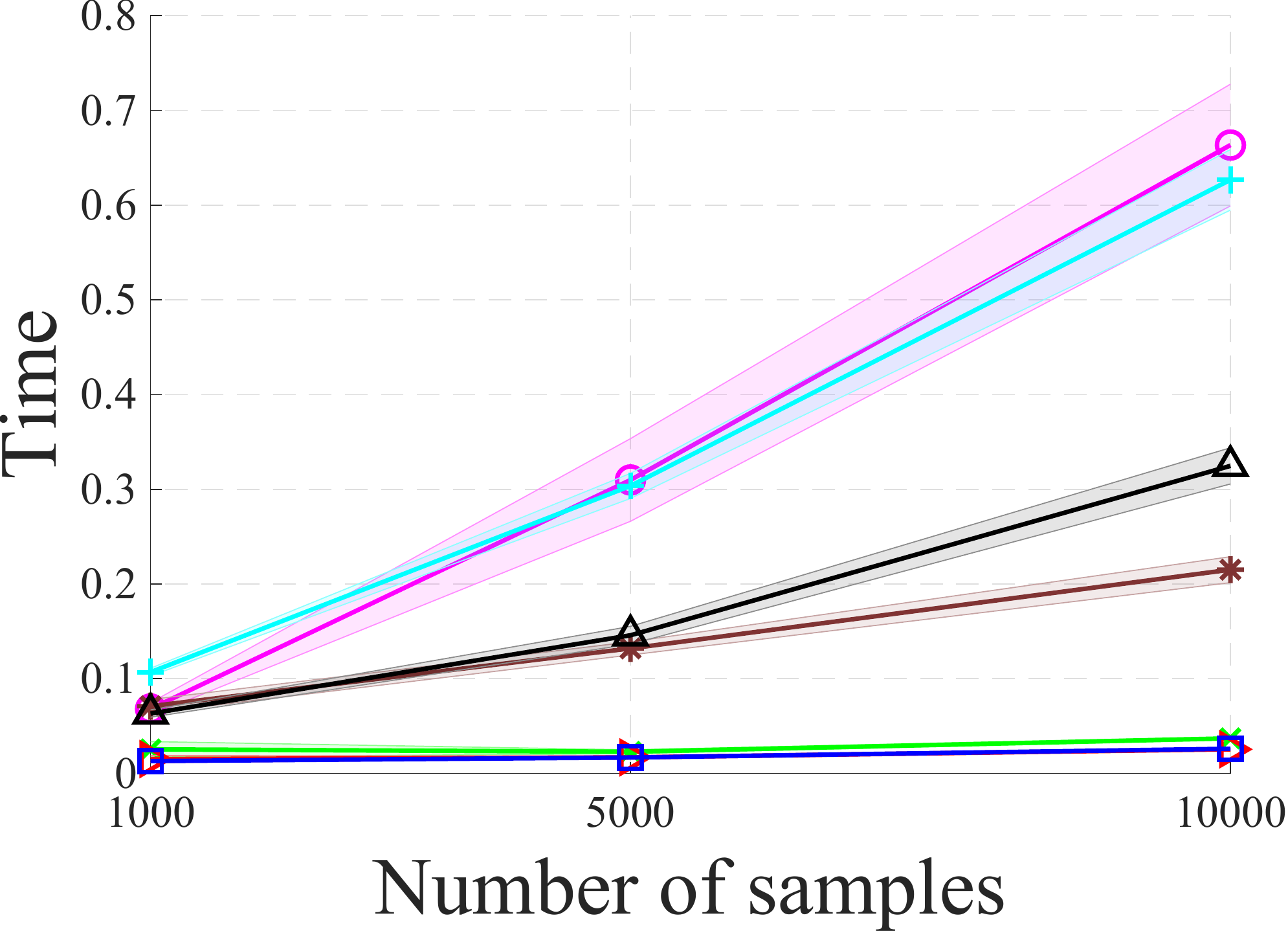}
            \hspace{0.05\textwidth}
            \includegraphics[width=0.25\textwidth]{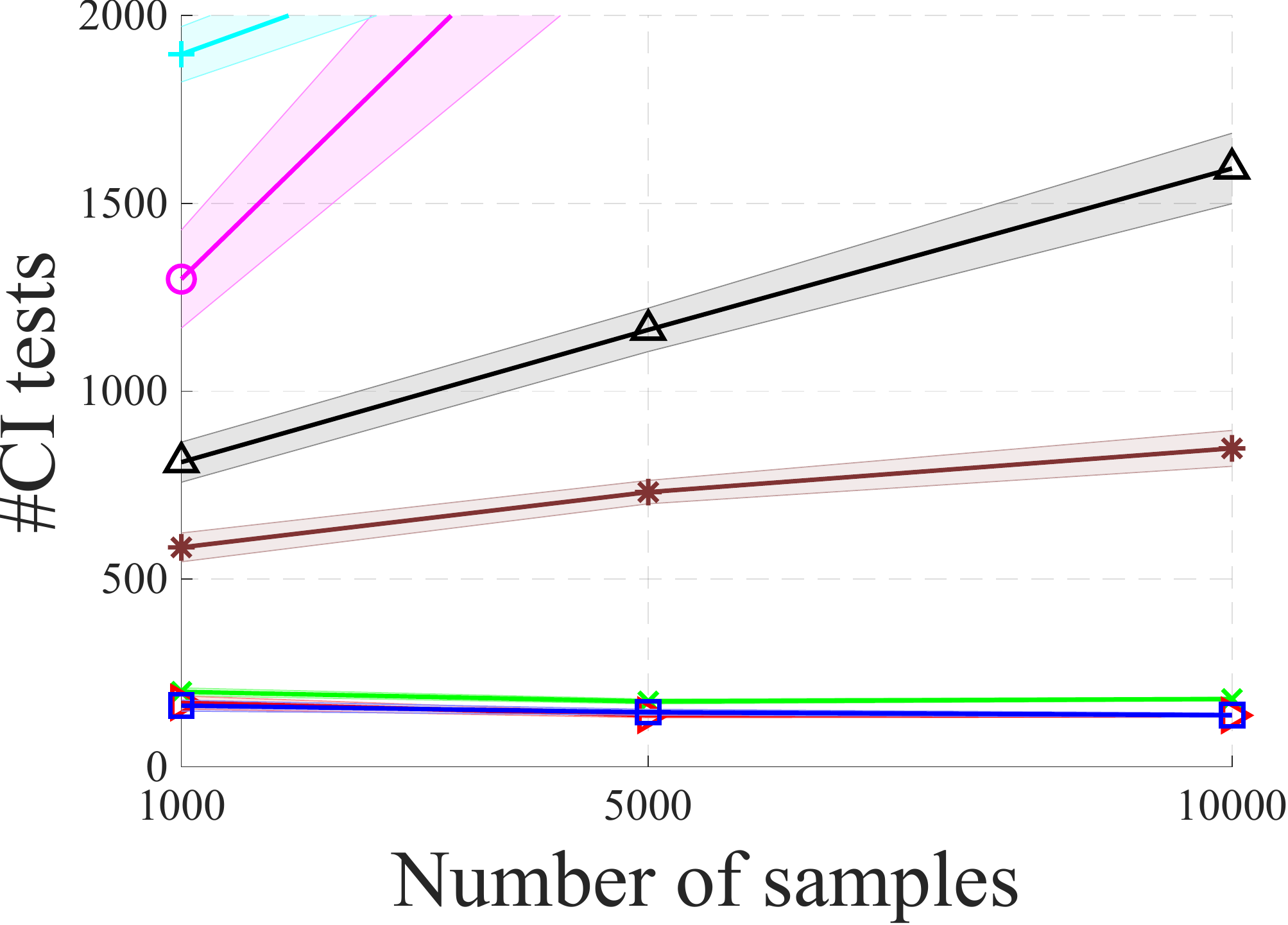}
            \hspace{0.05\textwidth}
            \includegraphics[width=0.25\textwidth]{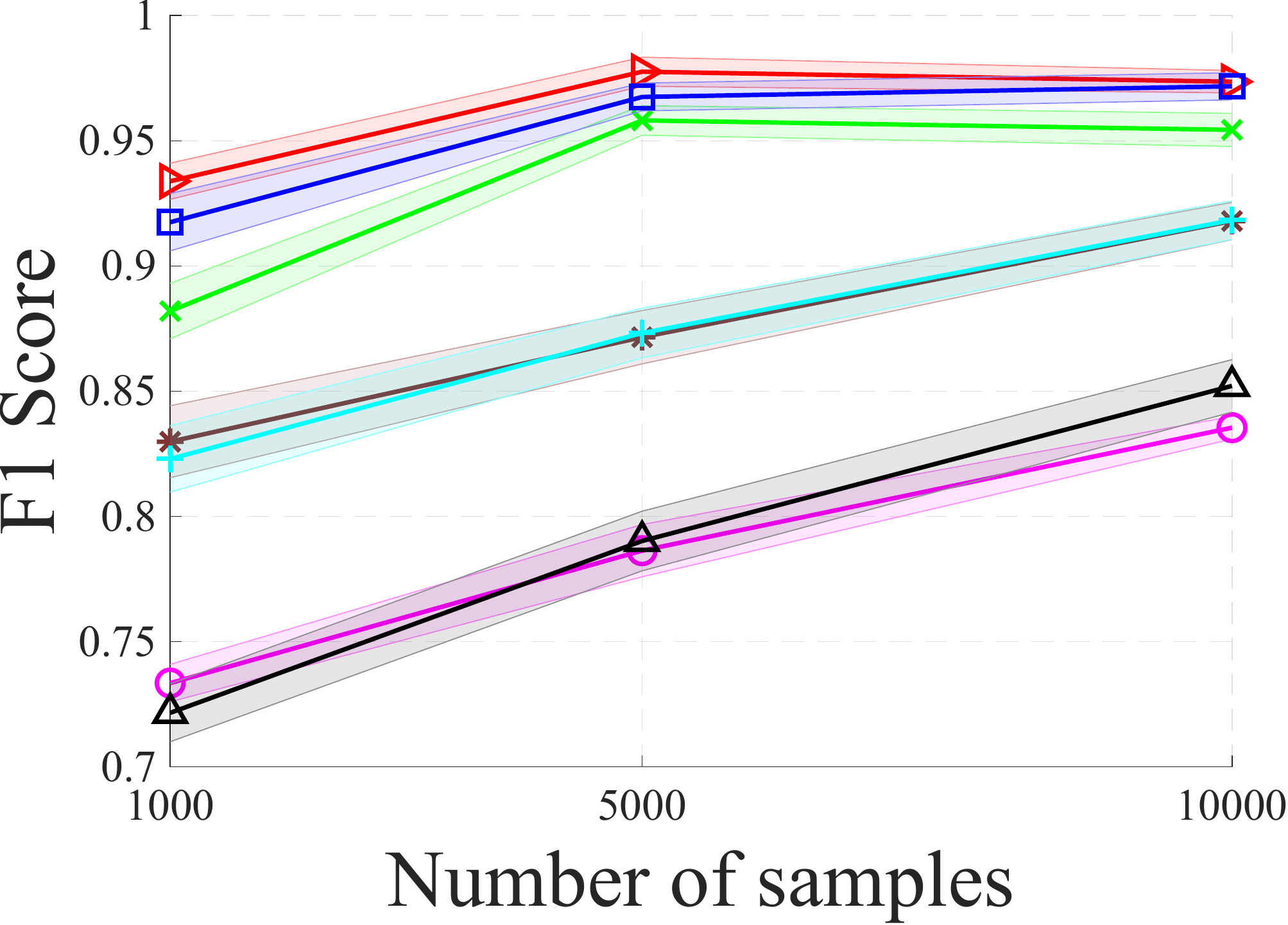}
            \caption{On data; Insurance.}
            \label{fig: Insurance}
        \end{subfigure}
        \begin{subfigure}[b]{1\textwidth}
            \centering
            \includegraphics[width=0.25\textwidth]{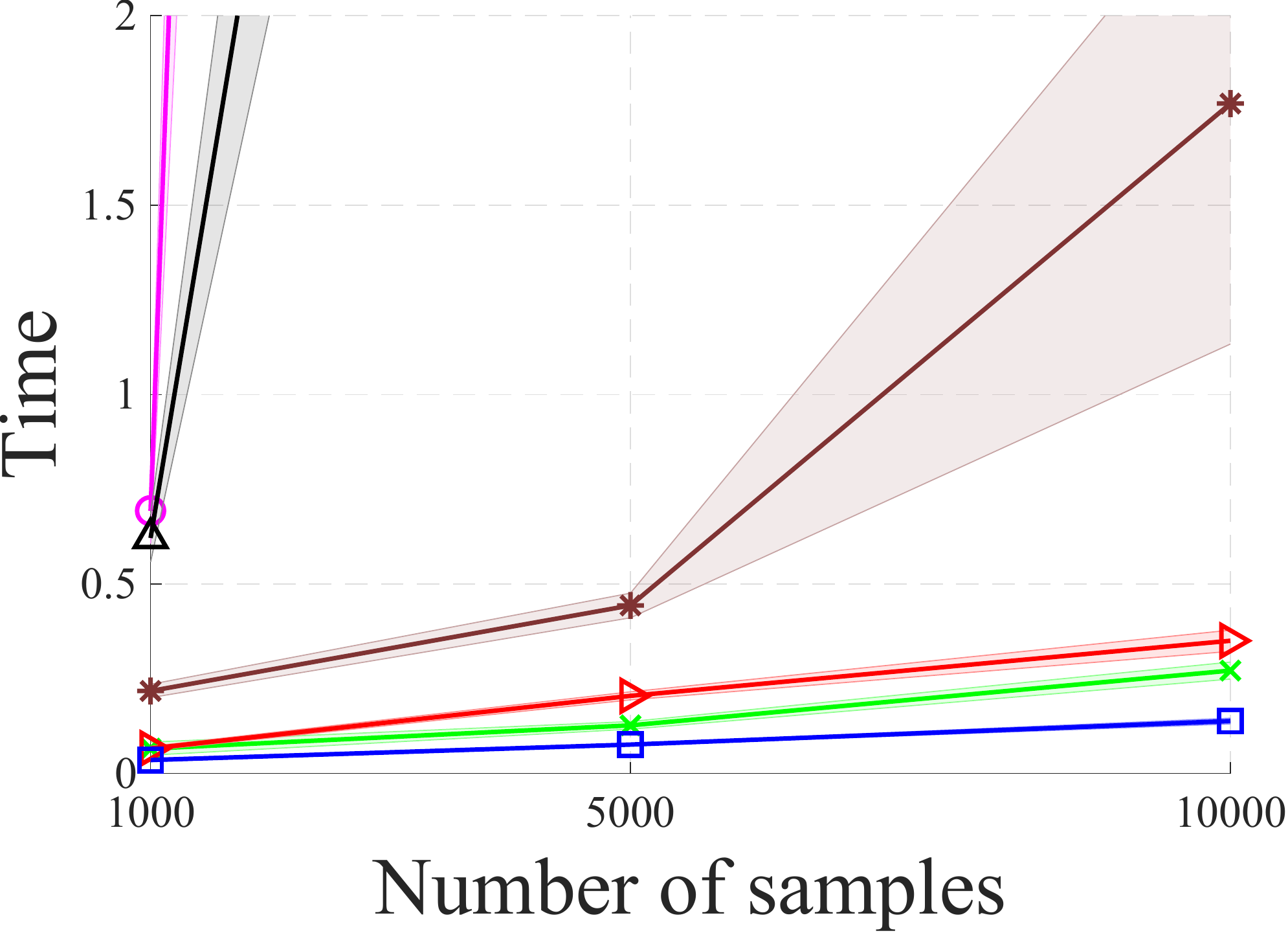}
            \hspace{0.05\textwidth}
            \includegraphics[width=0.25\textwidth]{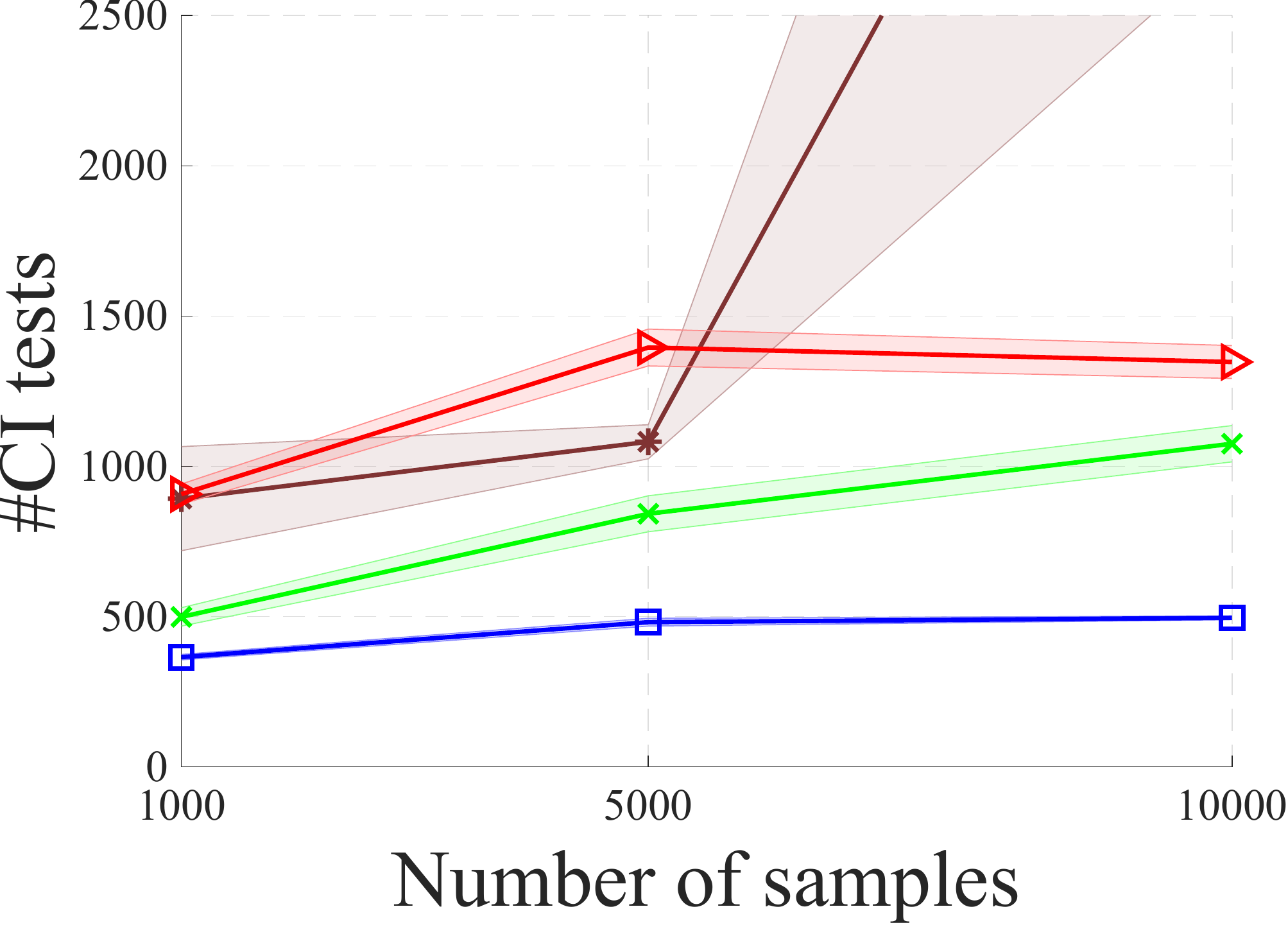}
            \hspace{0.05\textwidth}
            \includegraphics[width=0.25\textwidth]{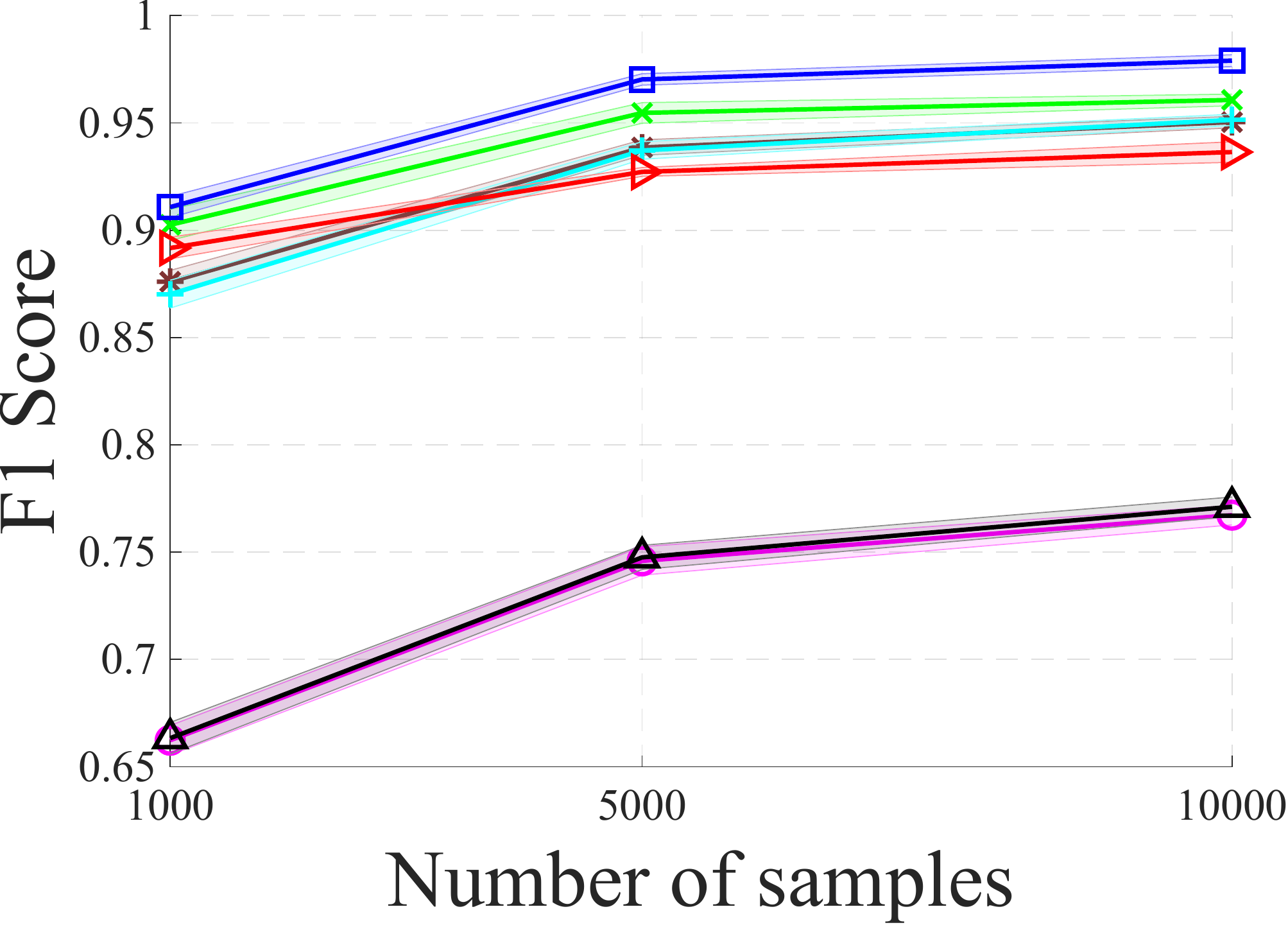}
            \caption{On data; Hepar2.}
            \label{fig: Hepar2}
        \end{subfigure}
        \caption{Performance of various algorithms on two other real-world structures.}
        \label{fig: real-world 2}
    \end{figure*}

    Figure \ref{fig: real-world 2} demonstrates the performance of the BN learning algorithms on Hepar2 and Insurance structures. The experimental setup in this figure is the same as Figure \ref{fig: real-world}. 
    As seen in Figures \ref{fig: Insurance} and \ref{fig: Hepar2}, our algorithms outperform other algorithms in both accuracy and complexity. 
    
\begin{figure*}[ht] 
        \centering
        \captionsetup{justification=centering}
            \centering
         \includegraphics[width=0.32\textwidth]{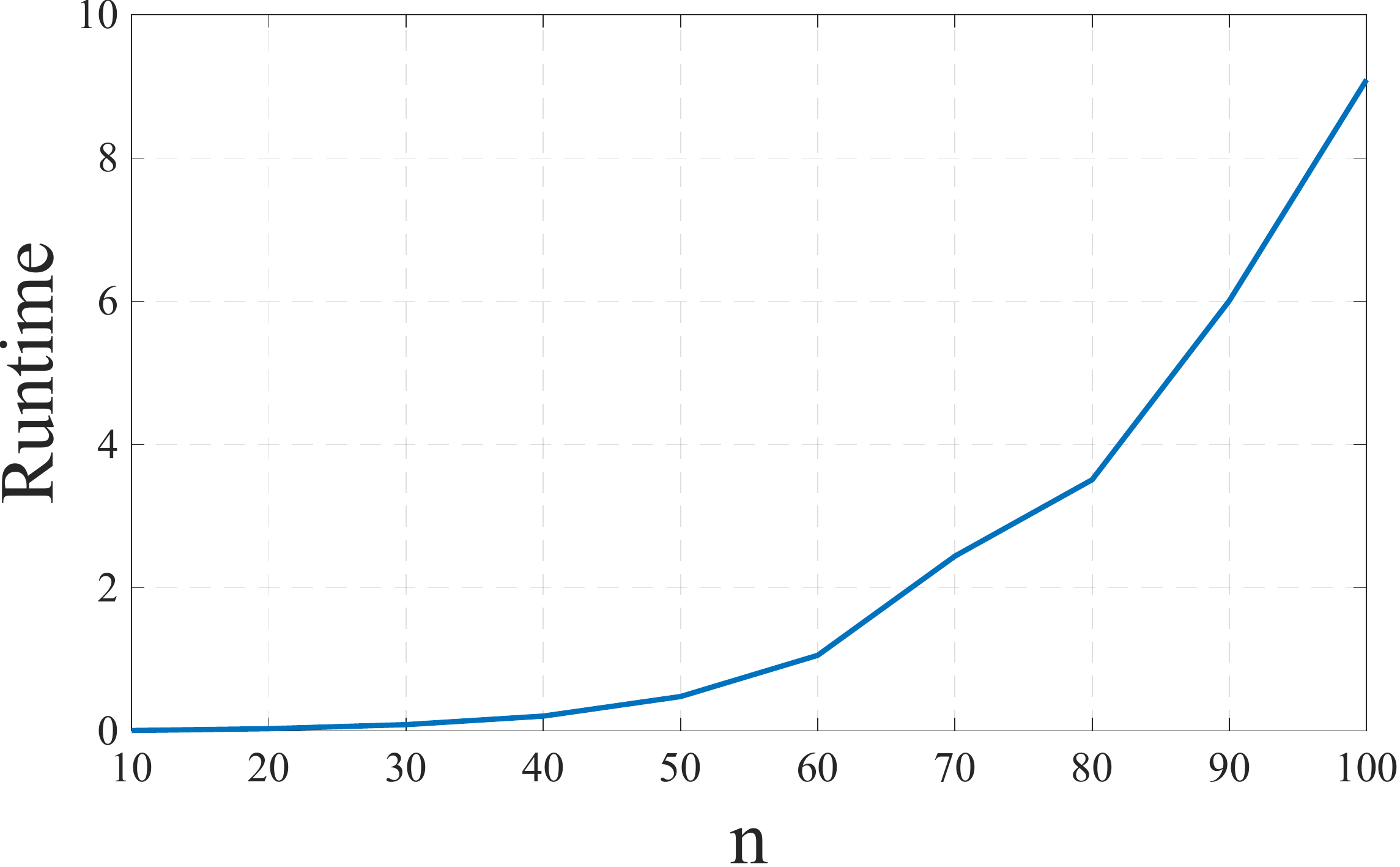}
            \caption{Runtime of TC algorithm for computing Markov boundaries on G(n,p) models where, $p=n^{-0.72}$.}
            \label{fig: TC runtim}
    \end{figure*}

    Figure \ref{fig: TC runtim} shows the runtime of TC algorithm for random Erdos-Renyi graphs when $p=n^{-0.72}$. 
    
\begin{figure*}[!ht] 
        \centering
        \captionsetup{justification=centering}
        \begin{subfigure}[b]{\textwidth}
            \centering
            \includegraphics[width=0.5\textwidth]{Figures/leg.pdf}
        \end{subfigure}
        \begin{subfigure}[b]{\textwidth}
            \includegraphics[width=0.25\textwidth]{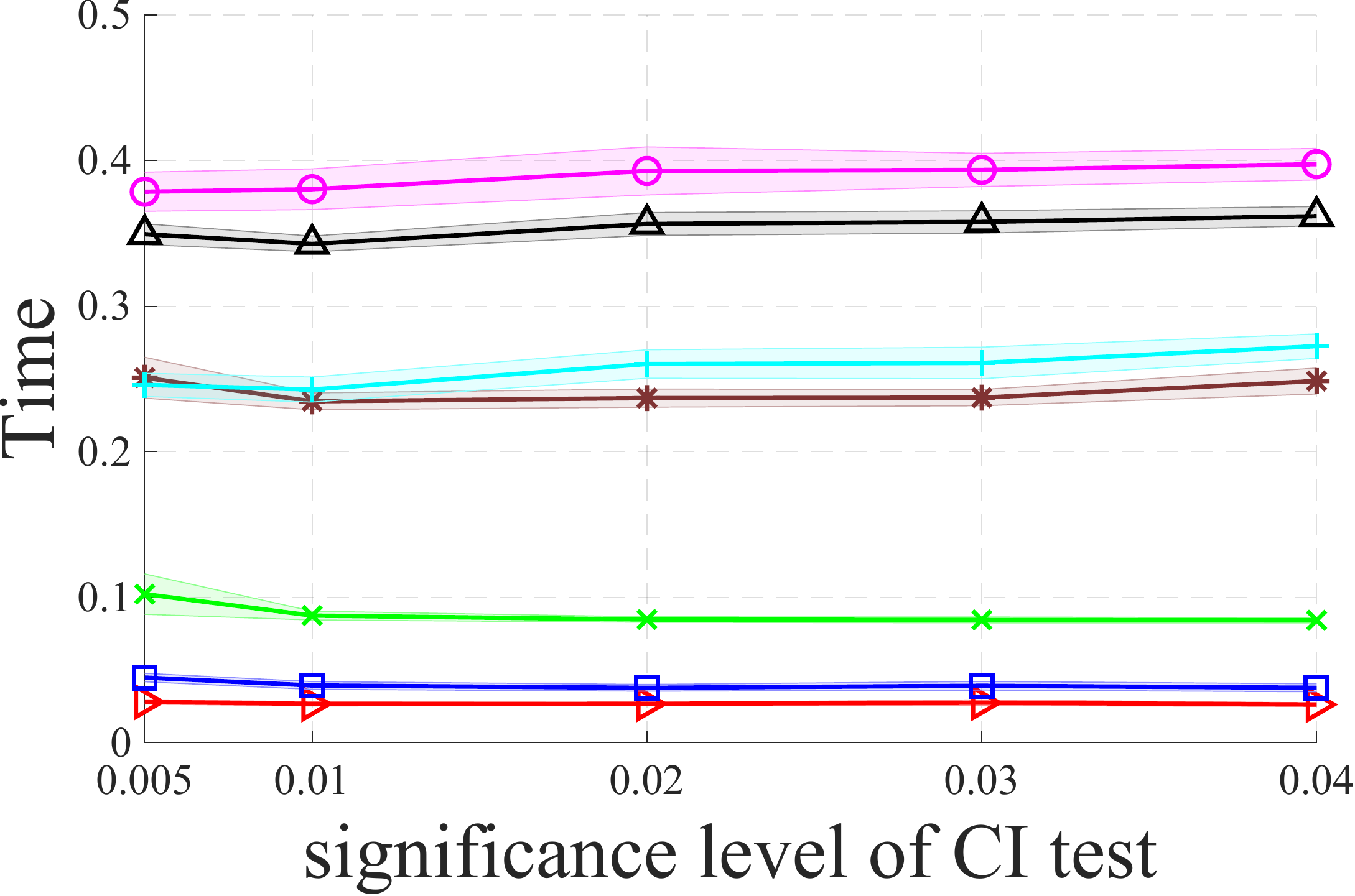}
            \hspace{0.05\textwidth}
            \includegraphics[width=0.25\textwidth]{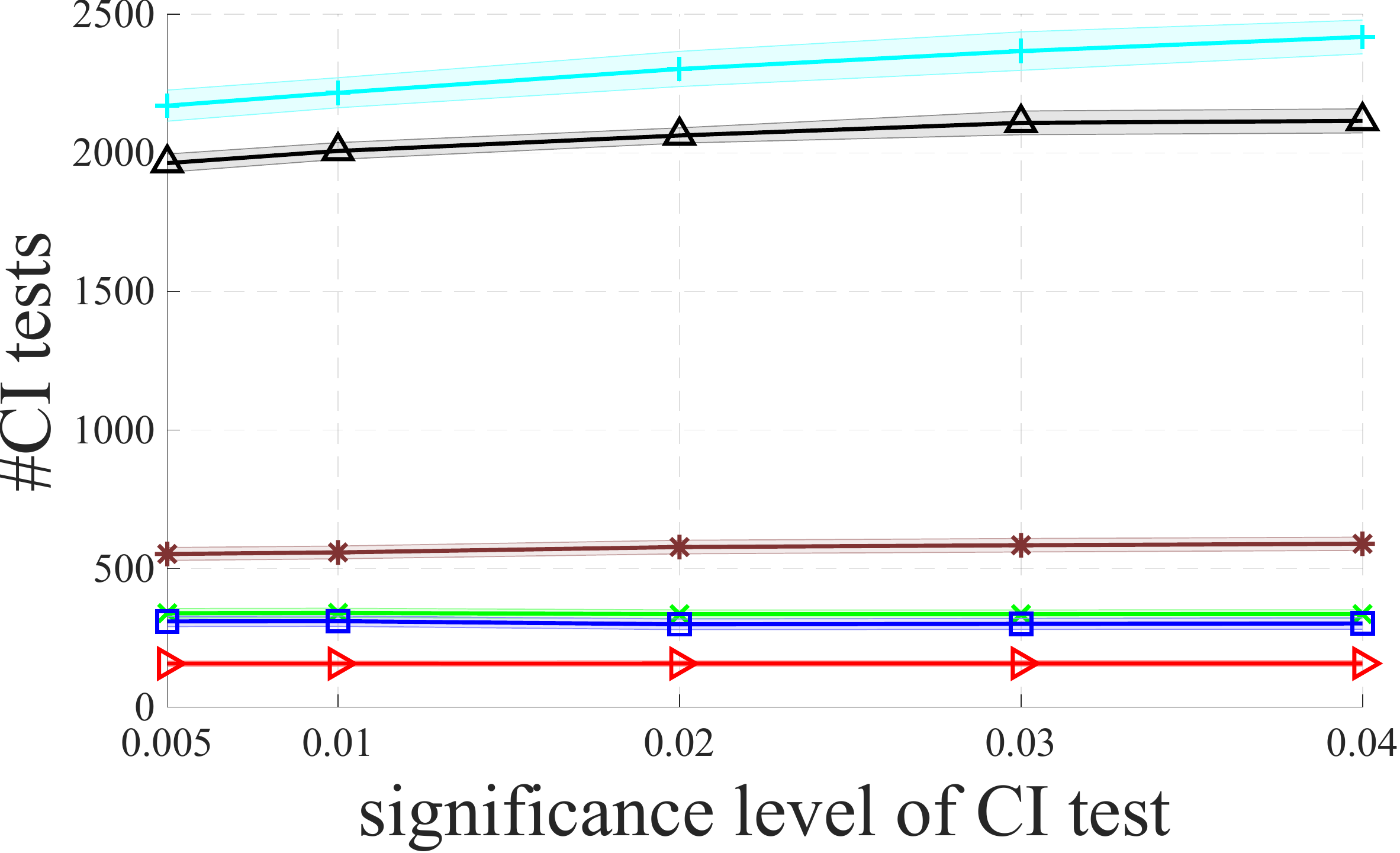}
            \hspace{0.05\textwidth}
            \includegraphics[width=0.25\textwidth]{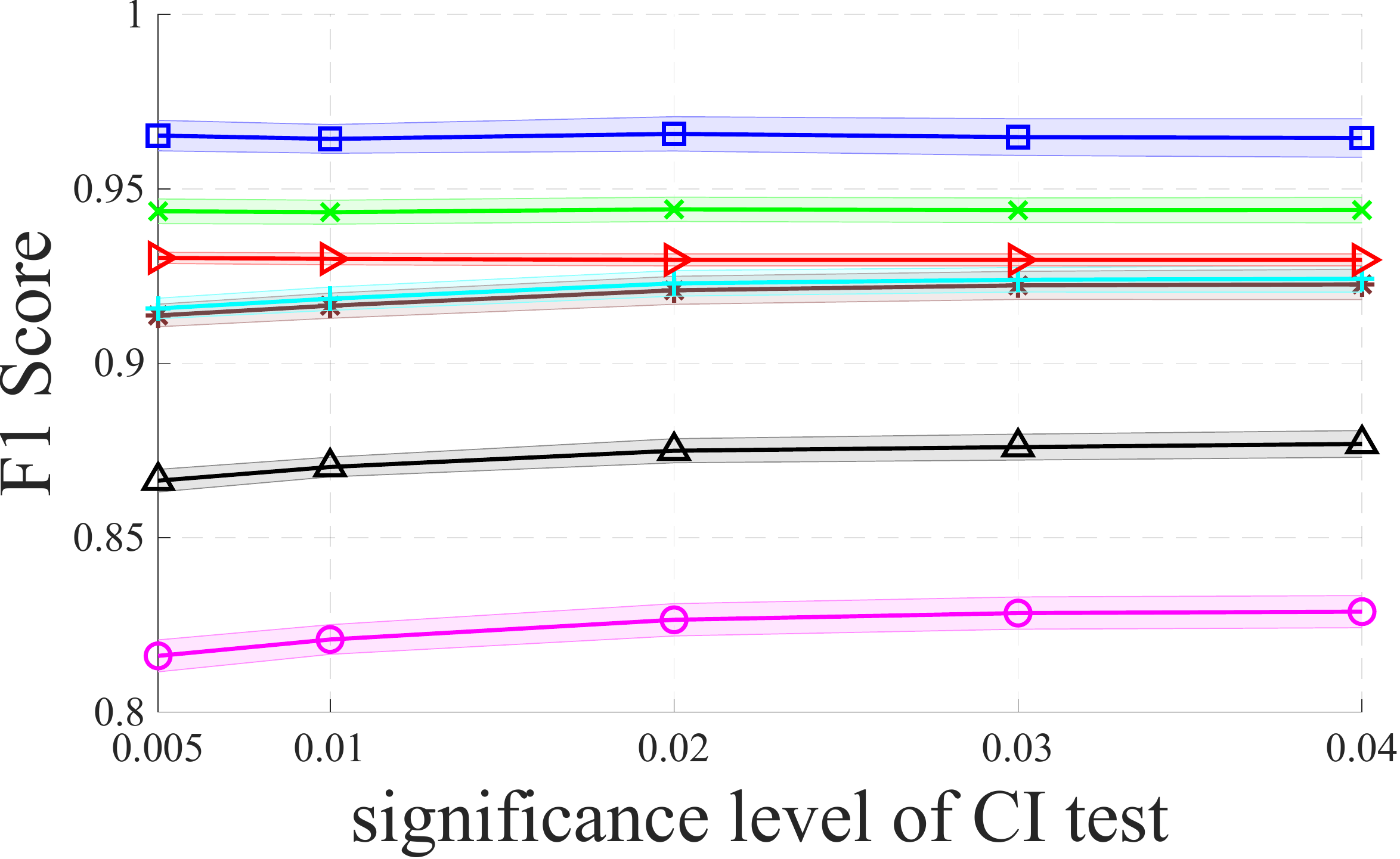}
        \end{subfigure}
        \caption{Performance of various algorithms for different values of significance level of CI test on Diabetes structure.}
        \label{fig: diabete-alpha}
    \end{figure*}

    As mentioned in Section \ref{sec: experiments}, alternative values of significance level of CI test do not change our experimental results. Figure \ref{fig: diabete-alpha} illustrates the performance of the BN learning algorithms for different values of the significance level of CI test on Diabetes structure when the number of samples is 5000. 
    As seen in this Figure, the choice of the significance level does not have any considerable effect on our results.
    
    Table \ref{table: real-world} demonstrates the experimental results on the real-world structures under two scenarios, namely the oracle setting and the finite sample setting.
    In the latter scenario (finite sample), the algorithms have access to a dataset with 10,000 samples of each variable.
    The number of CI tests, Average Size of Conditioning sets (ASC), Accuracy of Learned Separating Sets (ALSS), and runtime of the algorithms are reported after Markov boundary discovery.
    Structural Hamming Distance (SHD) is calculated as the sum of the number of extra edges and missing edges.
    Analogous to the the formerly reported results, both RSL algorithms outperform other algorithms in terms of both accuracy and computational complexity.
    
    \begin{table*}[!ht] 
	    \fontsize{9}{10.5}\selectfont
	    \centering
	    \begin{tabular}{N M{1.2cm}|M{0.9cm}| M{1.12cm}| M{1.12cm}M{1.12cm} M{1.12cm} M{1.12cm}M{1.12cm}}
     		\toprule
			& \multicolumn{3}{c|}{\multirow{2}{*}{}}
			& Insurance
			& Hepar2
			& Diabetes
			& Andes*
			& Pigs
			\\
			& \multicolumn{3}{c|}{}
			& $n=27$
			& $n=70$ 
			& $n=104$ 
			& $n=223$
			& $n=441$
 			\\
			\hline
			& \multirow{7}{*}{$RSL_D$}
			& \multirow{2}{*}{Oracle}
			& \#CI tests
			& 124
			& \textbf{477}
			& 250
			& \textbf{1,623}
			& \textbf{1,059}
			 \\
			&
			&
			& ASC
			& 1.18
			& 3.38
			& \textbf{0.55}
			& 2.20
			& \textbf{0.55}
			 \\
			\cline{3-4}
			&
			& \multirow{5}{*}{\shortstack{finite\\sample}}
			& runtime
			& 0.06
			& \textbf{0.18}
			& 0.09
			& \textbf{0.51}
			& 0.63
			\\
			&
			&
			& F1-score
			& \textbf{0.99}
			& \textbf{0.99}
			& \textbf{0.96}
			& \textbf{0.86}
			& \textbf{0.99}
			\\
			&
			&
			& precision
        	& \textbf{1}
        	& 0.99 
        	& 0.96
        	& 0.88
        	& \textbf{0.99}
        	\\
        	&
        	&
        	& recall
        	& \textbf{0.98}
        	& 0.97
        	& 0.95
        	& 0.84
        	& \textbf{0.99}
        	\\
        	&
        	&
        	& SHD
        	& \textbf{1}
        	& \textbf{5}
        	& \textbf{13}
        	& \textbf{94}
        	& 11
        	\\
        	&
        	&
        	& ALSS
        	& 0.91
        	& \textbf{0.89}
        	& 0.87
        	& \textbf{0.68}
        	& \textbf{0.99}
			\\
			\hline
			& \multirow{7}{*}{$RSL_\omega$}
			& \multirow{2}{*}{Oracle}
			& \#CI tests
			& \textbf{118}
			& 1,218
			& \textbf{140}
			& 2,884
			& \textbf{1,059}
			\\
			&
			&
			& ASC
			& 1.42 
			& 2.72
 			& 1.27
 			& 2.40
 			& \textbf{0.55}
 			\\
			\cline{3-4}
			&
			& \multirow{5}{*}{\shortstack{finite\\sample}}
			& runtime
			& \textbf{0.04}
			& 0.49
			& \textbf{0.06}
			& 0.94
			& 0.63
			\\
			&
			&
			& F1-score
			& 0.98
			& 0.94
			& 0.92
			& 0.82
			& \textbf{0.99}
            \\
            &
            &
            & precision
        	& \textbf{0.98}
        	& 0.89
        	& 0.85
        	& 0.77
        	& \textbf{0.99}
        	\\
        	&
        	&
        	& recall
        	& 0.98
        	& \textbf{0.99}
        	& \textbf{1}
        	& \textbf{0.88}
        	& \textbf{0.99}
        	\\
        	&
        	&
        	& SHD
        	& 2
        	& 16
        	& 25
        	& 130
        	& 11
        	\\
        	&
        	&
        	& ALSS
        	& \textbf{0.92}
        	& 0.86
        	& \textbf{0.97}
        	& 0.60
        	& \textbf{0.99}
			\\
			\hline
			& \multirow{7}{*}{MARVEL}
			& \multirow{2}{*}{Oracle}
			& \#CI tests
			& 148
			& 1,663
			& 420
			& 6,072
			& 1,119
 			\\
			&
			&
			& ASC
			& \textbf{1.08}
			& \textbf{2.34}
 			& 0.71
 			& \textbf{1.42}
 			& 0.58
 			\\
			\cline{3-4}
			&
			& \multirow{5}{*}{\shortstack{finite\\sample}}
			& runtime
			& 0.11
			& 0.36
			& 0.18
			& 2.92
			& \textbf{0.29}
			\\
			&
			&
			& F1-score
			& \textbf{0.99}
			& 0.95
			& 0.95
			& 0.77
			& \textbf{0.99}
			\\
			&
			&
			& precision
        	& \textbf{1}
        	& 0.94
        	& 0.93
        	& 0.80
        	& 0.98
        	\\
        	&
        	&
        	& recall
        	& \textbf{0.98}
        	& 0.95
        	& 0.96
        	& 0.74
        	& \textbf{0.99}
        	\\
        	&
        	&
        	& SHD
        	& \textbf{1}
        	& 13
        	& 16
        	& 150
        	& 14
        	\\
    	    \hline
			& \multirow{7}{*}{CS}
			& \multirow{2}{*}{Oracle}
			& \#CI tests
			& 1,079
			& 34,083
 			& 901
 			& 45,697
 			& 20,634
 			\\
			&
			&
			& ASC
			& 2.65
			& 8.25
 			& 2.15
 			& 5.20
 			& 7.94
 			\\
			\cline{3-4}
			&
			& \multirow{5}{*}{\shortstack{finite\\sample}}
			& runtime
			& 0.38
			& 0.78
			& 0.52
			& 10.98
			& 6.99
			\\
			&
			&
			& F1-score
			& 0.93
			& 0.93
			& 0.93
			& 0.79
			& \textbf{0.99}
            \\
            &
            &
            & precision
        	& \textbf{1}
        	& \textbf{1}
        	& 0.99
        	& 0.97
        	& \textbf{0.99}
        	\\
        	&
        	&
        	& recall
        	& 0.86
        	& 0.87
        	& 0.88
        	& 0.66
        	& \textbf{0.99}
        	\\
        	&
        	&
        	& SHD
        	& 7
        	& 16
        	& 19
        	& 121
        	& 12
        	\\
        	 \hline
	
			& \multirow{7}{*}{GS}
			& \multirow{2}{*}{Oracle}
			& \#CI tests
			& 1,683
			& 67,408
 			& 1,923
 			& 56,512
 			& 146,888
 			\\
			&
			&
			& ASC
			& 2.88
			& 7.85
 			& 2.51
 			& 5.37
 			& 8.11
 			\\
			\cline{3-4}
			&
			& \multirow{5}{*}{\shortstack{finite\\sample}}
			& runtime
			& 0.75
			& 204.85
			& 0.61
			& 12.36
			& 8.08
			\\
			&
			&
			& F1-score
			& 0.93
			& 0.74
			& 0.93
			& 0.78
			& \textbf{0.99}
			\\
			&
			&
			& precision
        	& \textbf{1}
        	& 0.97
        	& \textbf{1}
        	& \textbf{0.98}
        	& \textbf{0.99}
        	\\
        	&
        	&
        	& recall
        	& 0.86
        	& 0.60
        	& 0.88
        	& 0.65
        	& \textbf{0.99}
        	\\
        	&
        	&
        	& SHD
        	& 7
        	& 51
        	& 18
        	& 122
        	& \textbf{6}
        	\\
			\hline
			& \multirow{7}{*}{PC}
			& \multirow{2}{*}{Oracle}
			& \#CI tests
			& 4,912
			& $>10^7$
 			& 3,873
 			& 40,773
 			& $>10^{13}$
 			\\
			&
			&
			& ASC
			& 3.23
			& NA
			& 2.20
 			& 4.30
 			& NA
 			\\
			\cline{3-4}
			& 
			& \multirow{5}{*}{\shortstack{finite\\sample}}
			& runtime
			& 0.35
			& 9.51
			& 0.64
			& 1.26
			& NA
			\\
			&
			&
			& F1-score
			& 0.85
			& 0.76
			& 0.87
			& 0.74
			& NA
			\\
			&
			&
			& precision
        	& \textbf{1}
        	& \textbf{1}
        	& 0.97
        	& \textbf{0.98}
        	& NA
        	\\
        	&
        	&
        	& recall
        	& 0.74
        	& 0.61
        	& 0.80
        	& 0.59
        	& NA
        	\\
        	&
        	&
        	& SHD
        	& 13
        	& 48
        	& 34
        	& 141
        	& NA
			\\
			\hline
		    &\multirow{7}{*}{MMPC}
			& \multirow{2}{*}{Oracle}
			& \#CI tests
			& 4,582
			& $> 10^6$
			& 3,543
			& $>10^6$
			& $>10^6$
 			\\
			&
			&
			& ASC
			& 3.35
			& NA
 			& 2.23
 			& NA
 			& NA
 			\\
			\cline{3-4}
			&
			& \multirow{5}{*}{\shortstack{finite\\sample}}
			& runtime
			& 0.56
			& 198.8
			& 0.71
			& 2.70
			& NA
			\\
			&
			&
			& F1-score
			& 0.82
			& 0.74
			& 0.84
			& 0.69
			& NA
			\\
			&
			&
			& precision
        	& 0.90
        	& 0.97
        	& 0.89
        	& 0.89
        	& NA
        	\\
        	&
        	&
        	& recall
        	& 0.74
        	& 0.60
        	& 0.80
        	& 0.57
        	& NA
        	\\
        	&
        	&
        	& SHD
        	& 17
        	& 51
        	& 45
        	& 169
        	& NA
			\\
    \bottomrule
	    \end{tabular}
	    \caption{Performance of various algorithms on real-world structures. The structure indicated with star is not diamond-free. NA indicates that the corresponding value is unknown due to a very long runtime.}
	    \label{table: real-world}
    \end{table*}

\clearpage
\onecolumn

\end{document}